\def\paperID{10772}
\def\confName{CVPR}
\def\confYear{2024}

\def\paperTitle{AUEditNet: Dual-Branch Facial Action Unit Intensity Manipulation with \\ Implicit Disentanglement}

\def\authorBlock{
    Shiwei Jin $^{1}$, 
    Zhen Wang $^{2}$, 
    Lei Wang $^{2}$, 
    Peng Liu $^{2}$, 
    Ning Bi $^{2}$, 
    Truong Nguyen $^{1}$ \\
    $^{1}$ ECE Dept. UC San Diego, $^{2}$ Qualcomm Technologies, Inc. \\
    {\tt\small \{sjin, tqn001\}@eng.ucsd.edu, \{zhewang, wlei, peli, nbi\}@qti.qualcomm.com}
}

\newif\ifreview 
\newif\ifarxiv 
\newif\ifcamera \newcommand{\cameraready}{\cameratrue}
\newif\ifrebuttal 

\cameraready 

\pdfoutput=1
\documentclass[10pt,twocolumn,letterpaper]{article}
\ifreview \usepackage[review]{cvpr} \fi
\ifarxiv \usepackage[pagenumbers]{cvpr} \fi
\ifrebuttal \usepackage[rebuttal]{cvpr} \fi
\ifcamera \usepackage{cvpr} \fi

\usepackage{graphicx}	
\usepackage{amsmath}	
\usepackage{amssymb}	
\usepackage{booktabs}
\usepackage{times}
\usepackage{microtype}
\usepackage{epsfig}
\usepackage[table,xcdraw,dvipsnames]{xcolor}
\usepackage{caption}
\usepackage{float}
\usepackage{placeins}
\usepackage{color, colortbl}
\usepackage{stfloats}
\usepackage{enumitem}
\usepackage{tabularx}
\usepackage{xstring}
\usepackage{multirow}
\usepackage{xspace}
\usepackage{url}
\usepackage{subcaption}
\usepackage{xcolor}
\usepackage[hang,flushmargin]{footmisc}
\usepackage{ulem}

\ifcamera \usepackage[accsupp]{axessibility} \fi

% ...

\ifarxiv  \fi

\newcommand{\R}[1]{{%
    \textbf{%
        \ifstrequal{#1}{1}{\textcolor{red}{R#1}}{%
        \ifstrequal{#1}{2}{\textcolor{blue}{R#1}}{%
        \ifstrequal{#1}{3}{\textcolor{magenta}{R#1}}{%
        \ifstrequal{#1}{4}{\textcolor{teal}{R#1}}{%
                           \textcolor{cyan}{R#1}%
        }}}}%
    }%
}}

\usepackage{xr-hyper}

\makeatletter
\newcommand*{\addFileDependency}[1]{
  \typeout{(#1)}
  \@addtofilelist{#1}
  \IfFileExists{#1}{}{\typeout{No file #1.}}
}

\makeatother

\usepackage{arydshln}

\definecolor{cvprblue}{rgb}{0.21,0.49,0.74}
\usepackage[pagebackref,breaklinks,colorlinks,citecolor=cvprblue]{hyperref}
\usepackage[capitalize]{cleveref}
\crefname{section}{Sec.}{Secs.}
\crefname{table}{Table}{Tables}
\crefname{figure}{Fig.}{Figs.}

\begin{document}
\title{\paperTitle}
\author{\authorBlock}
\maketitle

\begin{abstract}
Facial action unit (AU) intensity plays a pivotal role in quantifying fine-grained expression behaviors, which is an effective condition for facial expression manipulation. 
However, publicly available datasets containing intensity annotations for multiple AUs remain severely limited, often featuring a restricted number of subjects. 
This limitation places challenges to the AU intensity manipulation in images due to disentanglement issues, leading researchers to resort to other large datasets with pretrained AU intensity estimators for pseudo labels.
In addressing this constraint and fully leveraging manual annotations of AU intensities for precise manipulation, we introduce AUEditNet.
Our proposed model achieves impressive intensity manipulation across 12 AUs, trained effectively with only 18 subjects. 
Utilizing a dual-branch architecture, our approach achieves comprehensive disentanglement of facial attributes and identity without necessitating additional loss functions or implementing with large batch sizes.
This approach offers a potential solution to achieve desired facial attribute editing despite the dataset’s limited subject count.
Our experiments demonstrate AUEditNet's superior accuracy in editing AU intensities, affirming its capability in disentangling facial attributes and identity within a limited subject pool. 
AUEditNet allows conditioning by either intensity values or target images, eliminating the need for constructing AU combinations for specific facial expression synthesis. 
Moreover, AU intensity estimation, as a downstream task, validates the consistency between real and edited images, confirming the effectiveness of our proposed AU intensity manipulation method. 
\end{abstract}
\section{Introduction}
\label{sec:intro}
Facial action units (AUs), serving as anatomical indicators of facial muscle movements, have been effectively utilized as conditions for fine-grained facial expression editing in images \cite{pumarola2018ganimation, ling2020toward}. 
The manipulation of AU intensities offers advantages such as 
objective quantification on a six-integer-level ordinal scale defined by the Facial Action Coding System (FACS) \cite{ekman1978facial}, 
the ability to generate over $7000$ combinations in observed facial expressions with a small number of AUs ($30$) \cite{scherer1982emotion}, 
and the potentials for continuous intensity manipulation, instead of the category-based expression editing \cite{ding2018exprgan}. 
However, public datasets containing intensity annotations for over $10$ AUs are constrained by limited subject counts, and frame-level AU intensity annotation requires expert involvement and extensive works. 
As a result, current AU intensity manipulation methods \cite{pumarola2018ganimation, ling2020toward, tripathy2020icface} often resort to the pretrained AU intensity estimator \cite{baltrusaitis2018openface} to obtain predicted annotations for datasets with larger subject pools, sidestepping the reliance on expert-labeled datasets with a restricted number of subjects. 

On the other hand, the semantic richness in latent space and high-quality generation capability of StyleGAN \cite{karras2020analyzing} have facilitated the development of facial attribute editing methods \cite{harkonen2020ganspace, shen2020interfacegan, dalva2022vecgan, do2023quantitative, lyu2023deltaedit} that enable targeted modifications without affecting other attributes and identity. 
However, searching unified editing directions in the latent space for attribute editing typically requires substantial data from numerous subjects to disentangle the target attributes from others and identity. 
Limited number of subjects may lead to overfitting issues and poor generalization to new faces.

Considering these, it is challenging to search disentangled editing directions for manipulating intensities of multiple AUs based on the data from limited subjects. 
To address this, we propose a method to manipulate intensities of $12$ AUs within the $W^+$ latent space \cite{abdal2019image2stylegan} of StyleGAN \cite{karras2020analyzing} for high-resolution face image synthesis using only $18$ subjects' data. 
Specifically, we introduce a novel pipeline designed to enforce disentanglement within the network, even with a dataset containing a limited number of subjects compared to the number of target facial attributes we aim to edit. 
This approach offers a potential solution to achieve desired facial attribute editing despite the dataset's limited subject count. 
To summarize, our contributions are as follows:
\begin{itemize}[leftmargin=2em]
    \item Achieve accurate AU intensity manipulation in high-resolution synthesized face images conditioned by AU intensity values or target images without requiring network retraining or extra AU estimators. 
    \item Introduce an architecture designed to disentangle target attributes from others and identity, even when working with data containing very few subjects compared to the number of target facial attributes we aim to edit. 
    \item Propose the encoding of labels to match the level-wise disentangled structure of latent vectors in $W^+$ to avoid entangled labels as conditions for editing. 
    \item Demonstrate the ability to manipulate float or negative AU intensities while generating consistent results, despite the training set labels encompassing six levels.  
\end{itemize}

\section{Related Work}
\label{sec:related}

\subsection{AU Intensity Manipulation}
\label{subsec:au_intensity_manipulation}
GANimation \cite{pumarola2018ganimation} is an early work that utilizes AU intensities as conditions for facial expression manipulation. 
However, it suffers from attention mechanism issues that could result in overlap artifacts in regions where facial deformations occur \cite{pumarola2020ganimation}. 
Ling \textit{et al.} \cite{ling2020toward} propose using the relative AU intensities between the source and target images as conditions, avoiding the direct addition of new attributes onto the existing expression \cite{tripathy2020icface}.
Alternatively, ICface \cite{tripathy2020icface} introduces a two-stage editing pipeline. 
The initial stage transforms the input image into a neutral one with all AU intensities set to zero, and the second stage maps this neutral status to the final output, depicting the desired driving attributes with two independent generators. 
However, the architecture of ICface is redundant and resource-intensive. 
FACEGAN \cite{tripathy2021facegan} utilizes AU representations to construct facial landmarks for expression transfer, reducing the potential of identity leakage from the target image. 
These methods place greater emphasis on facial expressions compared to AUs, both in terms of their editing goals and evaluation criteria. 

\subsection{Image Editing in Latent Space}
\label{subsec:latent_manipulation}
The latent space working with StyleGAN2 \cite{karras2020analyzing} is well-known of its meaningful and highly disentangled properties.
Several unsupervised methods \cite{harkonen2020ganspace, voynov2020unsupervised, yuksel2021latentclr, shen2021closed} search editing directions in the latent space without the need for attributes labels. 
For instance, GANSpace \cite{harkonen2020ganspace} employs principal component analysis to identify semantic editing directions in the latent space. 
In contrast, supervised methods \cite{shen2020interfacegan, dalva2022vecgan, jin2023redirtrans, do2023quantitative} typically rely on pretrained attribute estimators or attribute labels. 
InterFaceGAN \cite{shen2020interfacegan}, for example, utilizes a binary support vector machine \cite{noble2006support} to estimate hyperplanes for the corresponding attribute editing. 
Furthermore, some methods \cite{patashnik2021styleclip, zhu2022one, lyu2023deltaedit} use the CLIP loss \cite{radford2021learning} to enable text-driven image manipulation. 
These methods usually handle identity information effortlessly since commonly used datasets contain a much larger number of subjects compared to the attributes involved in the editing process.
However, in certain cases with a limited number of subjects included, the identity issue becomes significant. 
Therefore, in our work, we introduce a novel architecture designed to implicitly disentangle identity information from multiple attributes, even when dealing with a restricted number of subjects.


\section{Proposed Method}
\label{sec:method}
\begin{figure*}  
\begin{center}  
    \includegraphics[width=0.95\linewidth]{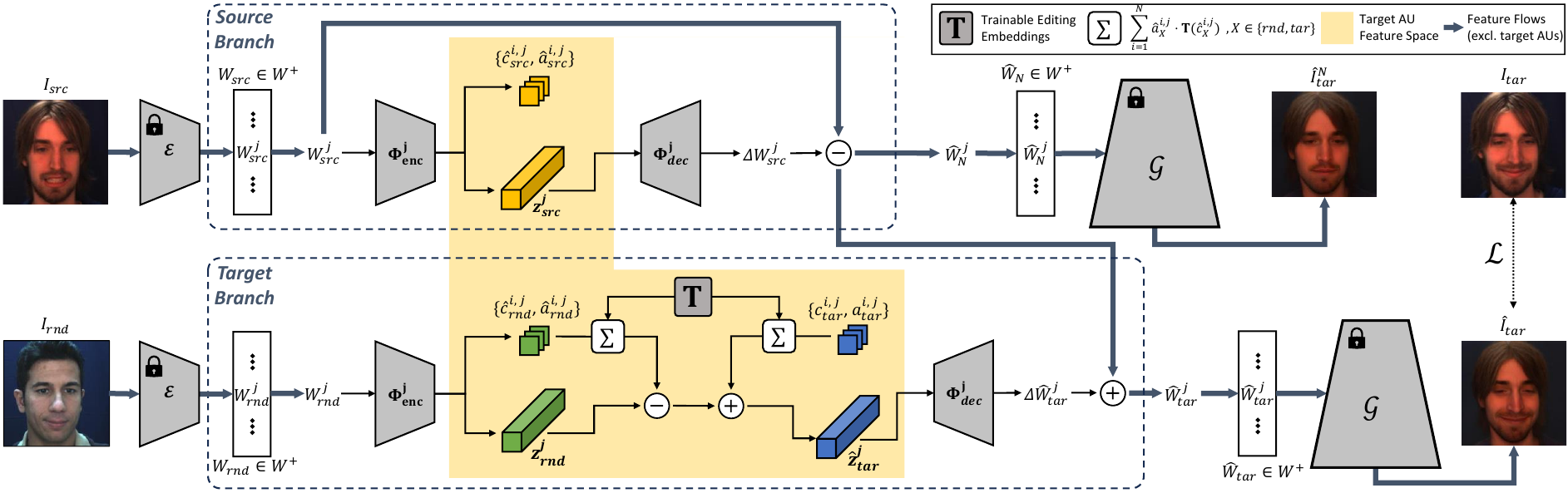}
    \caption{Overall scheme of the proposed AUEditNet. 
    AUEditNet has a dual-branch structure that separately addresses source attribute removal (\textit{Source Branch}) and target attribute addition (\textit{Target Branch}). 
    The \textit{Source Branch} aims at removing the original status in $I_{src}$, maintaining other attributes and identity while keeping them distinct from the feature space of target facial attributes (highlighted in \textit{yellow}). 
    The \textit{Target Branch} focuses on determining an edited direction $\Delta \hat W^j_{tar}$ for the new status of the target facial attribute, ensuring its independence from identity and other facial attributes. 
    Instead of applying this branch directly to $I_{src}$, we randomly select another image $I_{rnd}$, facilitating implicit disentanglement of attributes and identity.
    The \textit{blue bold arrows} present feature flows excluding the target facial attributes. 
    In this configuration, AUEditNet guarantees that these flows remain outside the embedding space of the target facial attributes. 
    } 
    \label{fig:network}
\end{center} 
\end{figure*}

\subsection{Problem Setting}
\label{subsec:problem_setting}
Our objective is to develop an intermediate module within the pretrained GAN inversion pipeline that enables the modification of specific facial attributes in input face images based on target conditions, while preserving the individual's identity and leaving other attributes unaffected.
A crucial aspect of achieving this lies in effectively disentangling the target facial attributes from others and from identity. 
Prior works \cite{harkonen2020ganspace, shen2020interfacegan} focused on identifying global editing directions in latent space for desired facial attributes by analyzing data from thousands of subjects. 
The data includes a significantly larger number of subjects than the number of facial attributes aiming to edit. 
Consequently, it is common for different subjects in the dataset to share the same facial attributes. 
This characteristic naturally facilitates the disentanglement of identity-related influences from the identified global editing directions for the corresponding attribute editing. 

However, while data availability from numerous subjects is abundant, obtaining fine-grained labels poses challenges. 
The significant tradeoff between data collection and annotation efforts, particularly when expert annotation is necessary, can hinder the inclusion of detailed labels.
In specific face image editing tasks, such as AU intensity manipulation, multi-level intensity labels offer advantages over binary labels (activated or not). 
Yet, datasets with intensity labels covers more AUs often comprise fewer subjects. 
The limited subject pool in facial attribute editing may blend identity features with the target attributes, complicating disentanglement processes.
To address this, we propose a novel framework named \textit{AUEditNet}. 
This architecture enables seamless intensity adjustments across $12$ AUs in face images, even when trained on a restricted dataset containing only $18$ subjects.


\subsection{Pipeline Overview}
\label{subsec:pipeline_overview}
Consistent with previous works \cite{dalva2022vecgan, jin2023redirtrans, lyu2023deltaedit}, we use a GAN inversion pair that consists of an encoder $\mathcal{E}$ and a generator $\mathcal{G}$ to achieve the transformation between the image space and the latent space. 
All editing occurs in the latent space. 
During training, we use a pair of images $I_{src}$, $I_{tar}$ from one subject, while an additional face image $I_{rnd}$ is randomly chosen from another subject's data. 
The processes are visually depicted in Fig. \ref{fig:network} with detailed descriptions.

\vspace{-1em}
\paragraph{Feature Space for Target AUs.} 
Initially, we encode the input source image $I_{src}$ into the latent vectors $W_{src}=\mathcal{E}(I_{src})$. 
To achieve explicit control over facial attributes using conditions associated with physical interpretation, we perform additional encoding of the latent vector $W^j_{src}$, one level from the multi-level vectors $W_{src}$, through a trainable encoder $\Phi^j_{enc}$. 
Here, $j$ corresponds to the level index, taking into account the disentangled level-wise structure of $W_{src}$, as outlined in Sec. \ref{subsec:architecture}. 
For the purpose of this subsection, we can disregard this index. 
The outcome of this encoding process is as follows:  
\begin{equation}
\label{eq:1}
    \Phi^j_{enc}(W^j_{src}) = \left\{\hat{c}^{i,j}_{src}, \hat{a}^{i,j}_{src}, z^j_{src} \right\}, i \in [1, N],  
\end{equation}
where $N$ represents the number of facial attributes included in the editing task. 
In this Eq. \ref{eq:1}, $\hat{c}^{i,j}_{src}$ denotes whether the $i$-th facial attribute exists or not (AU is activated or not); $\hat{a}^{i,j}_{src}$ is the corresponding estimated detailed labels (AU intensities); $z^j_{src}$ is the embedding which acts as a medium for delivering information pertaining to the target facial attributes in this newly encoded space. 
$\hat{c}^{i,j}_{src}$ would select an editing direction from a globally trainable matrix $\mathbf{T}$ if the $i$-th facial attribute exists.
$\mathbf{T}$ contains $N$ editing directions, each possessing the same dimension as the embeddings. 
When a specific editing direction $\mathbf{T}(\hat{c}^{i,j}_{src})$ is chosen, we scale it with the estimated labels $\hat{a}^{i,j}_{src}$ to serve as an intensity control. 
This yields a normalized embedding $z^j_N = z^j_{src} - \sum_{i=1}^N \hat{a}^{i,j}_{src} \cdot \mathbf{T}(\hat{c}^{i,j}_{src})$. 
Ideally, $z^j_N$ exclusively represents a canonical status of the target facial attribute, free from any person-specific information. 
While it seems feasible to continue incorporating new target conditions into this normalized embedding for subsequent generation with edited attributes \cite{zheng2020self, jin2023redirtrans}, 
this approach has limitations.
\begin{itemize}
    \item It cannot ensure the \uline{complete exclusion of other attributes or identity features} from the normalized embedding. 
    \item Achieving optimal disentanglement of identity from target attributes requires training data that ideally \uline{encompasses as many subjects as possible} to attain the desired normalized embedding.
    \item A loss function is necessary to enforce normalized embeddings identical within a batch, which \uline{heavily relies on the batch size} and can be resource-intensive. 
\end{itemize}
Given these limitations, instead of directly adding target conditions to the source embedding, our approach adopts a dual-branch structure to physically prevent irrelevant attribute or identity features (indicated by \textit{blue bold arrows}) from infiltrating the feature space of target facial attributes (highlighted in \textit{yellow}), as illustrated in Fig. \ref{fig:network}. 

We introduce $I_{rnd}$ through the same processing steps with the shared-weights modules and build a normalized embedding instead of using the source one to compel the network to retain only the target-attribute related information within this encoded space during training. 
Finally, we introduce new conditions (the existence of the $i$-th attribute $c^{i,j}_{tar}$ and the corresponding detailed labels $a^{i,j}_{tar}$). 
This yields the edited embedding $\hat{z}^j_{tar} = z^j_{rnd} - \sum_{i=1}^N \hat{a}^{i,j}_{rnd} \cdot \mathbf{T}(\hat{c}^{i,j}_{rnd}) + \sum_{i=1}^N a^{i,j}_{tar} \cdot \mathbf{T}(c^{i,j}_{tar})$. 
During testing, we directly use source normalized embedding for efficiency considerations. 

\vspace{-2em}
\paragraph{Source Latent Vectors Editing.} For all other facial attributes and identity information, our goal is to preserve them within the original latent space \cite{jin2023redirtrans}. 
We input the source embedding $z^j_{src}$ and the edited target embedding $\hat{z}^j_{tar}$ into the decoder $\Phi^j_{dec}$ to obtain the residuals $\Delta W^j_{src}$ and $\Delta \hat{W}^j_{tar}$ respectively, which are used for editing $W^j_{src}$.
The purpose of $\Delta W^j_{src}$ is to capture the source status of the target facial attributes in the input image, while $\Delta \hat{W}^j_{tar}$ stores the new status. 
Rather than solely assessing the result with the new status, we propose to supervise both outcomes through the following expressions: 
\begin{equation}
\label{eq:2}
\left\{
    \begin{aligned}
        \hat W^j_N &= W^j_{src} - \Delta W^j_{src}, \\
        \hat W^j_{tar} &= \hat W^j_N + \Delta \hat{W}^j_{tar}, \\
    \end{aligned}
\right.
\end{equation}
where $\hat W^j_N$ represents the intermediate editing resulting from the removal of the source status of the aimed facial attributes, and $\hat W^j_{tar}$ is the outcome achieved by incorporating the target conditions based on $\hat W^j_N$. 
After replacing the latent vector at the index $j$ in $W_{src}$ with $\hat W^j_N$ (or $\hat W^j_{tar}$), we obtain the final edited latent vectors $\hat W_N$ (or $\hat W_{tar}$) for image generation. 
$\hat I_{tar}^N = \mathcal{G}(\hat W_N)$ represents a synthesized face image with zero intensities (deactivation) for all AUs, while $\hat I_{tar} = \mathcal{G}(\hat W_{tar})$ is generated based on the target intensities. 

\subsection{Multi-level Architecture}
\label{subsec:architecture}
The latent space used for editing is the $W^+$ space \cite{abdal2019image2stylegan}, compatible with StyleGAN \cite{karras2020analyzing}. 
Latent vectors in $W^+$ exhibit a multi-level structure, allowing them to control different semantic levels of images \cite{yang2021semantic}. 
The level is indexed by $j$ and $j \in [1, M]$, where $M \leq 18$ due to the dimension of $W^+$. 
Instead of attempting to reintegrate the disentangled features by levels of the latent vectors in $W^+$ using a single editing module, we opt for multiple independent editing modules.
Each of these modules is responsible for editing a specific level of the latent vectors. 
However, as we lack information regarding the relationship between the aimed facial attributes and the level index $j$, supervising the level-wise $\hat c^{i, j}_{src}, \hat a^{i, j}_{src}$ could be challenging. 
This is particularly true for the AU intensity editing task, where it's not reasonable to expect that a single level estimated results $\hat c^{i, j}_{src}, \hat a^{i, j}_{src}$ can accurately represent $12$ AU intensities across various areas of the face. 
Incorrectly estimating the AU intensity of the input image could disrupt the normalization process and consequently affect the final manipulation accuracy. 
To address this, we draw inspiration from the concept of a latent space for images and propose the creation of another `latent space' for labels. 
We use a fully connected network $\Psi_{enc}$ to encode the labels of the target conditions $(c^i_{tar}, a^i_{tar})$ for the $i$-th facial attribute into corresponding level-wise pseudo-labels $(c^{i, j}_{tar}, a^{i, j}_{tar})$ to suit the editing needs of each level. 
Given each level's estimated $(\hat c^{i, j}_{src}, \hat a^{i, j}_{src})$, we decode them back to the label space, resulting in estimated source labels $(\hat c^{i}_{src}, \hat a^{i}_{src})$ through $\Psi_{dec}$. 
The entire process can be summarized as follows:
\begin{equation}
\label{eq:3}
\left\{
    \begin{aligned}
        & \Psi_{enc}(c^i_{tar}, a^i_{tar}) = c^{i, j}_{tar}, a^{i, j}_{tar}, \\
        & \Psi_{dec}(\hat c^{i, j}_{src}, \hat a^{i, j}_{src}) = \hat c^{i}_{src}, \hat a^{i}_{src}, \\
    \end{aligned}
\right.
\end{equation}
where $i \in [1, N]$, $j \in [1, M]$, and the subscripts `\textit{src}' and `\textit{tar}'can be interchanged if we switch the roles of the source and target images during training.

\subsection{Objectives}
\label{subsec:objectives}
During training, AUEditNet requires source and target images from the same subject. 
And the random image can be randomly picked from other subjects. 
We train AUEditNet by minimizing the following loss:
\begin{equation}
\label{eq:4}
    \mathcal{L} = \lambda_{R}\mathcal{L}_R + \lambda_{P}\mathcal{L}_P + \lambda_{F}\mathcal{L}_{F} + \lambda_{ID}\mathcal{L}_{ID} + \lambda_{L}\mathcal{L}_{L}. 
\end{equation}

\vspace{-1.5em}
\paragraph{Pixel-wise and Perceptual Losses.} We minimize both the pixel-wise loss $\mathcal{L}_R$ and the perceptual loss $\mathcal{L}_P$ \cite{johnson2016perceptual} between the edited image $\hat I_{tar}$, which is generated based on the provided target conditions, and the actual target image $I_{tar}$. 
\begin{equation}
\label{eq:5}
\begin{aligned}
    \mathcal{L}_R &= \| \hat I_{tar} - I_{tar} \|_2, \\
    \mathcal{L}_P &= \| F_{pcept}(\hat I_{tar}) - F_{pcept}(I_{tar}) \|_2, \\
\end{aligned}
\end{equation}
where $F_{pcept}(\cdot)$ denotes the perceptual feature extractor.

\vspace{-1em}
\paragraph{Pretrained Function Loss.} Following the prior works \cite{zheng2020self, jin2023redirtrans}, the pretrained function loss $\mathcal{L}_F$ focuses on task-relevant inconsistencies between $\hat I_{tar}$ and $I_{tar}$.
The inconsistencies include both intermediate activation feature maps $\{f_k, k \in [1, K]\}$ and estimation results derived from a network $F_{pre}(\cdot)$, which is pretrained on the specific task (e.g. AU intensity estimation). 
\begin{equation}
\label{eq:6}
\begin{aligned}
    \mathcal{L}_F = & \frac1K\sum_{k=1}^K \| f_k(\hat I_{tar}) - f_k(I_{tar}) \|_2 \\ 
    + &\frac{1}{N}\| F_{pre}(\hat I_{tar}) - F_{pre}(I_{tar}) \|_2,
\end{aligned}
\end{equation}
where $K$ is the number of chosen layers from $F_{pre}$. 

\vspace{-1em}
\paragraph{Identity Loss.} 
We restrict the ID similarity between the real image $I_{tar}$ and two generated images $\hat I_{tar}; \hat I^N_{tar}$ based on the pretrained ArcFace network \cite{deng2019arcface}, denoted as $F_{id}$. 
\begin{equation}
\label{eq:7}
    \mathcal{L}_{ID} = \sum^{\hat I \in \{\hat I_{tar}, \hat I^N_{tar}\} } 1 - \langle F_{id}(\hat I), F_{id}(I_{tar})\rangle
\end{equation}

\vspace{-1.5em}
\paragraph{Label Loss.} We propose the label loss to supervise the transformation process in the `latent space' for labels through $\Psi_{enc}$ and $\Psi_{dec}$ as mentioned in Sec. \ref{subsec:architecture}. 
Let's assume we use the source image's labels ($c^{i}_{src}$ and $a^{i}_{src}$) as the target conditions for the $i$-th facial attribute.
In this scenario, the generated image should be identical to the source image. 
This implies that the estimated source embedding $z^j_{src}$ should match the edited embedding $\hat z^j_{tar}$ at each level.
In other words, the removal of the source status and the addition of the target status should be entirely consistent. 
As a result, the level-wise conditions ($c^{i, j}_{src}, a^{i, j}_{src}$) in the label's latent space, encoded from the source image's labels ($c^{i}_{src}, a^{i}_{src}$), should align with the estimated conditions ($\hat c^{i, j}_{src}, \hat a^{i, j}_{src}$). 
This corresponds to the second part of Eq. \ref{eq:8}, which supervises the learning of the encoder $\Psi_{enc}$. 
Supp. provides more details.

To further ensure that the level-wise estimation retains the information about the labels of the source image, we utilize the label decoder $\Psi_{dec}$ to guarantee that the estimated results ($\hat c^{i}_{src}, \hat a^{i}_{src}$) decoded from the level-wise conditions ($\hat c^{i, j}_{src}, \hat a^{i, j}_{src}$) are consistent with the source image's labels. 
Thus, we build the loss as follows:
\begin{equation}
\label{eq:8}
\begin{aligned}
    \mathcal{L}_L = \sum^{\xi\in\{c, a\}} \frac1N \sum_{i=1}^N 
    \Bigl(
        &\|\hat \xi^{i}_{src} - \xi^{i}_{src}\|_2 \\
      + \frac1M \sum_{j=1}^M &\| \hat \xi^{i, j}_{src} - \xi^{i, j}_{src} \|_2 
    \Bigl). 
\end{aligned}
\end{equation}

In summary, the first three terms in Eq. \ref{eq:4} ensure the generated image's similarity to the target image. 
The fourth term enforces identity consistency, and the final term supervises the learning of level-wise pseudo-labels for avoiding entangled labels as conditions and improving the attribute editing performance. 
The hyperparameters $\lambda_R, \lambda_P, \lambda_F, \lambda_{ID}, \lambda_L$ enable a balanced learning from these various losses. 
\section{Experiments}
\label{sec:experiments}
\subsection{Implementation Details}
\label{subsec:implementation}
We employed e4e \cite{tov2021designing} and StyleGAN2 \cite{karras2020analyzing} as the GAN inversion pair. 
We designed a Siamese network for the external AU intensity estimation for the pretrained function loss in Eq. \ref{eq:6}. 
This network takes a pair of face images from the same subject as the input and estimates the intensity difference of AUs between these two images. 
This design reduces the impact of subject-specific facial attributes. 
During training, we used the convolutional part of VGG-16 \cite{simonyan2015very} as the backbone to build the AU intensity estimation network $F_{pre}$. 
During test, we used a separate estimator $H_{est}$, which has the same architecture with ResNet-50 \cite{he2016deep} as the backbone. 
Importantly, this estimator $H_{est}$ was never exposed to the training phase but was trained on the same training dataset. 

We trained AUEditNet using the DISFA training subset \cite{mavadati2013disfa, mavadati2012automatic}. 
DISFA comprises of $27$ subjects and provides multi-level integral intensities for $12$ AUs, offering annotations for the largest number of AUs among publicly available datasets for AU intensity estimation. 
We used $18$ subjects for training and $9$ subjects for testing, following the data split used in \cite{li2018eac, shao2021jaa}. 
To assess AUEditNet, we used the DISFA test subset to evaluate its accuracy in manipulating AU intensities while preserving other attributes. 
Furthermore, we expanded our evaluation to encompass facial expressions, beyond AUs alone, by using the BU-4DFE dataset \cite{zhang2013high}. 
Our evaluation involved tasks related to expression transfer and data augmentation for AU intensity estimation.
For further assessment of out-of-domain editing performance, we incorporated CelebA-HQ \cite{karras2017progressive} and FFHQ \cite{karras2019style}, which both are the benchmarks for the high-quality human face image datasets.

\subsection{Evaluation Criteria}
\label{subsec:criteria}
We assess the performance of AUEditNet by examining the comparison between the generated image $\hat I_{tar}$ and the target image $I_{tar}$ from four perspectives: the accuracy of intensity editing in AUs, identity preservation, image similarity, and smile expression manipulation (illustrated in Sec. \ref{subsec:quantitative}). 

\vspace{-1.5em}
\paragraph{Accuracy of AU Intensity Manipulation.} We quantify the AU intensity manipulation performance in edited images by using the external pretrained ResNet-50 based estimator $H_{est}$, which is unseen during training. 
We report the Intra-Class Correlation (specifically ICC(3,1) \cite{shrout1979intraclass}) and mean squared error (MSE), both calculated for $12$ AUs, between the estimated values $H_{est}(\hat I_{tar})$ or $H_{est}(\hat I^N_{tar})$ and their intended target values. 

\vspace{-1.5em}
\paragraph{Identity Preservation.} A well-trained image editor should consistently maintain the identity given various provided conditions. 
To assess the similarity of identity, we measure the distance of embeddings between $\hat I_{tar}$ and $I_{tar}$ to assess the similarity of identity, where the embedding is extracted by a pretrained face recognition model \cite{adam2021face}. 

\vspace{-1.5em}
\paragraph{Image Similarity.} We employ two metrics: pixel-wise mean squared error and the Learned Perceptual Image Patch Similarity (LPIPS) \cite{zhang2018unreasonable} to measure the image similarity between $\hat I_{tar}$ and $I_{tar}$.

\subsection{Qualitative Evaluation}
\label{subsec:qualitative}
\begin{figure*}
    \captionsetup[subfigure]{labelformat=empty}
    \captionsetup[subfigure]{justification=centering}
    \centering
    \begin{subfigure}[t]{0.093\linewidth}
        \begin{minipage}{1\linewidth}
        \includegraphics[width=1\linewidth]{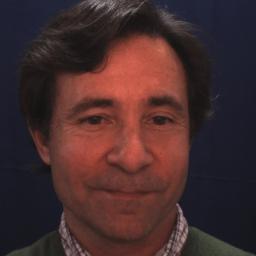}
        \includegraphics[width=1\linewidth]{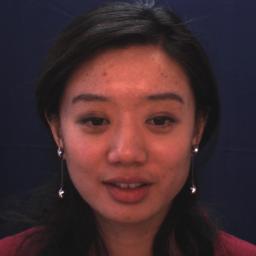}
        \includegraphics[width=1\linewidth]{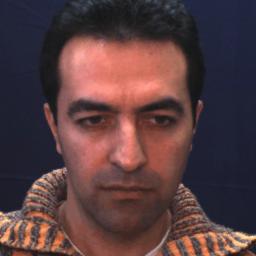}
        \includegraphics[width=1\linewidth]{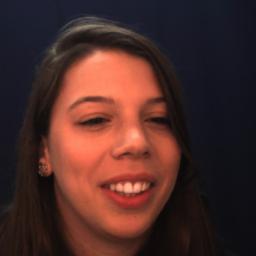}
        \end{minipage}
        \caption{Source}
  \end{subfigure}
  \hspace{-0.015\linewidth}
  \centering
    \begin{subfigure}[t]{0.093\linewidth}
        \begin{minipage}{1\linewidth}
        \includegraphics[width=1\linewidth]{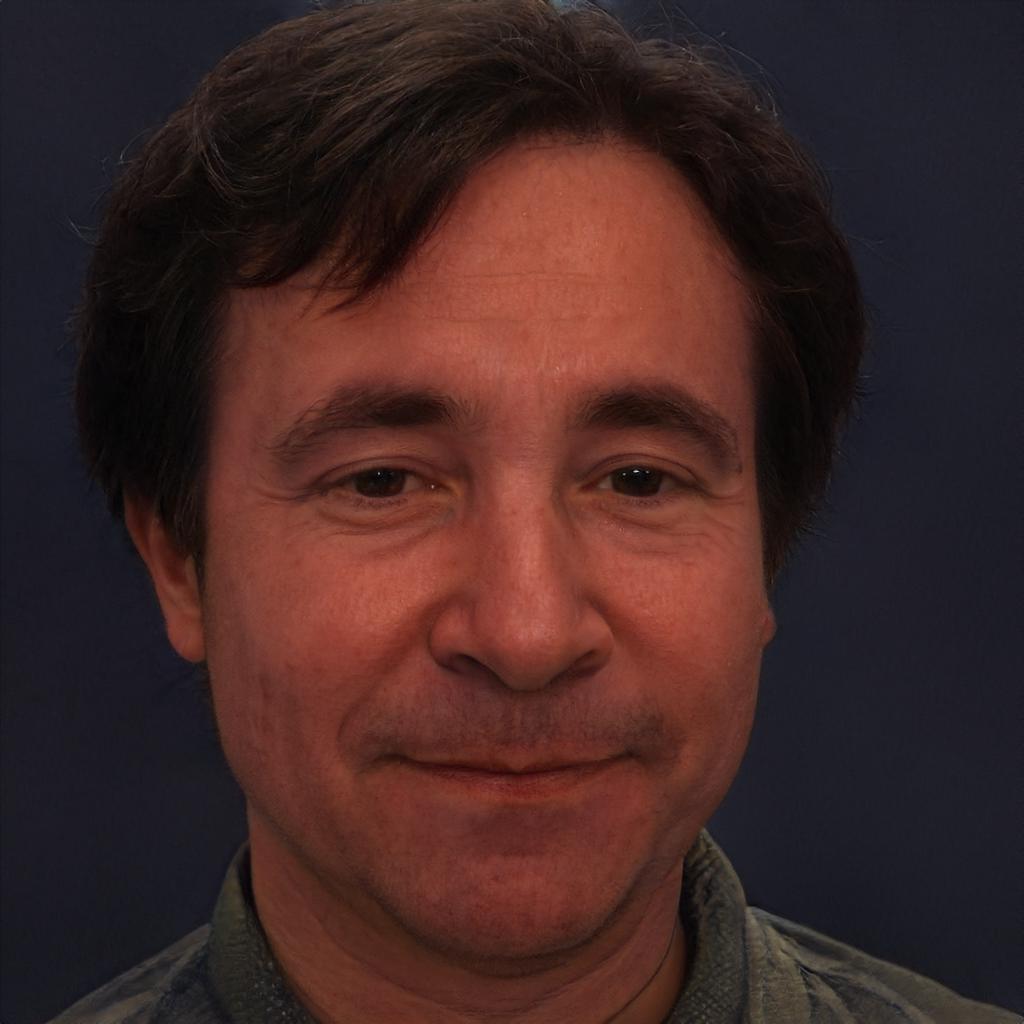}
        \includegraphics[width=1\linewidth]{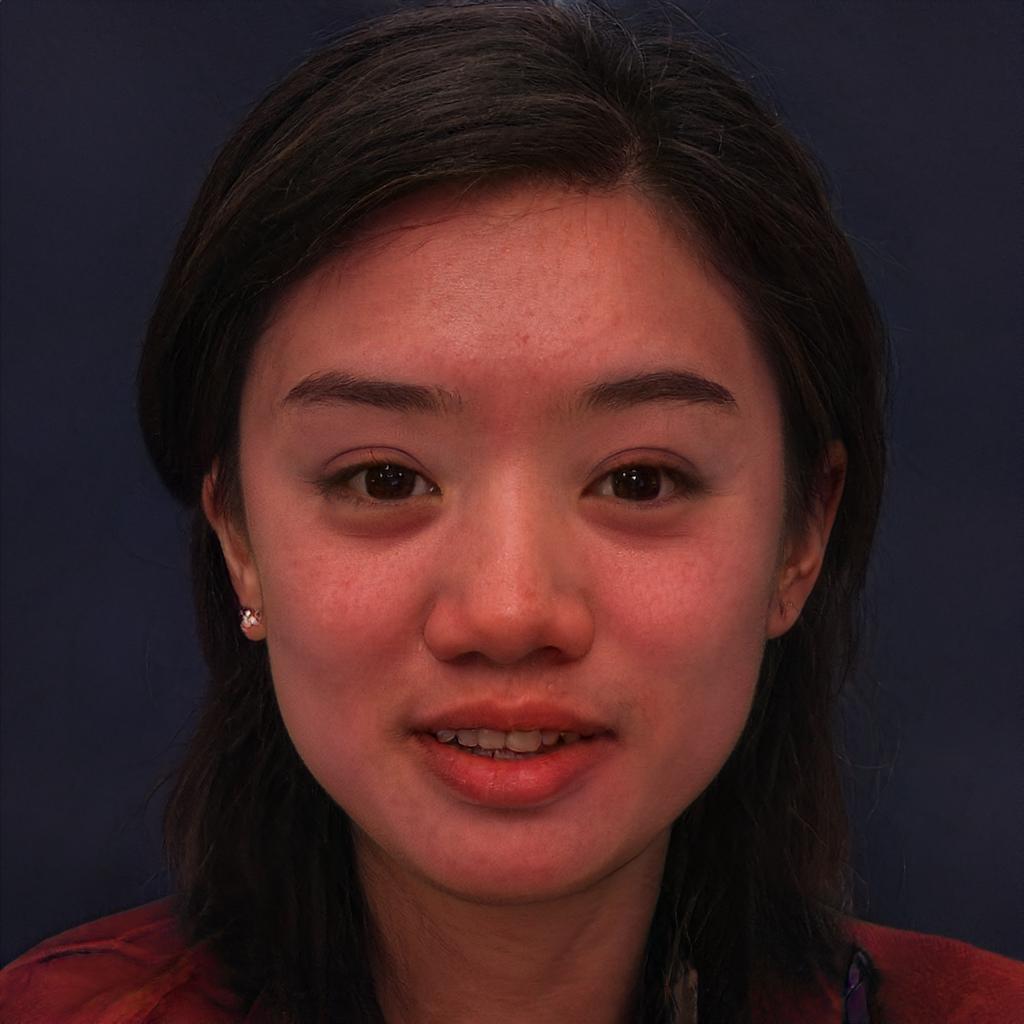}
        \includegraphics[width=1\linewidth]{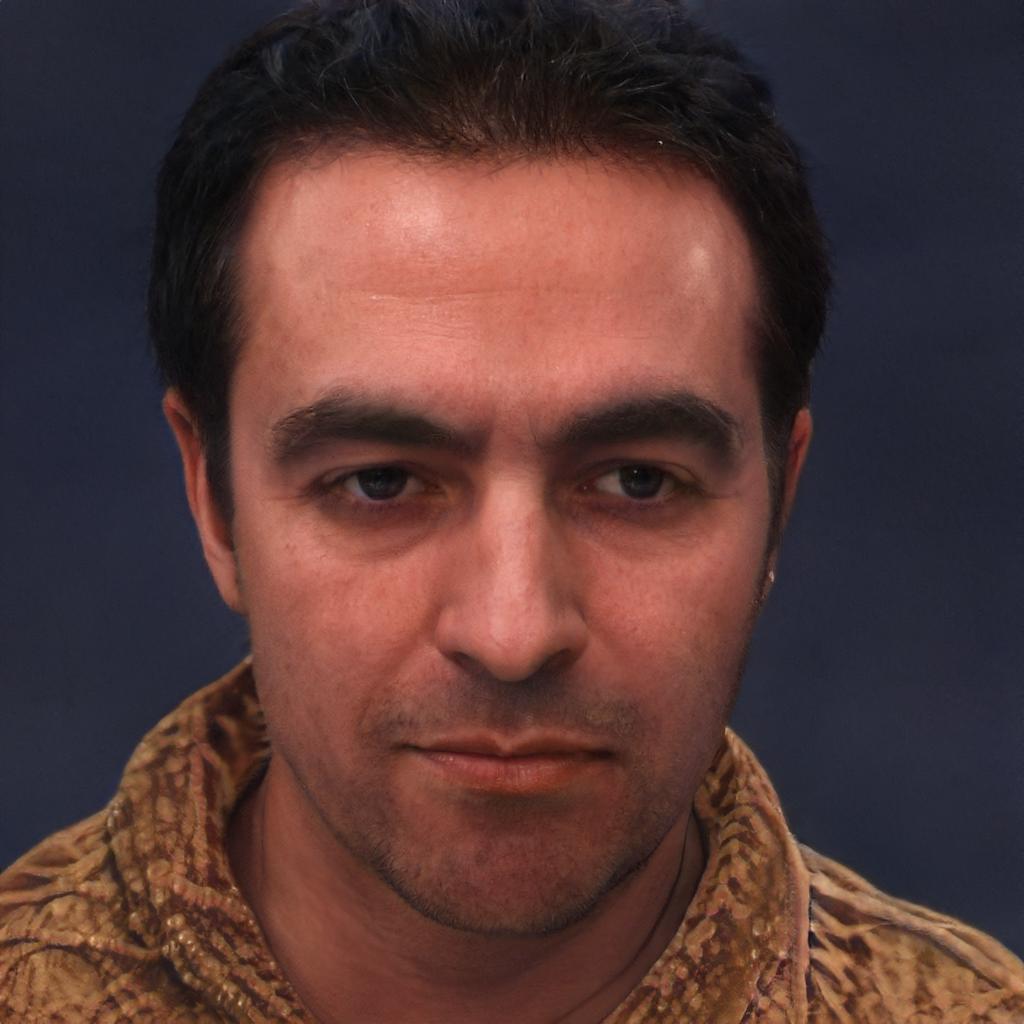}
        \includegraphics[width=1\linewidth]{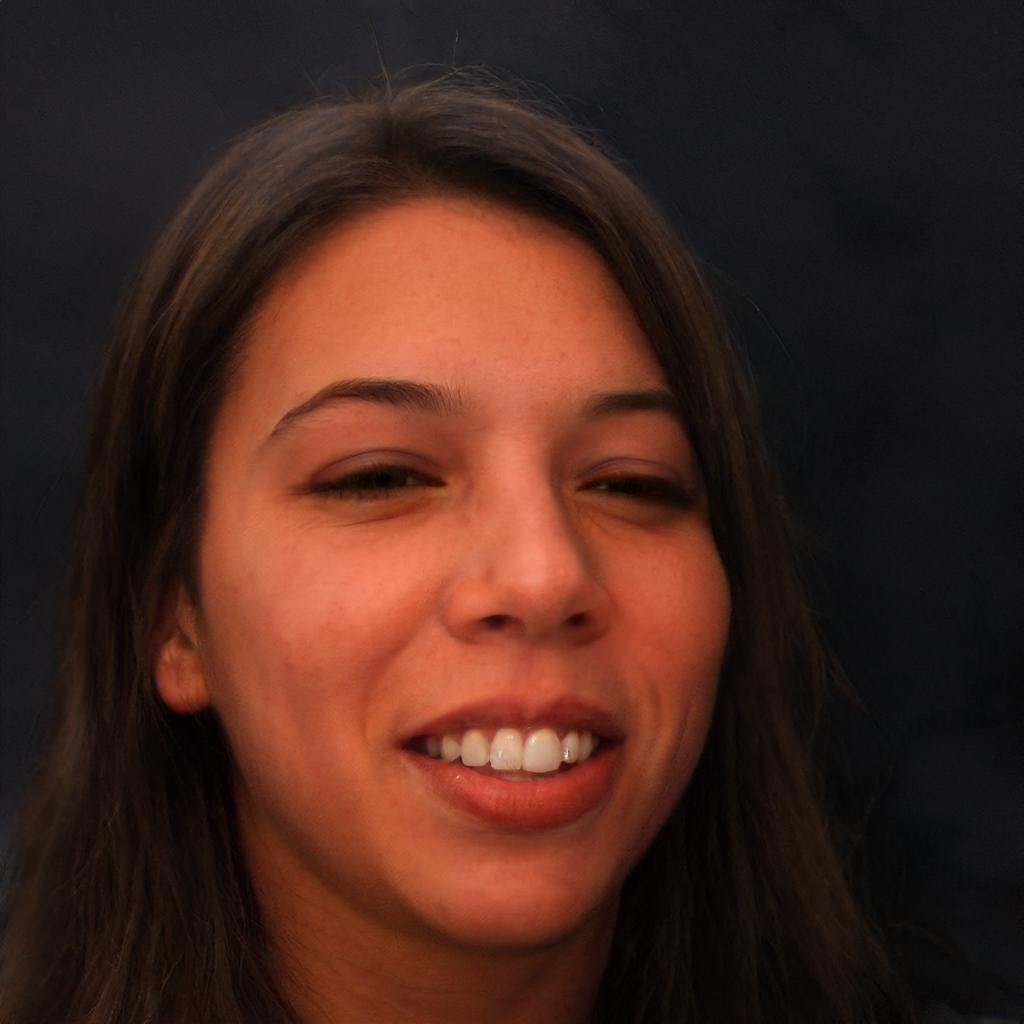}
        \end{minipage}
        \caption{e4e \\ Inversion}
  \end{subfigure}
  \centering
    \begin{subfigure}[t]{0.093\linewidth}
        \begin{minipage}{1\linewidth}
        \includegraphics[width=1\linewidth]{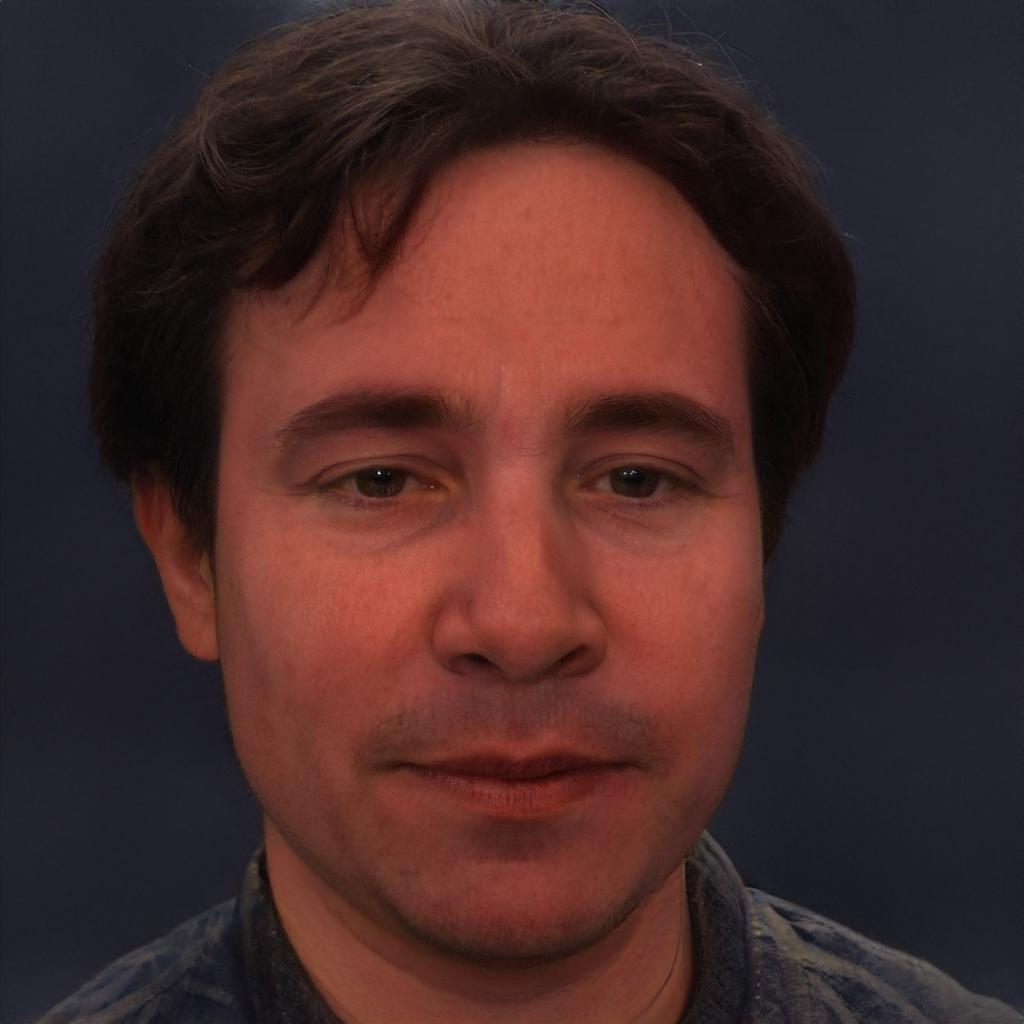}
        \includegraphics[width=1\linewidth]{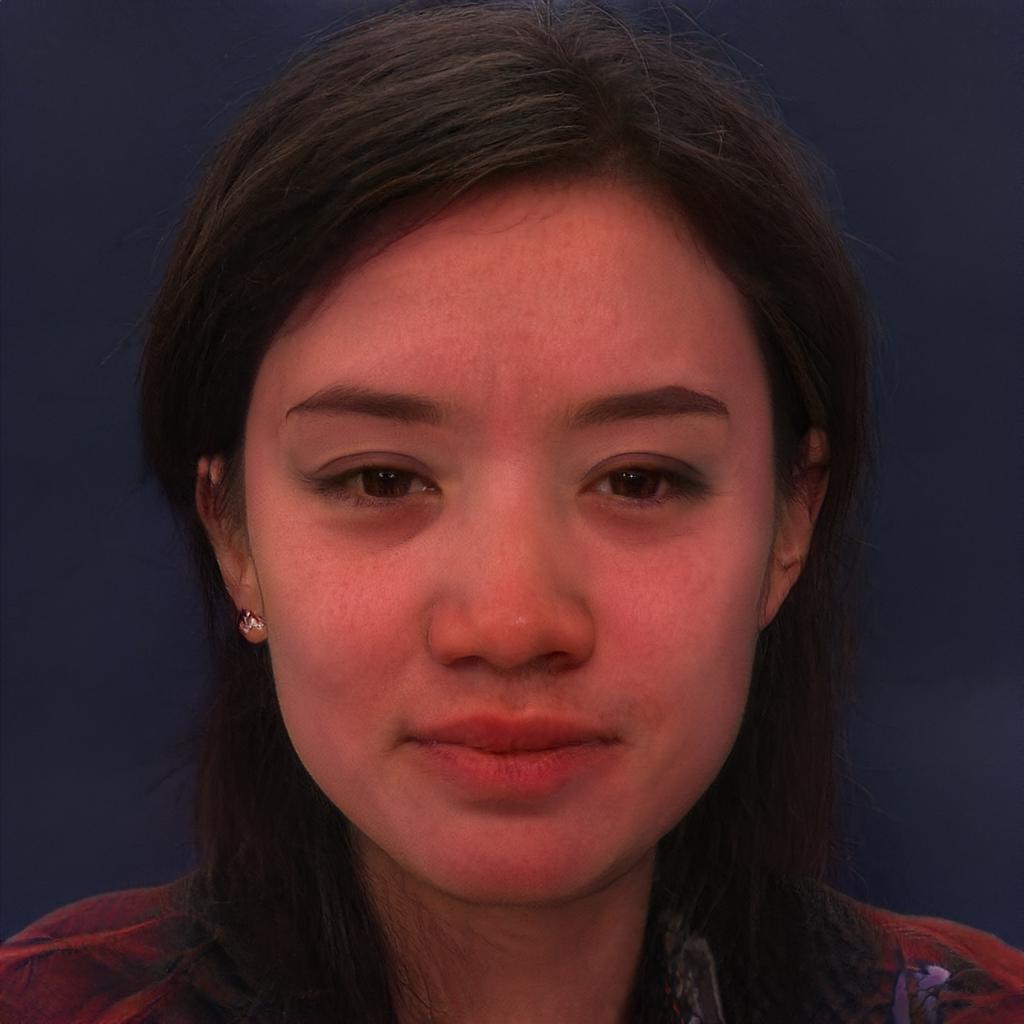}
        \includegraphics[width=1\linewidth]{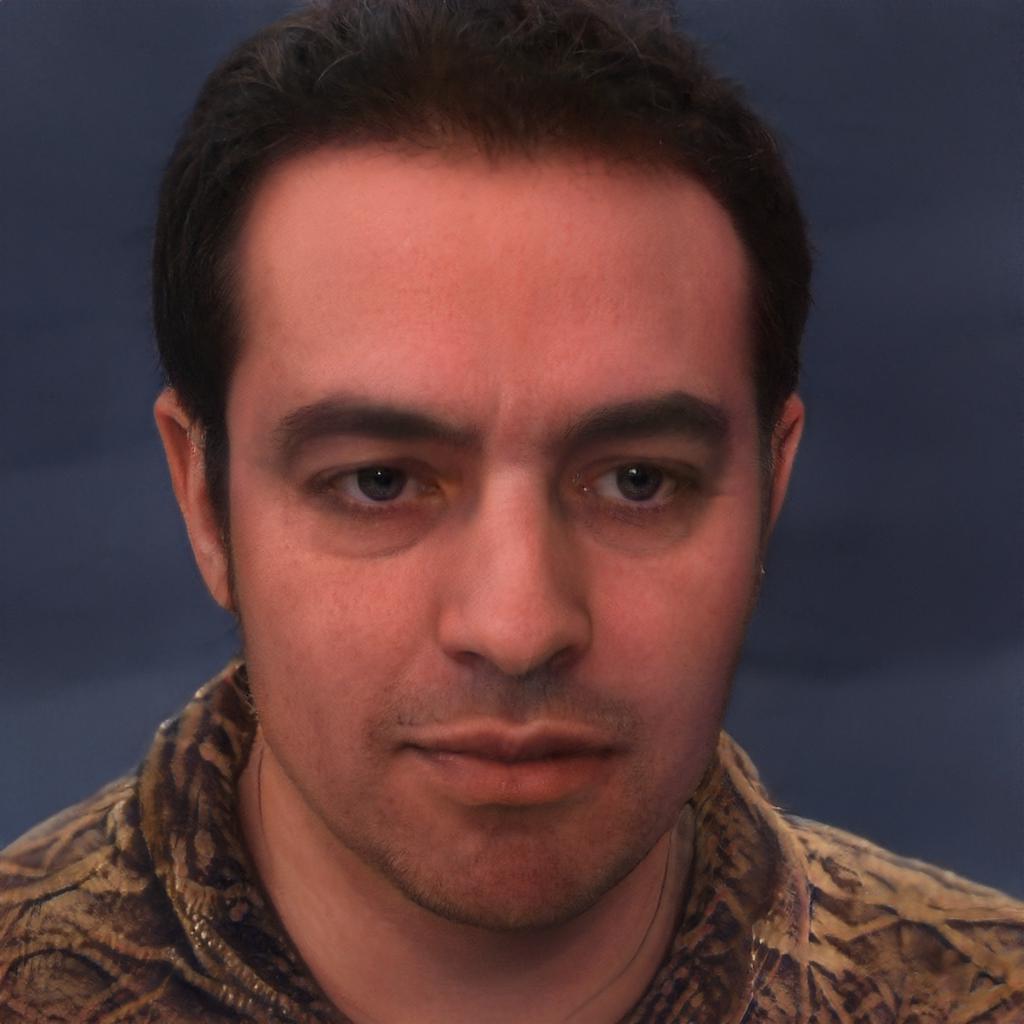}
        \includegraphics[width=1\linewidth]{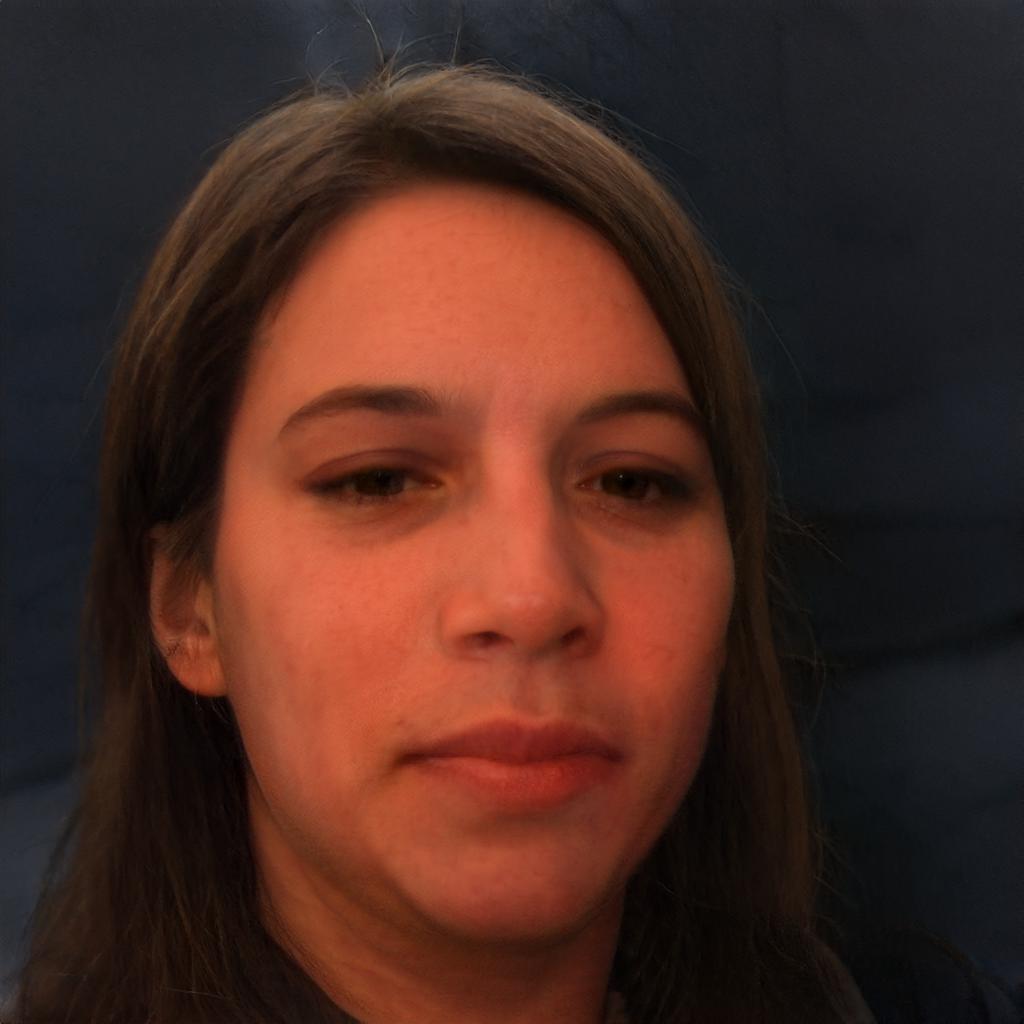}
        \end{minipage}
        \caption{AUEditNet \\ ($-$)}
  \end{subfigure}
  \hspace{-0.015\linewidth}
  \centering
    \begin{subfigure}[t]{0.093\linewidth}
        \begin{minipage}{1\linewidth}
        \includegraphics[width=1\linewidth]{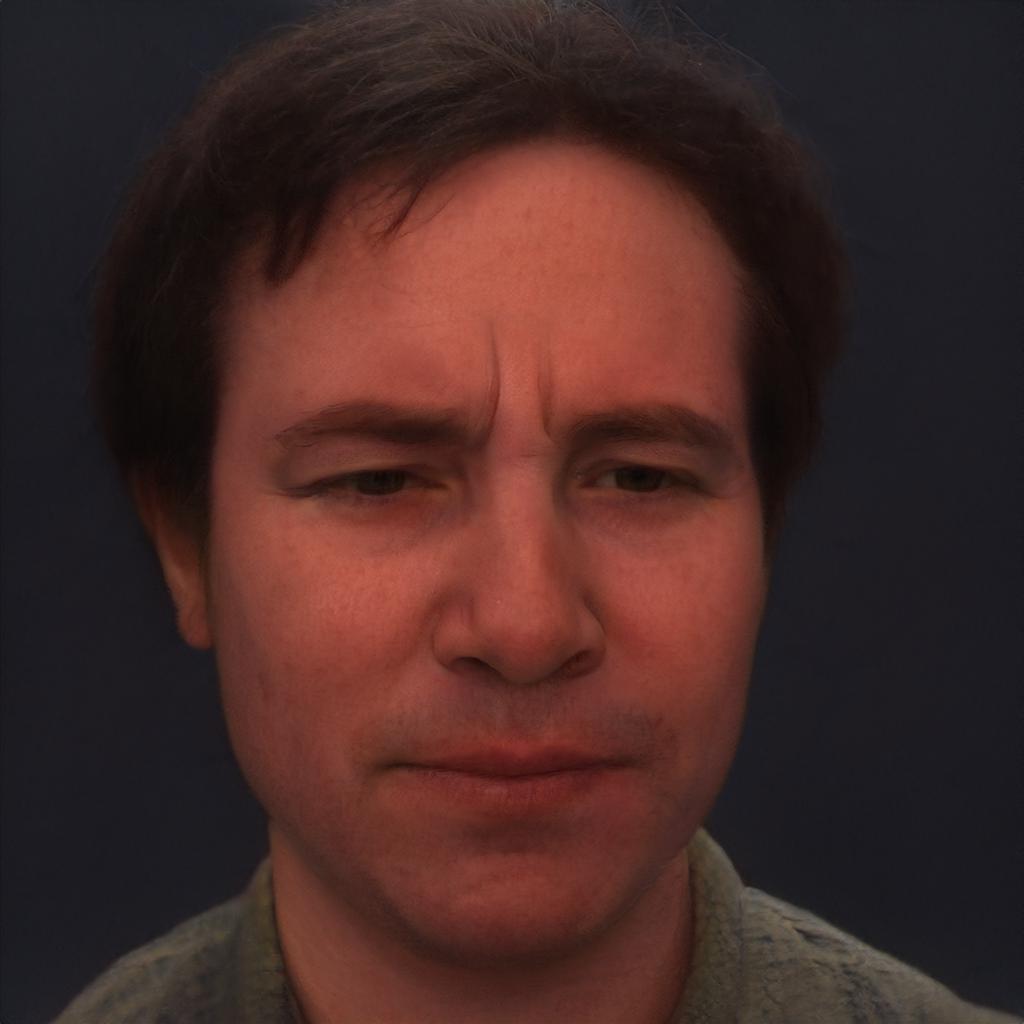}
        \includegraphics[width=1\linewidth]{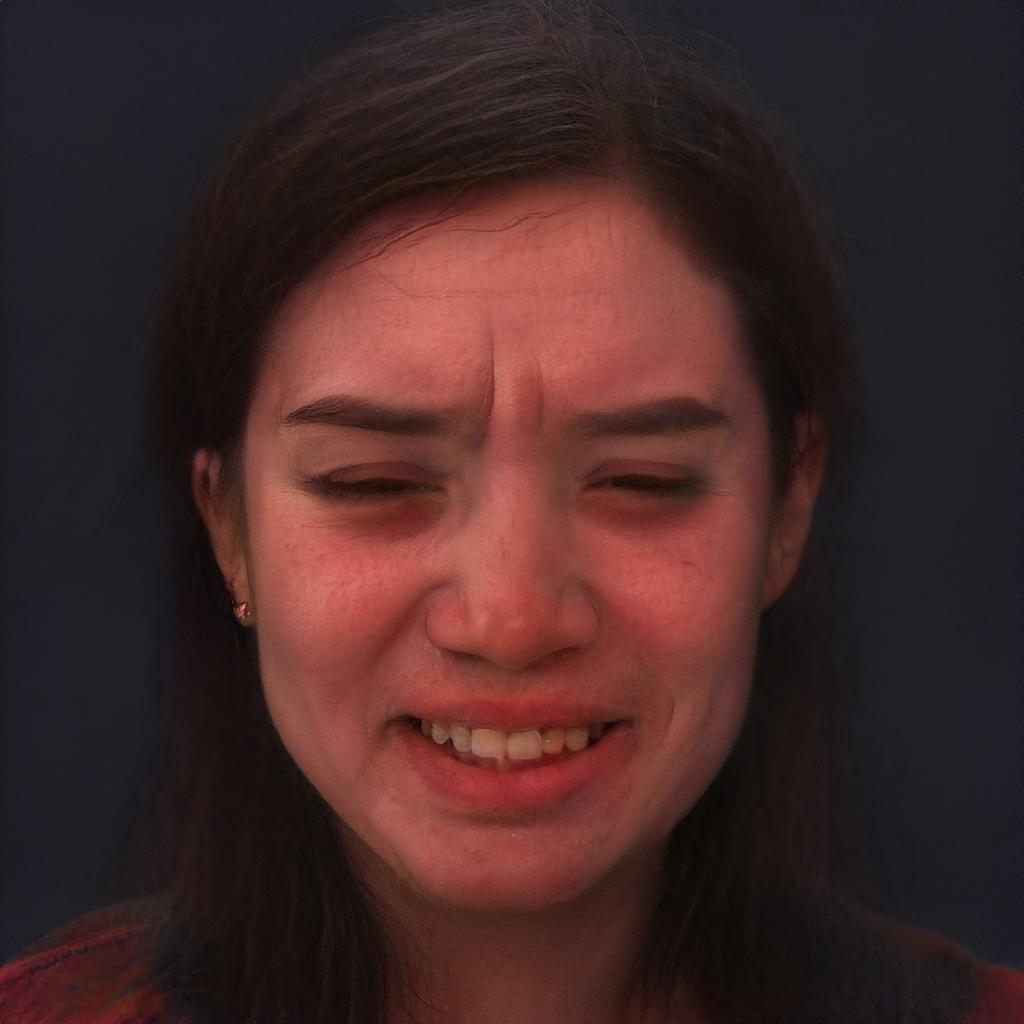}
        \includegraphics[width=1\linewidth]{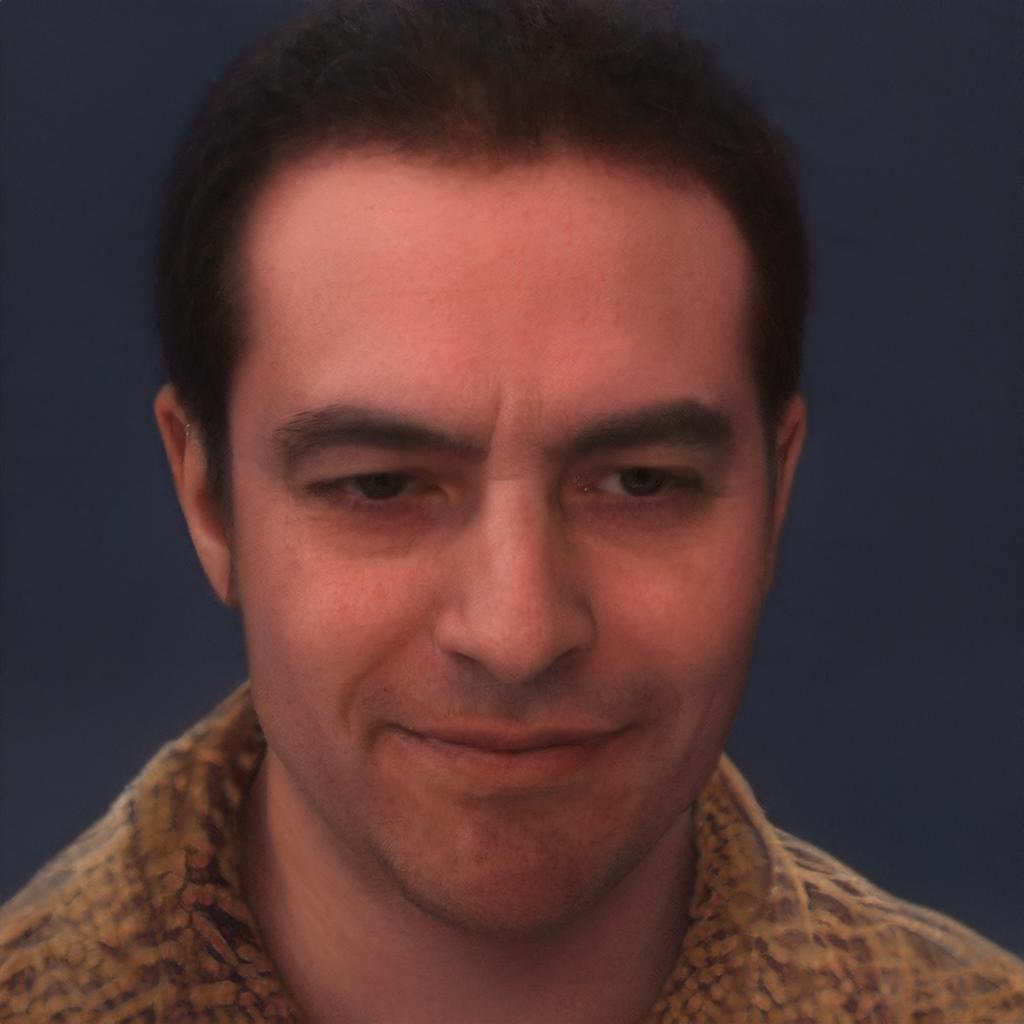}
        \includegraphics[width=1\linewidth]{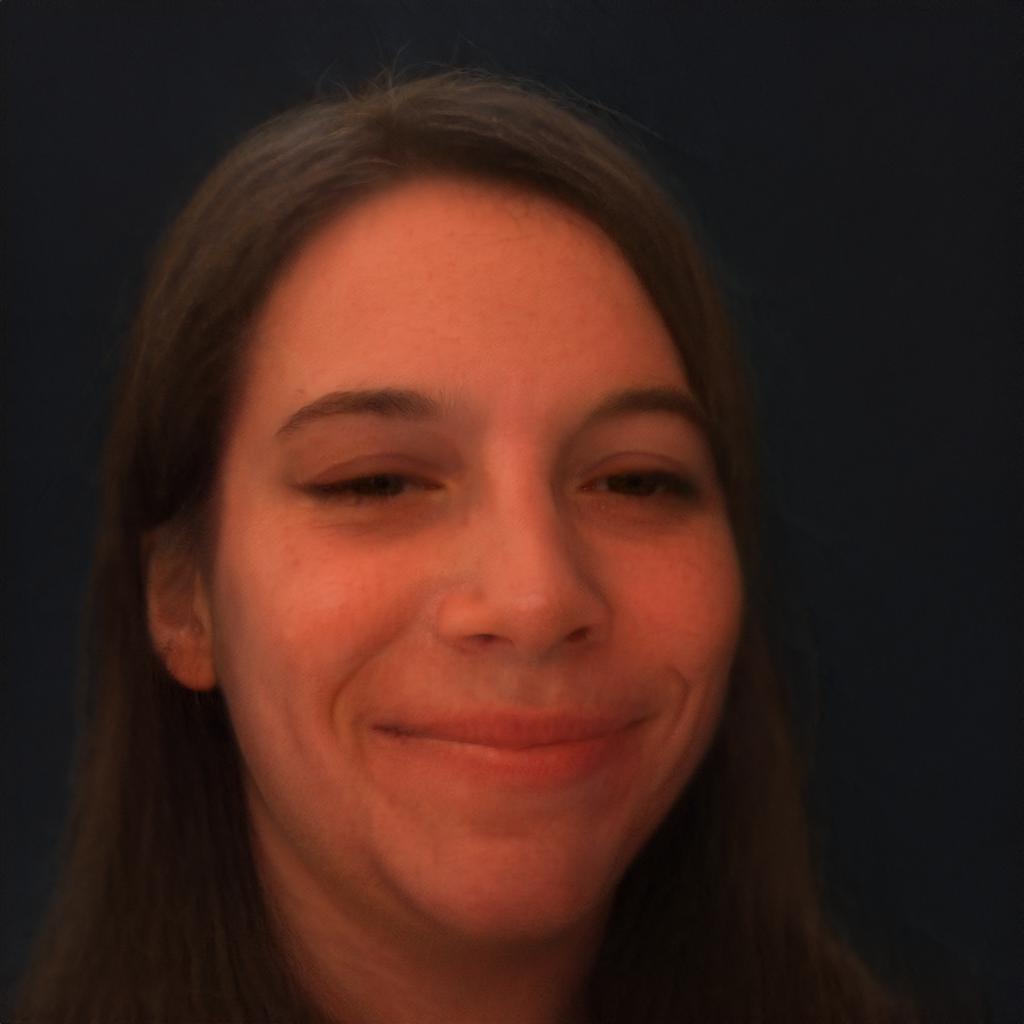}
        \end{minipage}
        \caption{AUEditNet \\ ($+$)}
  \end{subfigure}
  \centering
      \begin{subfigure}[t]{0.093\linewidth}
        \begin{minipage}{1\linewidth}
        \includegraphics[width=1\linewidth]{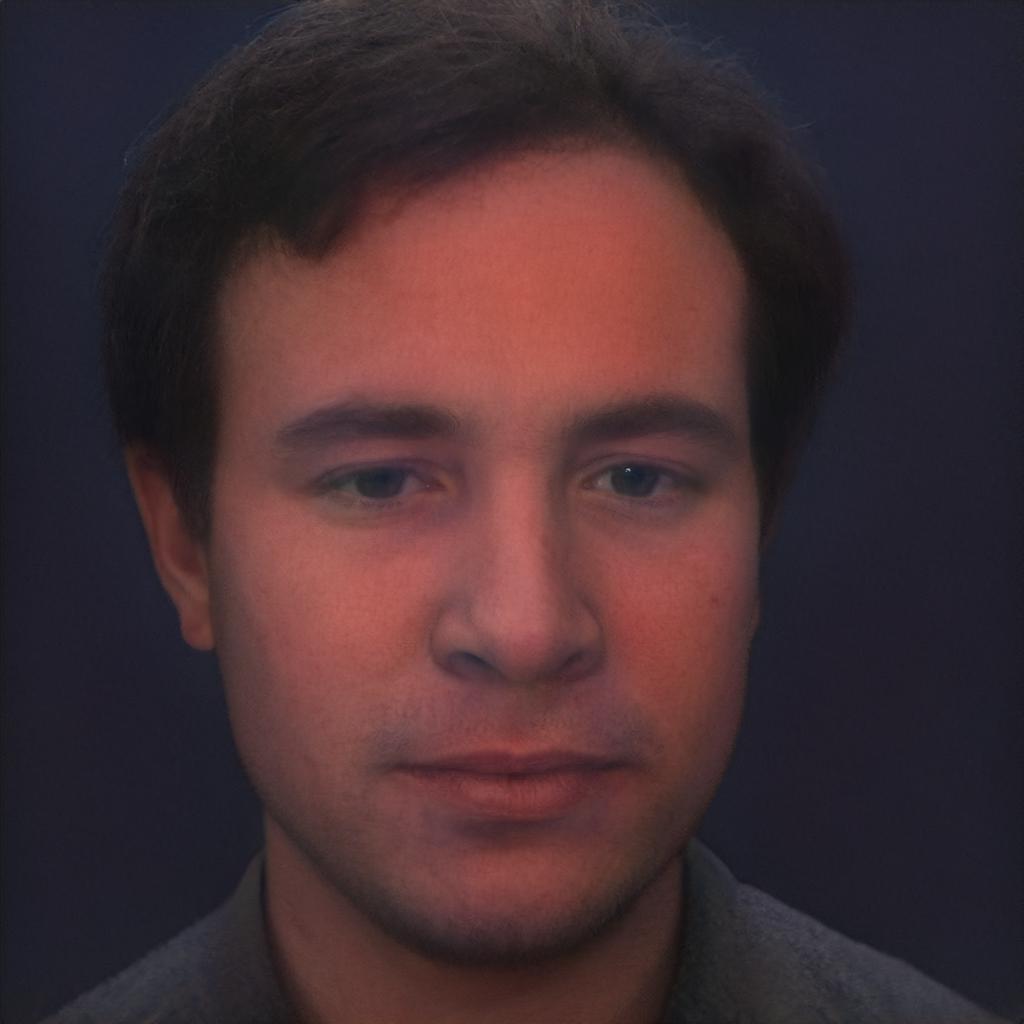}
        \includegraphics[width=1\linewidth]{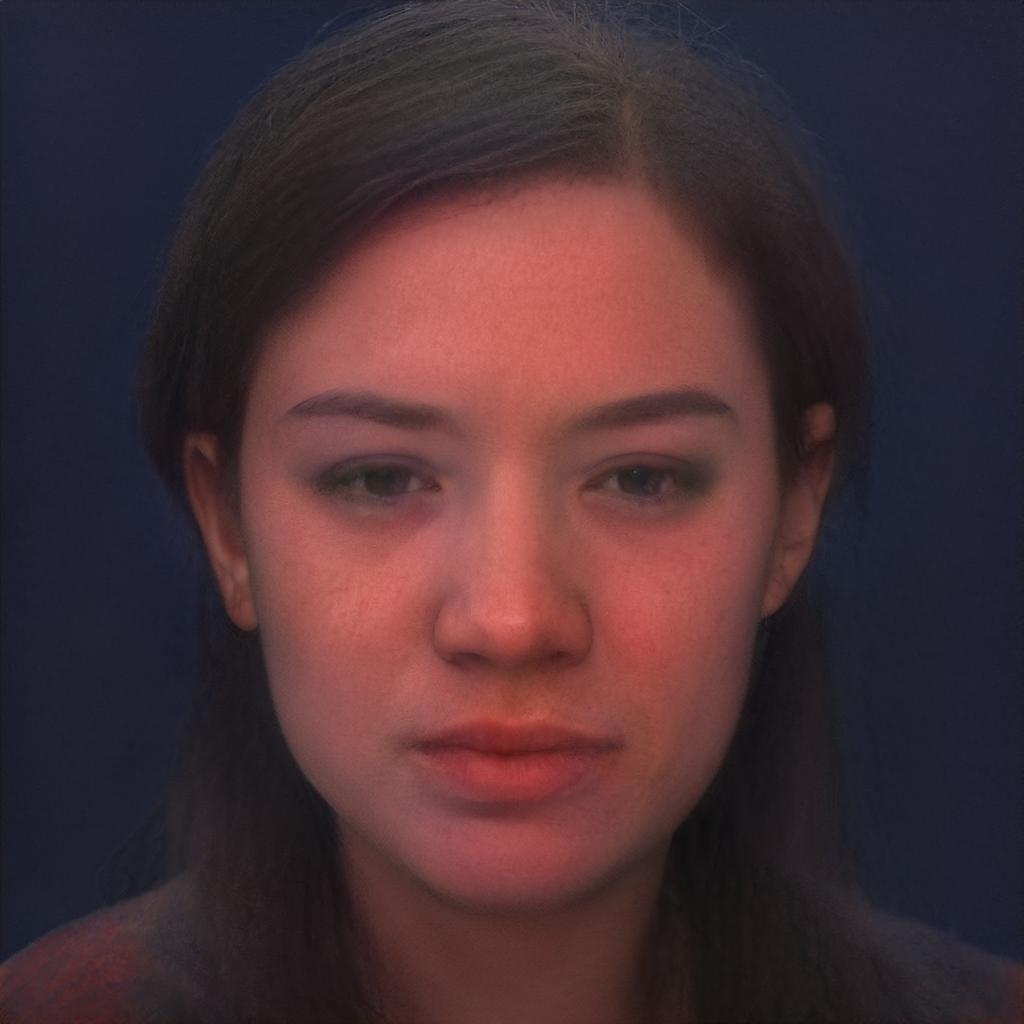}
        \includegraphics[width=1\linewidth]{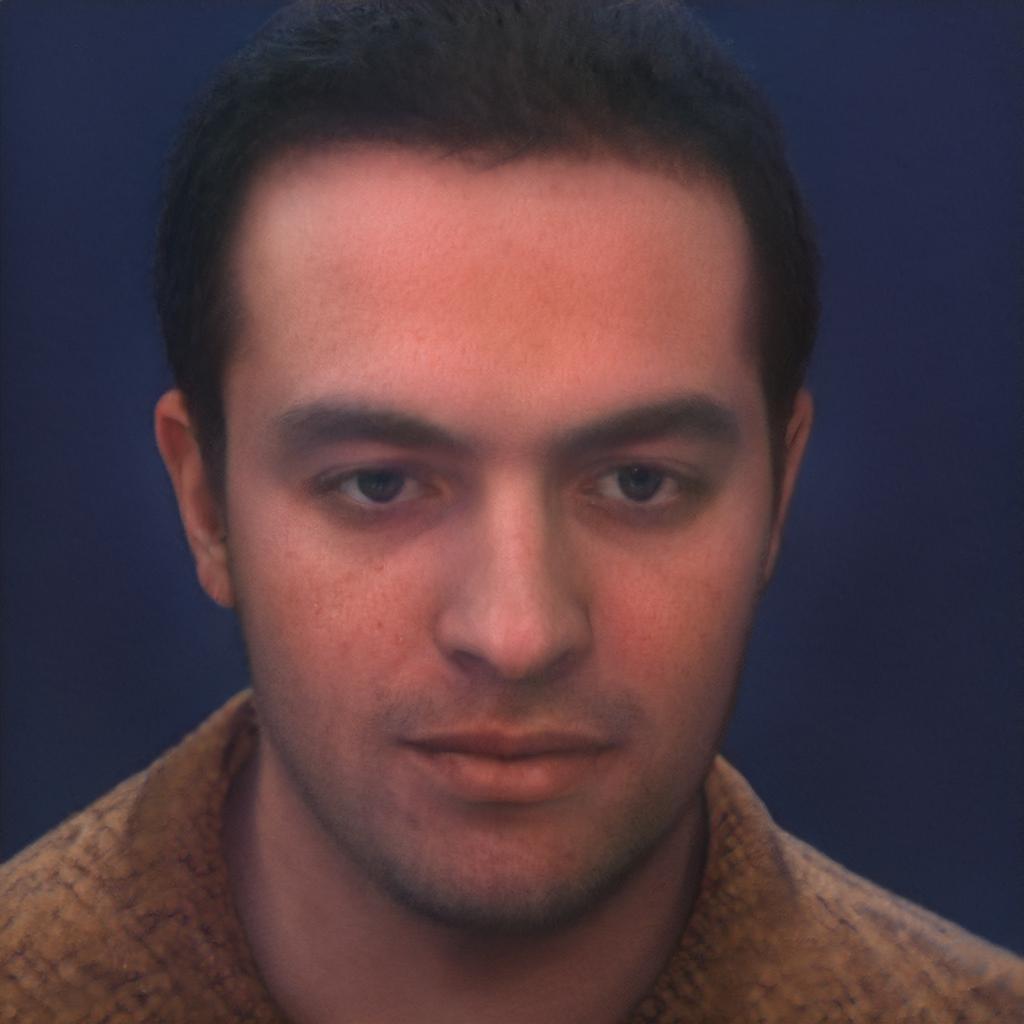}
        \includegraphics[width=1\linewidth]{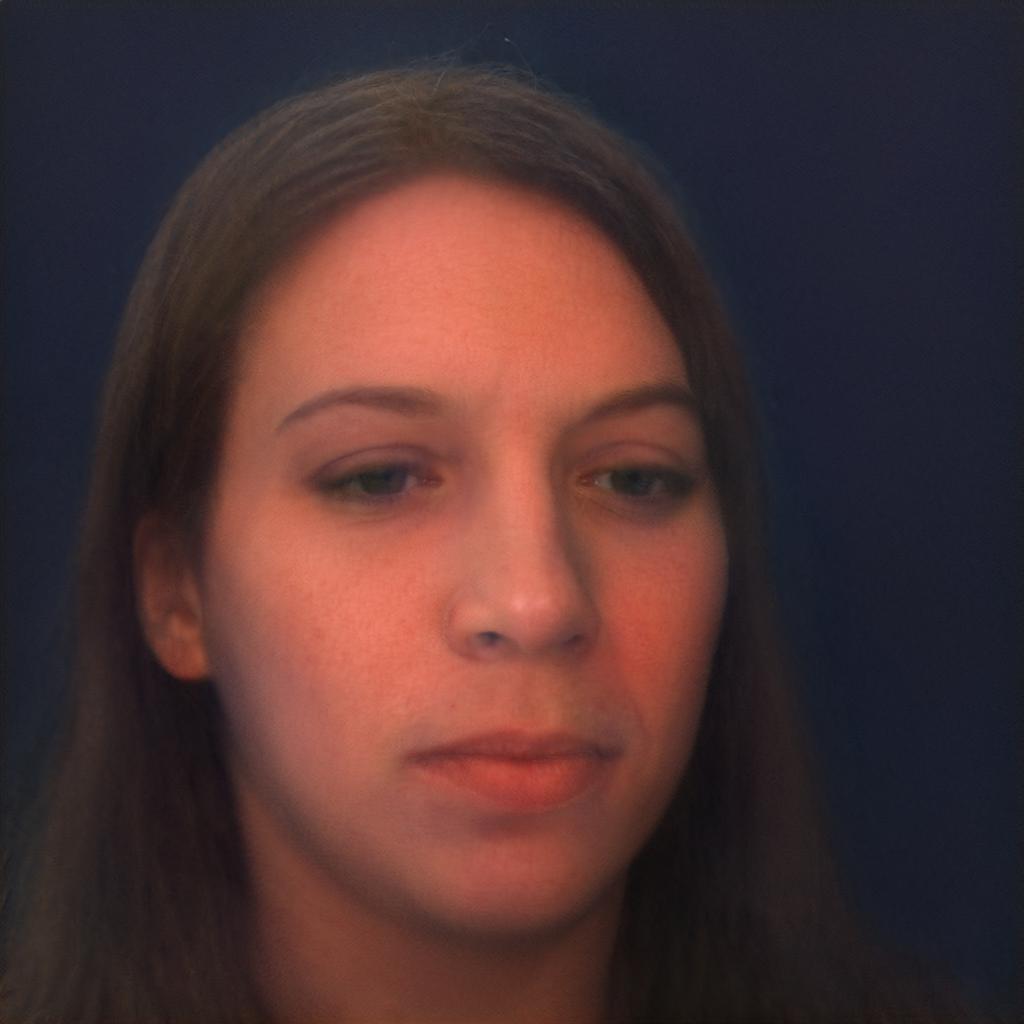}
        \end{minipage}
        \caption{ReDirTrans \\ ($-$)}
  \end{subfigure}
  \hspace{-0.015\linewidth}
  \centering
    \begin{subfigure}[t]{0.093\linewidth}
        \begin{minipage}{1\linewidth}
        \includegraphics[width=1\linewidth]{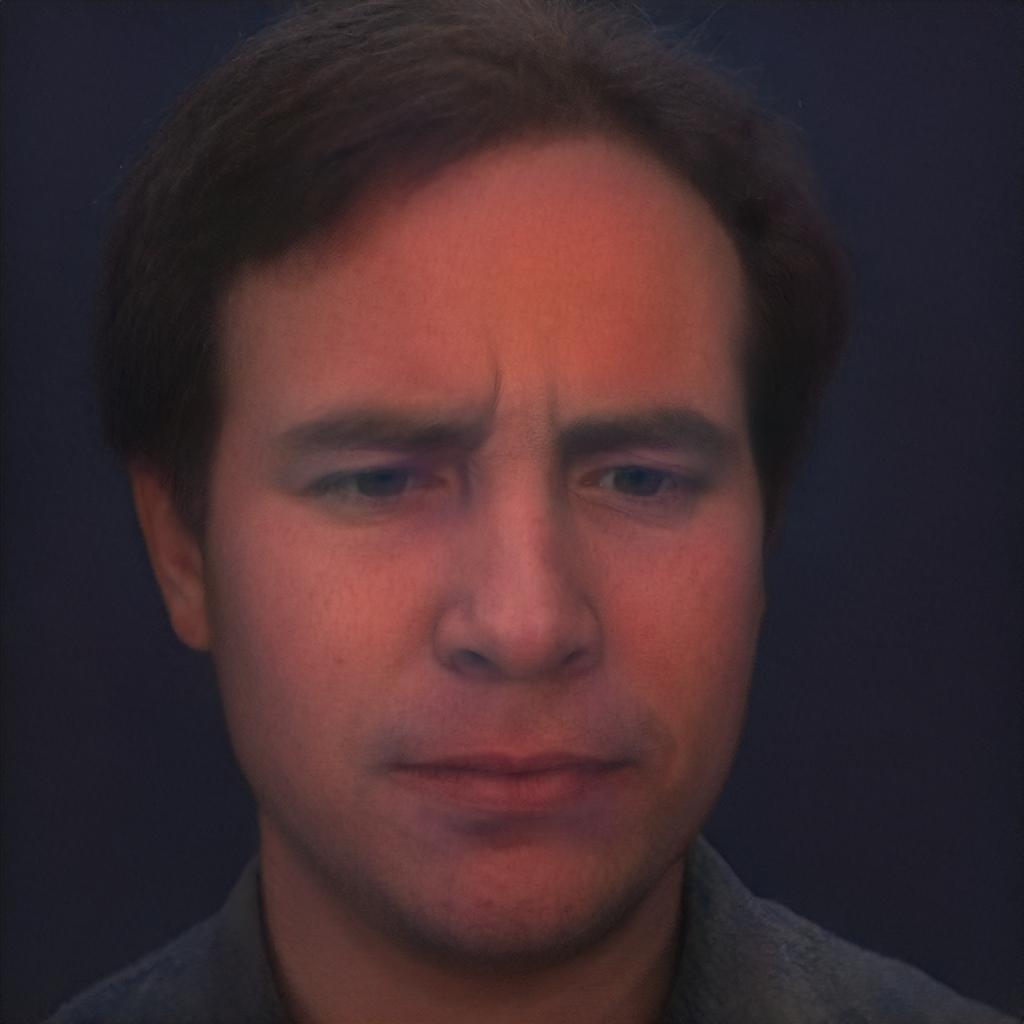}
        \includegraphics[width=1\linewidth]{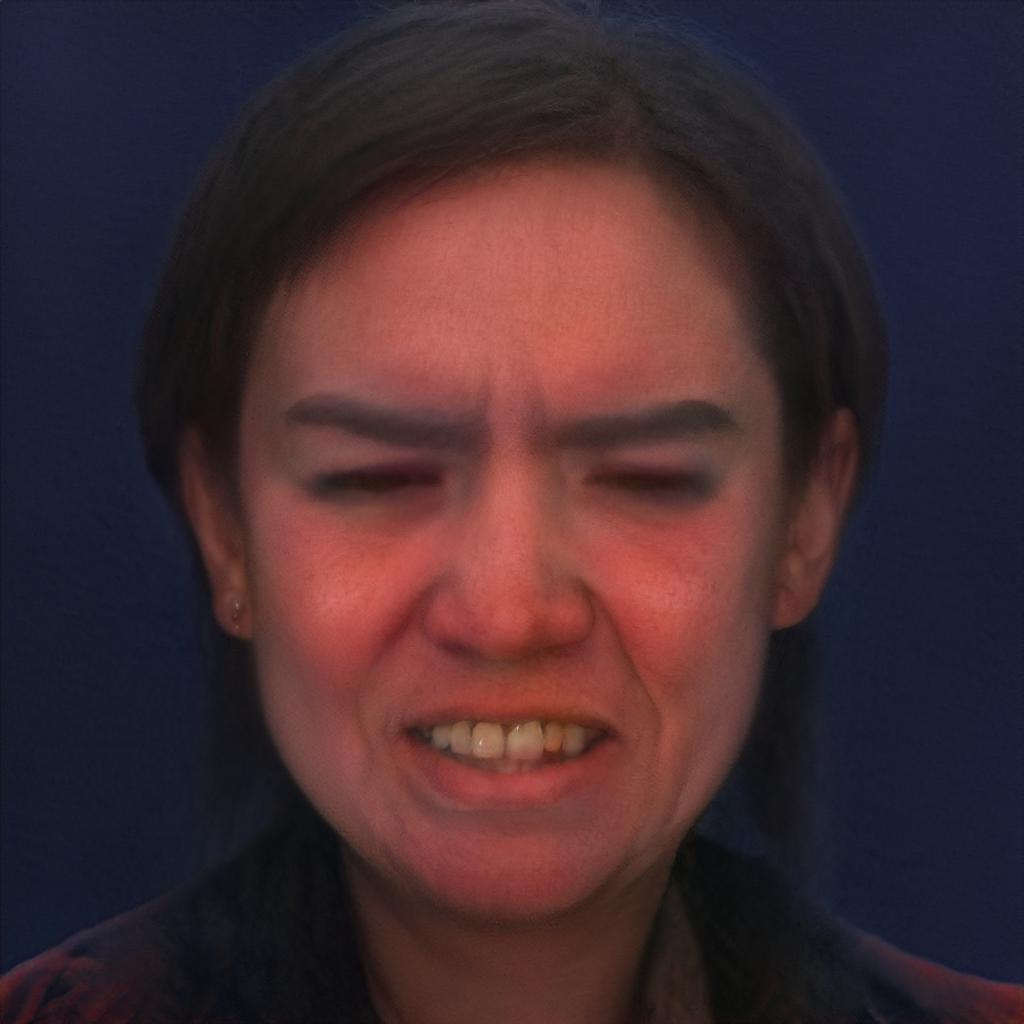}
        \includegraphics[width=1\linewidth]{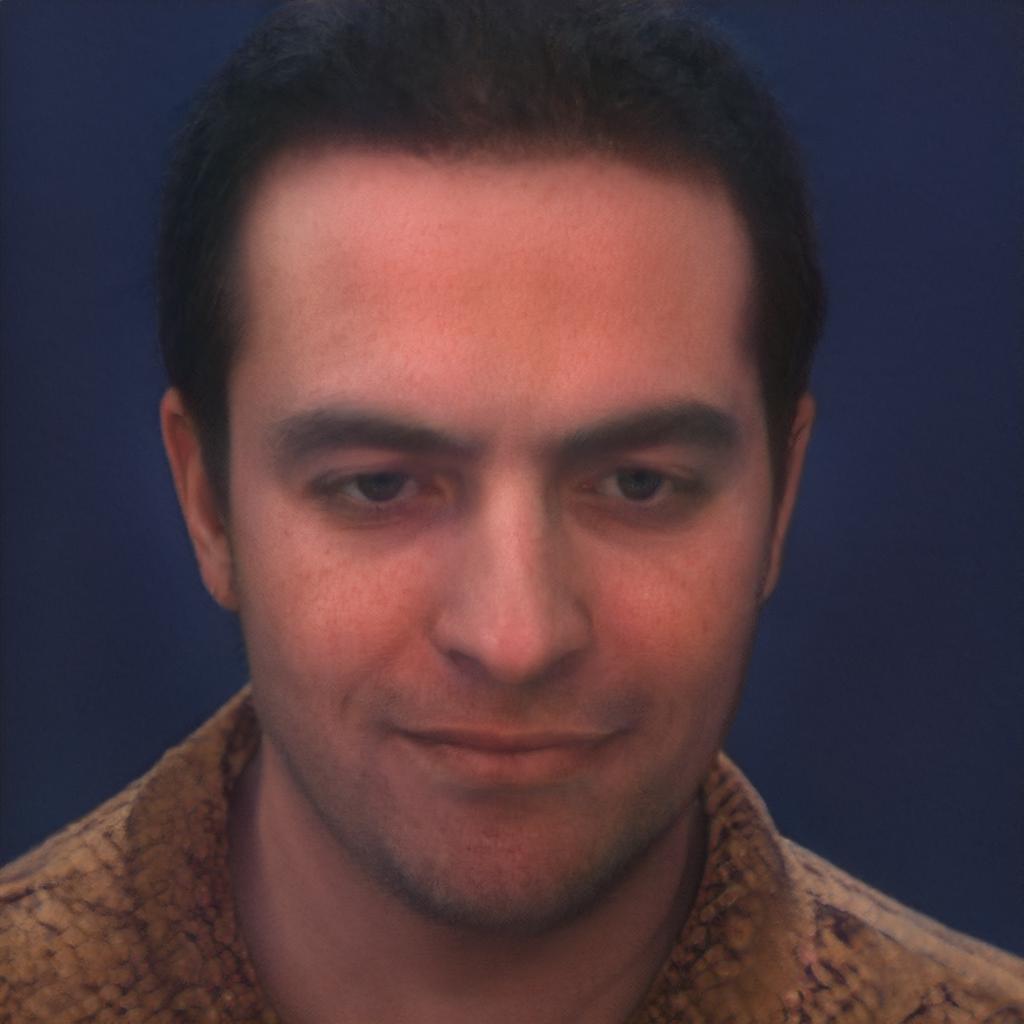}
        \includegraphics[width=1\linewidth]{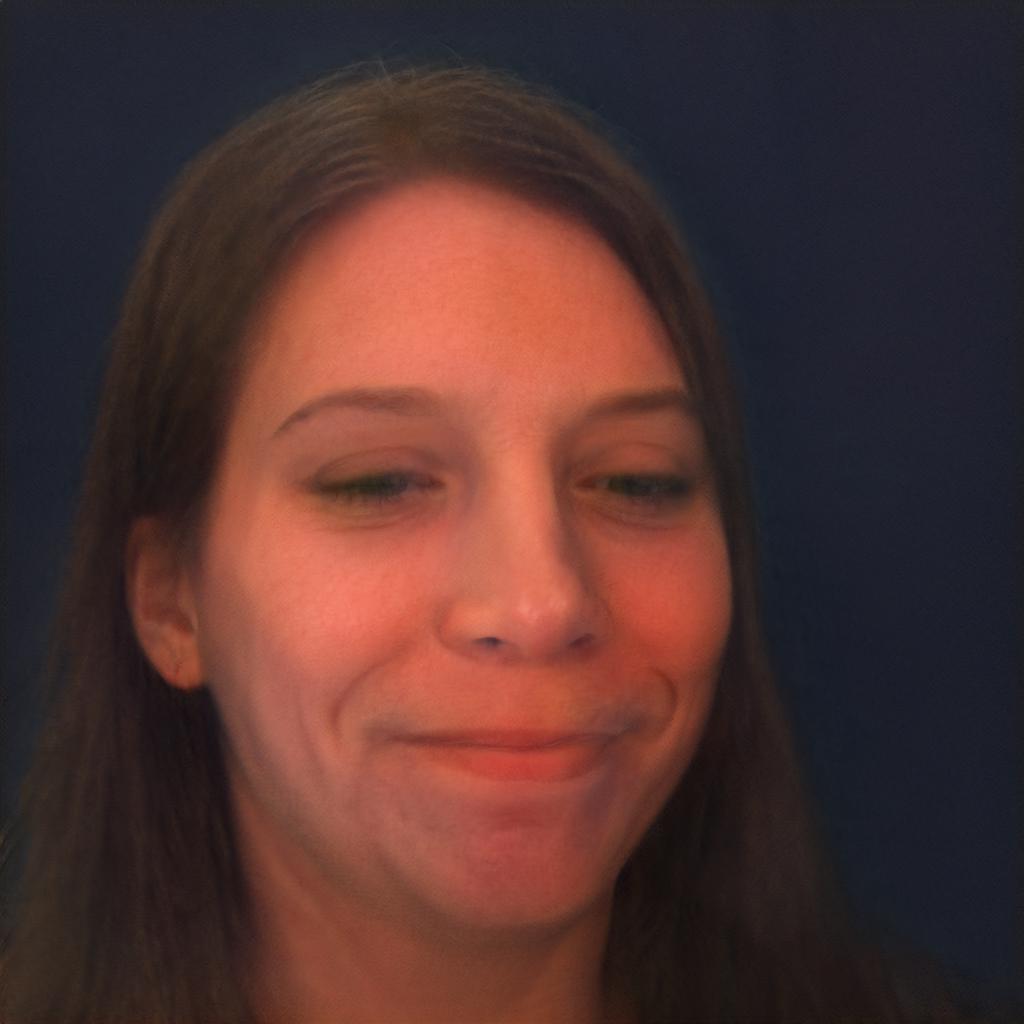}
        \end{minipage}
        \caption{ReDirTrans \\ ($+$)}
  \end{subfigure}
  \centering
    \begin{subfigure}[t]{0.093\linewidth}
        \begin{minipage}{1\linewidth}
        \includegraphics[width=1\linewidth]{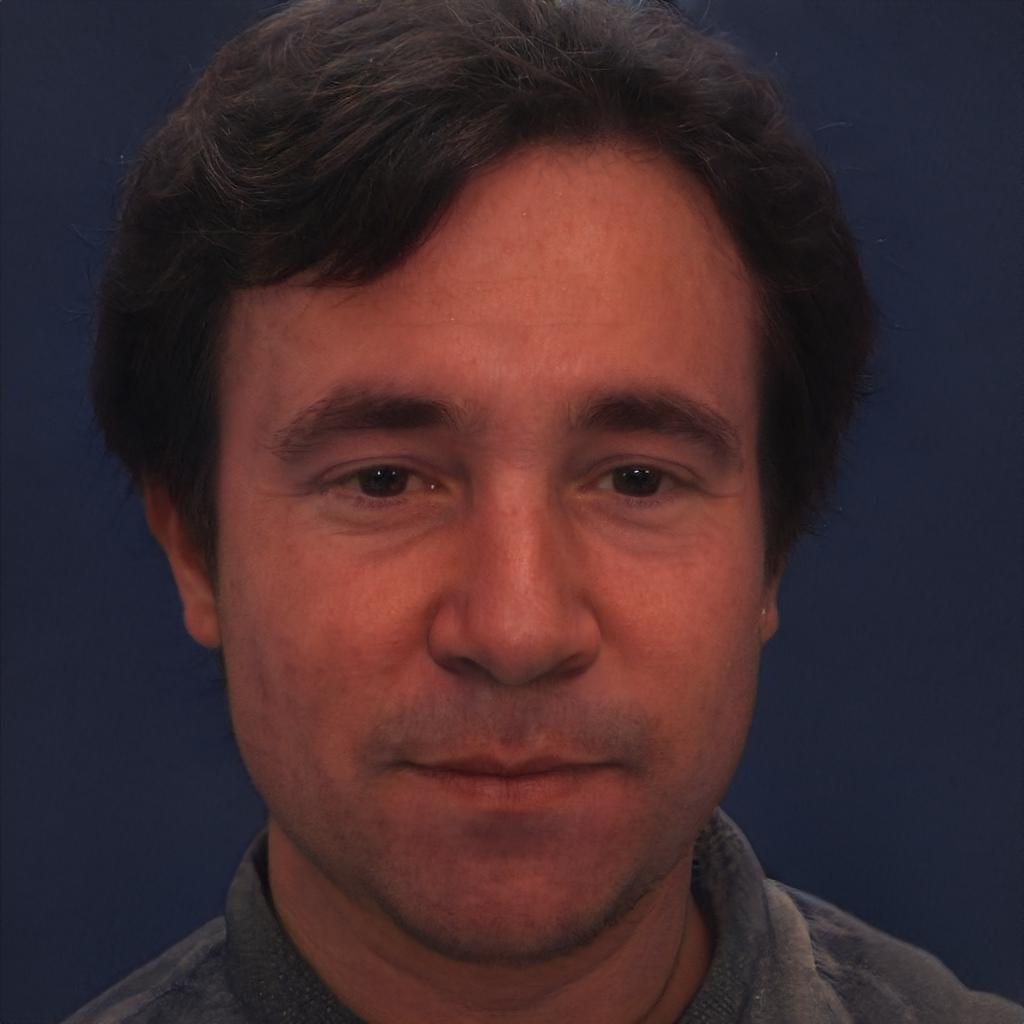}
        \includegraphics[width=1\linewidth]{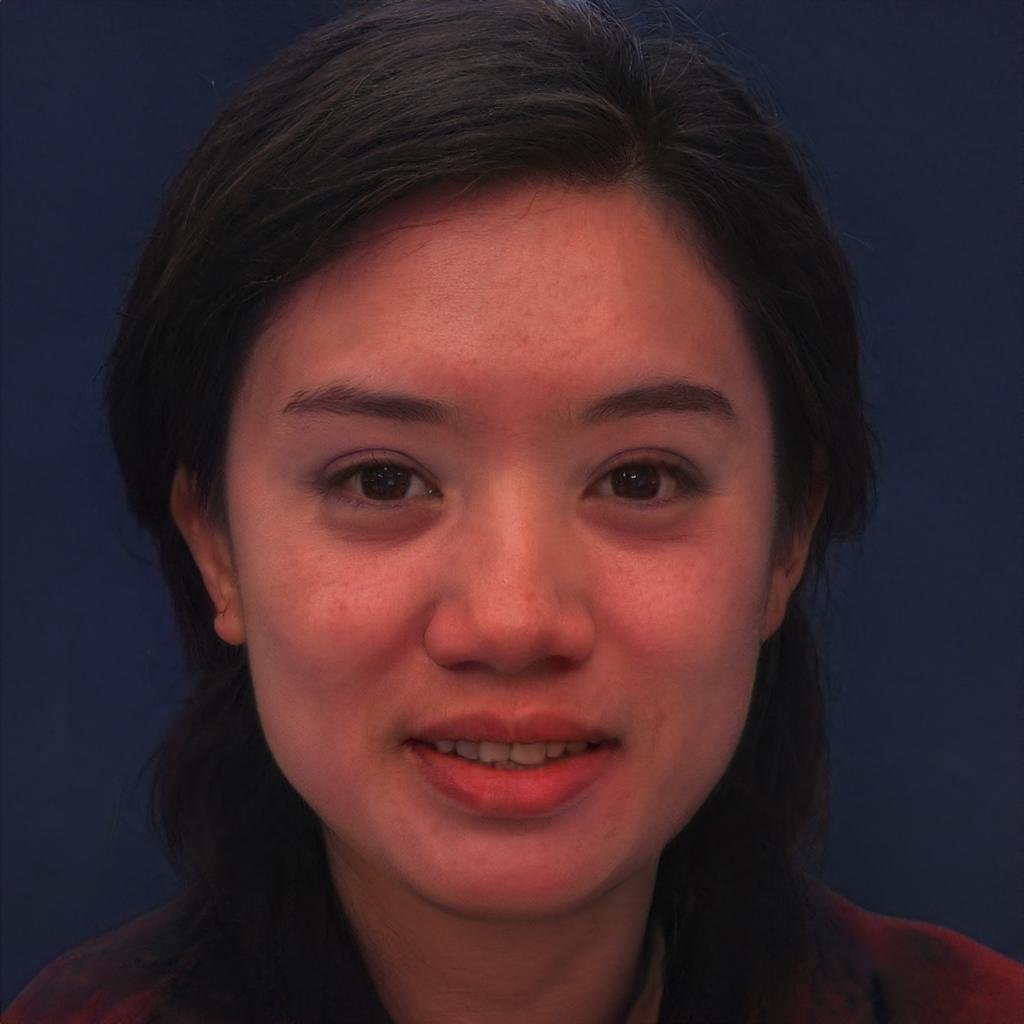}
        \includegraphics[width=1\linewidth]{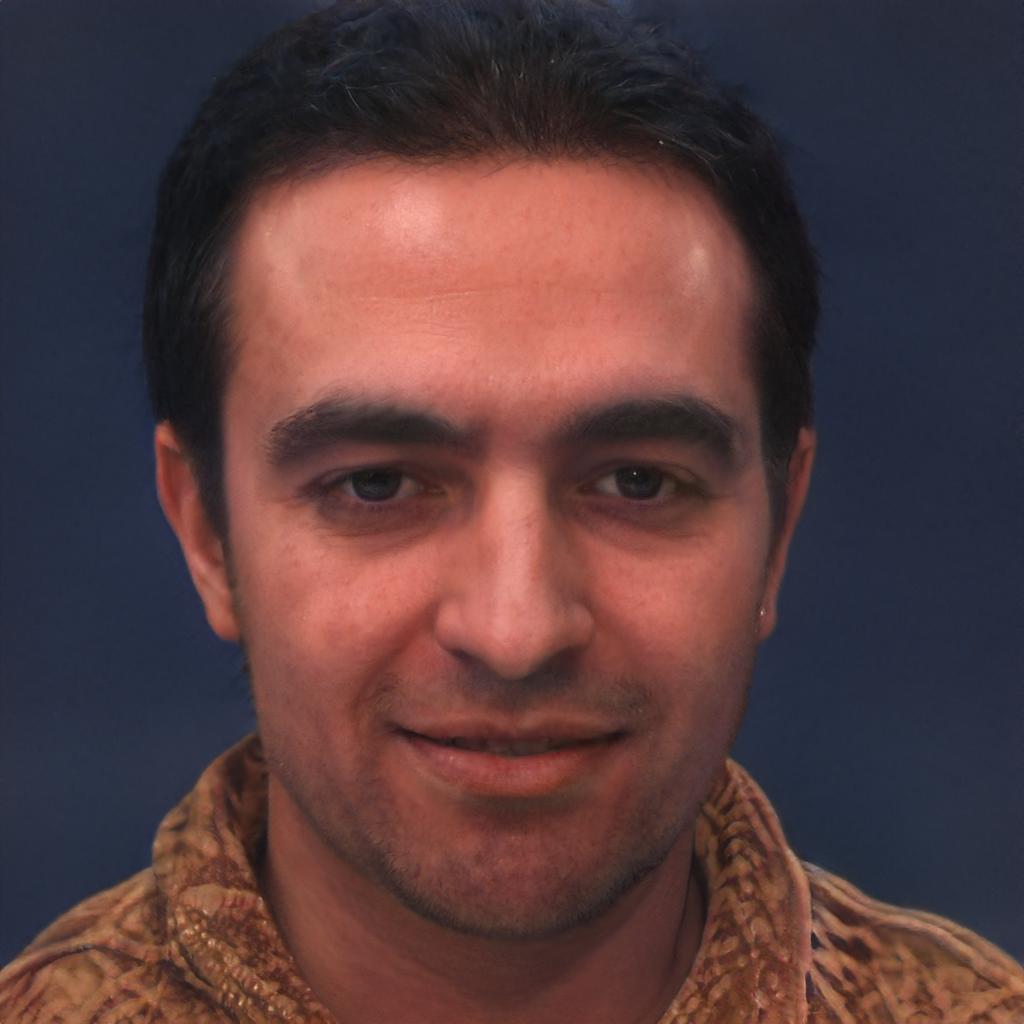}
        \includegraphics[width=1\linewidth]{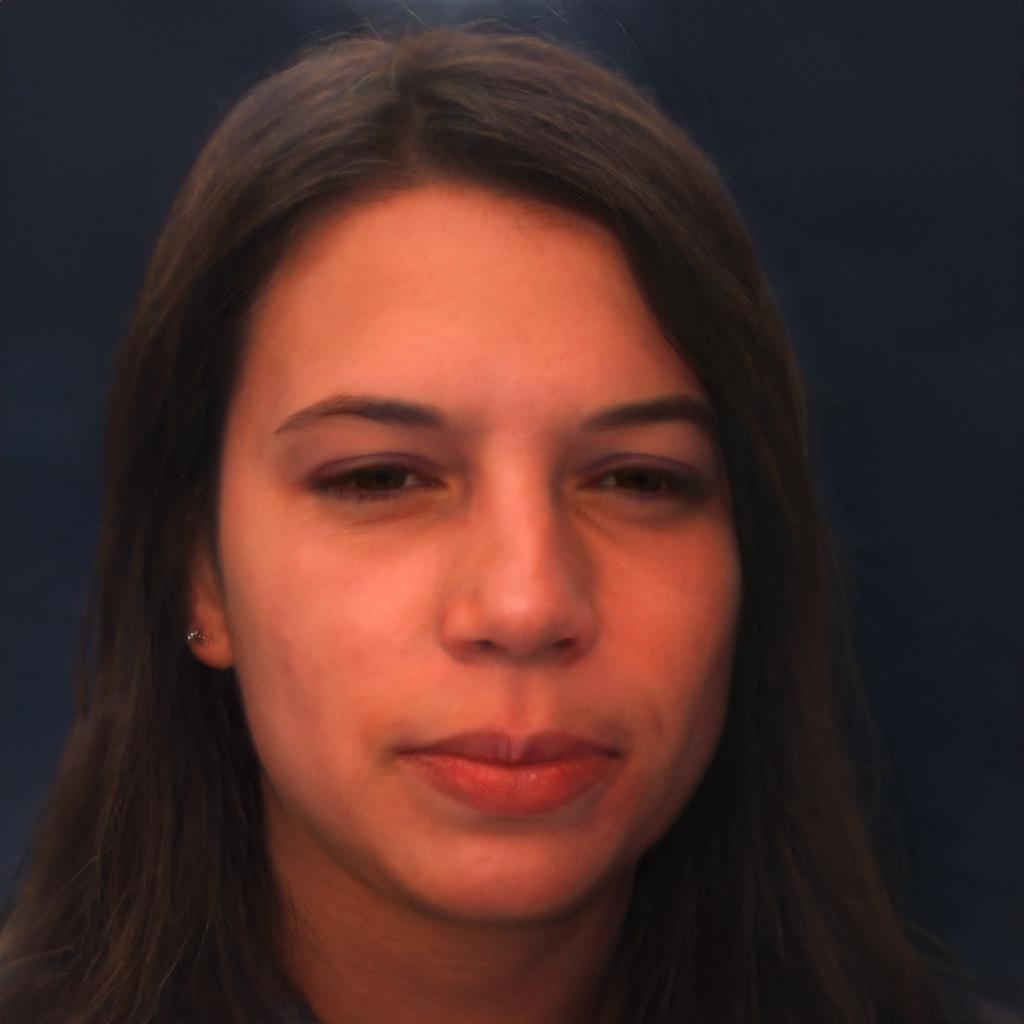}
        \end{minipage}
        \caption{DeltaEdit \\ ($+\Delta$)}
  \end{subfigure}
  \centering
    \begin{subfigure}[t]{0.093\linewidth}
        \begin{minipage}{1\linewidth}
        \includegraphics[width=1\linewidth]{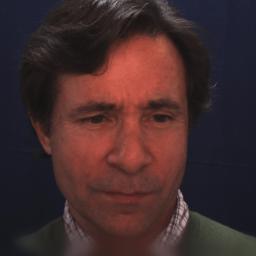}
        \includegraphics[width=1\linewidth]{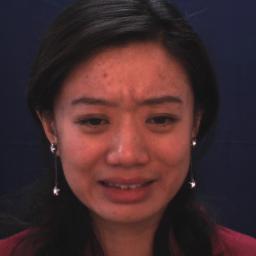}
        \includegraphics[width=1\linewidth]{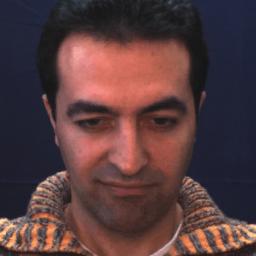}
        \includegraphics[width=1\linewidth]{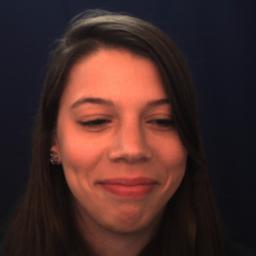}
        \end{minipage}
        \caption{Target}
  \end{subfigure}
  \caption{Comparison of AU intensity manipulation using target AU intensities in DISFA. 
  AUEditNet, ReDirTrans generate editing results that involve the removal ($-$) of source attributes and the addition ($+$) of target attributes. 
  DeltaEdit uses intensity differences between source and target images for attribute addition ($+\Delta$). 
  The removal ($-$) process yields 'neutral-like' face images with all AU intensities set to zero.
   } 
  \label{fig:disfa_comp}
\end{figure*}
\paragraph{Within-Dataset Evaluation}
Fig. \ref{fig:disfa_comp} illustrates a qualitative comparison of AU intensity manipulation based on provided target conditions. 
Both ReDirTrans \cite{jin2023redirtrans} and our proposed AUEditNet employ a two-step editing process to prevent potential attribute status mixing. 
After the source status removal, the generated images should exhibit all AU intensities set to zero, serving as benchmarks when all AUs are deactivated. 
ReDirTrans and AUEditNet demonstrate the ability to learn the desired AU movements, under both cases when deactivating all AUs or assigning new target intensities. 
However, ReDirTrans fails to preserve identity information in intermediate and final generated images. 
Additionally, ReDirTrans attempts to address color discrepancy between real and inverted images during AU editing, resulting in undesired color distortion in images. 
In contrast, AUEditNet focuses only on editing the aimed AUs' intensities, devoid of unrelated information, which is achieved through the dual-branch architecture. 
On the other hand, DeltaEdit \cite{lyu2023deltaedit} excels in maintaining identity information and other facial attributes. 
However, it is limited to learning noticeable AU movements and may ignore subtle motions such as eyebrow, cheek, and lip corner movements, potentially causing significant changes in the entire facial expression. 
AUEditNet successfully achieves accurate AU intensity editing under this two-phase editing process while maintaining identity. 

\begin{figure}
    \captionsetup[subfigure]{labelformat=empty}
    \captionsetup[subfigure]{justification=centering}
    \centering
        \begin{minipage}{0.3cm}
        \rotatebox{90}{\scriptsize{~~~~~~~~~~~ AU $25$  ~~~~~~~~~~~~~ AU $15$ ~~~~~~~~~~~~~~ AU $5$ ~~~~~~~~~~~~~~~ AU $4$ ~~~~~~~~~~~~~~~ AU $2$ ~~~~}}
        \end{minipage}%
    \begin{subfigure}[t]{0.187\linewidth}
        \begin{minipage}{1\linewidth}
        \includegraphics[width=1\linewidth]{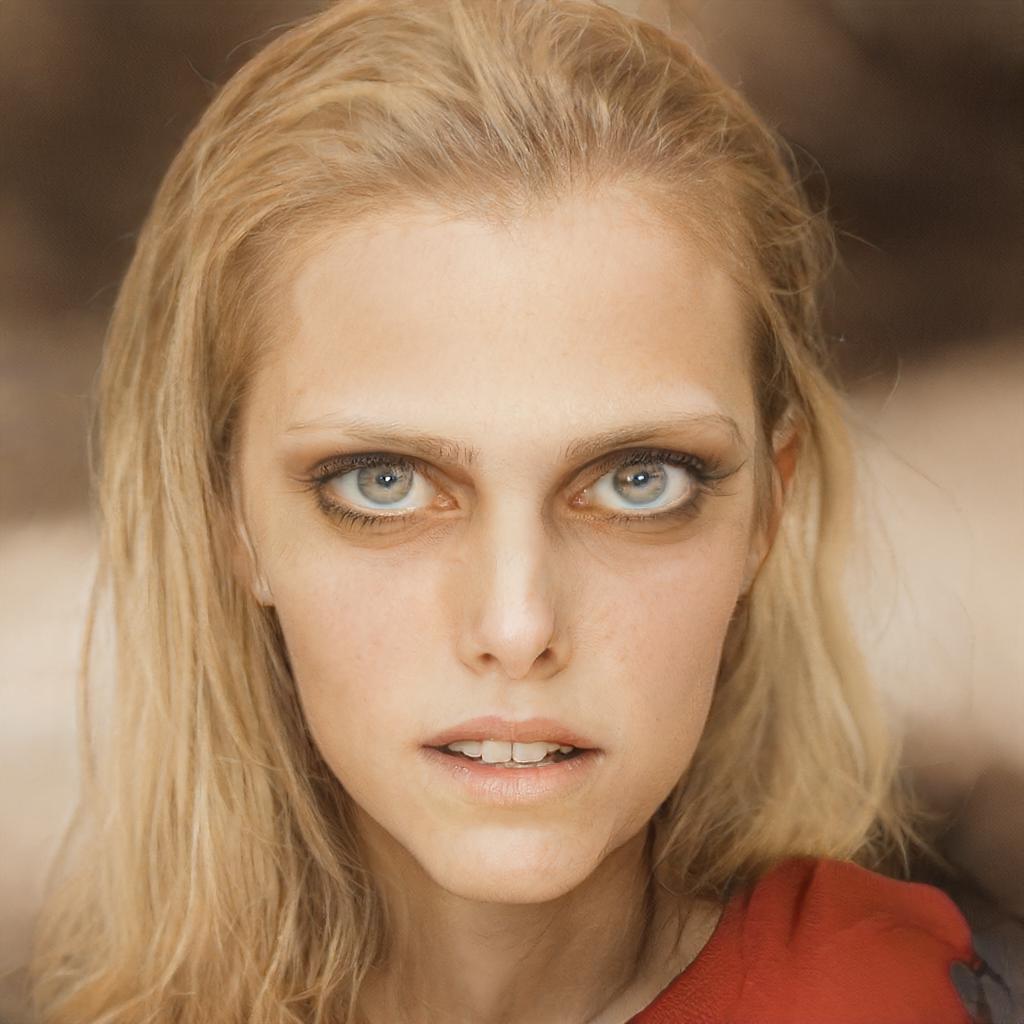}
        \includegraphics[width=1\linewidth]{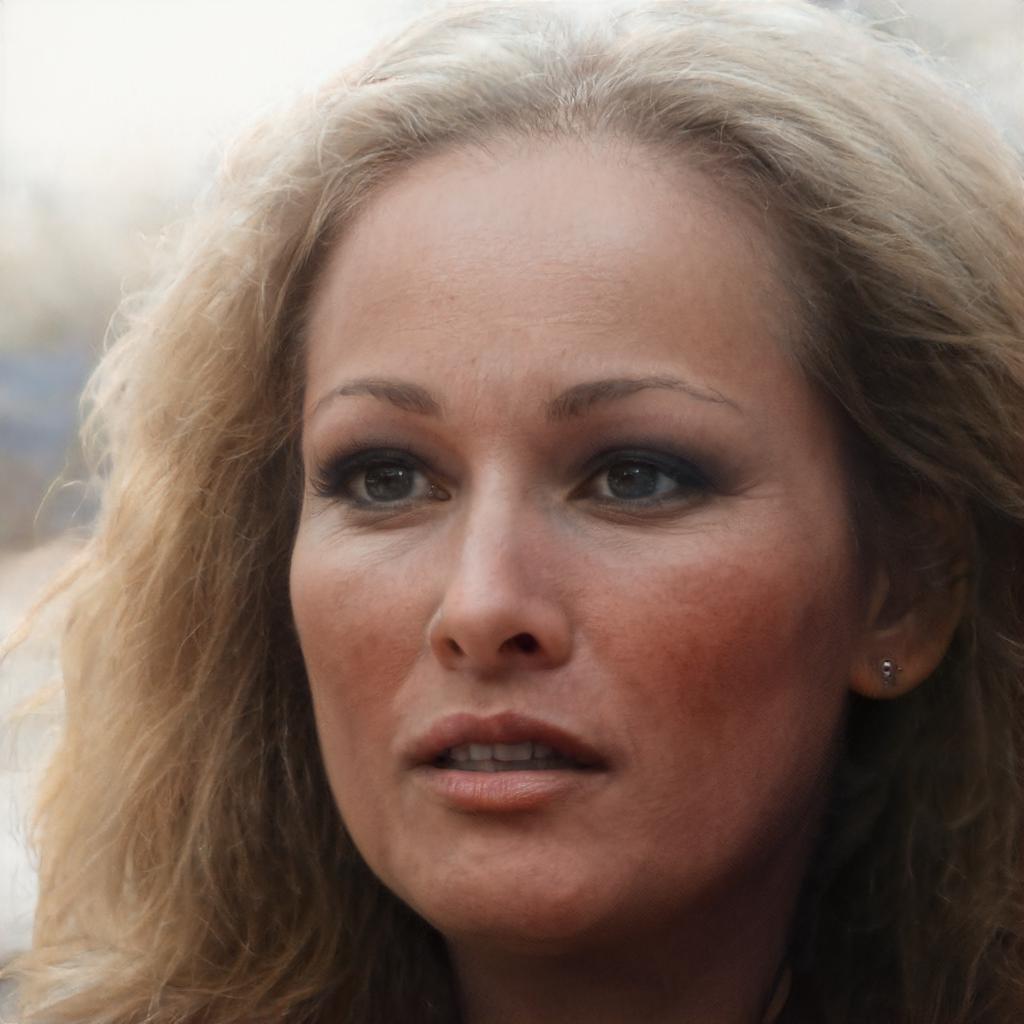}
        \includegraphics[width=1\linewidth]{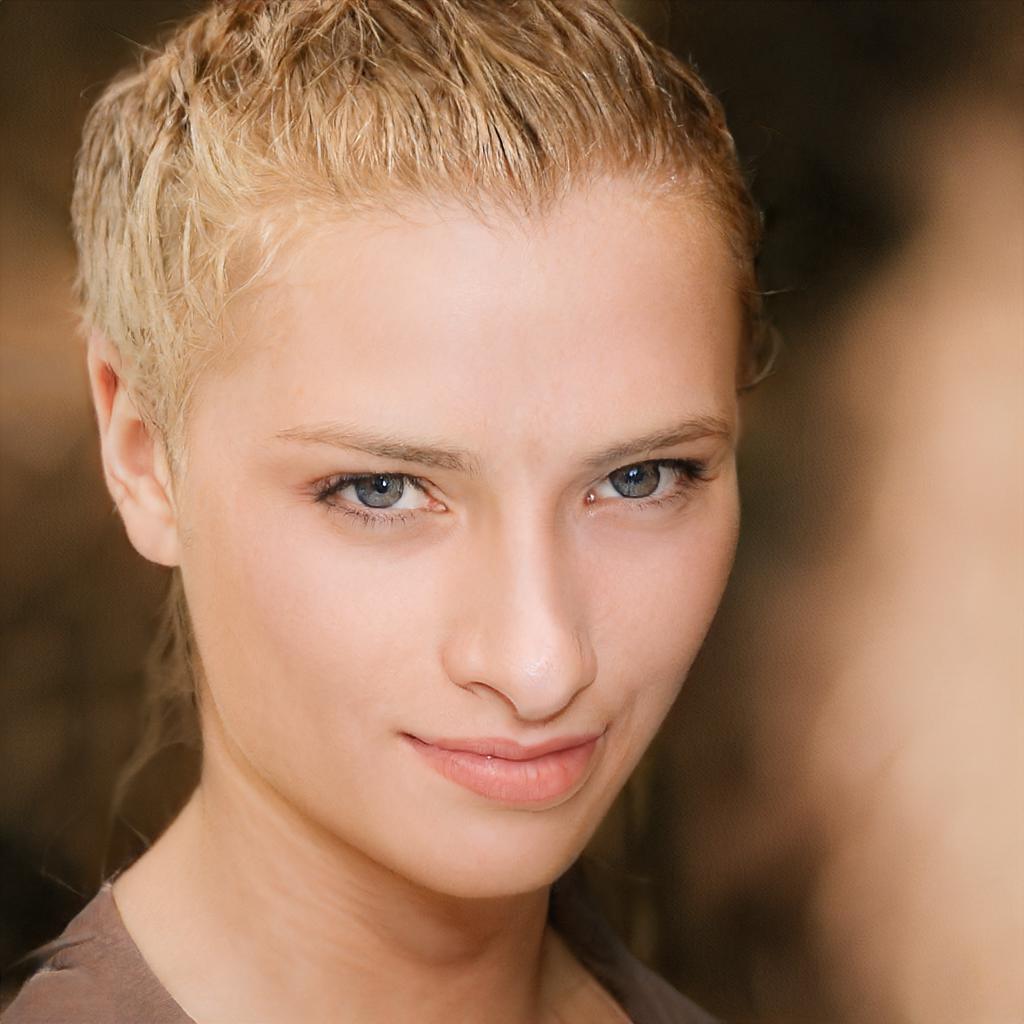}
        \includegraphics[width=1\linewidth]{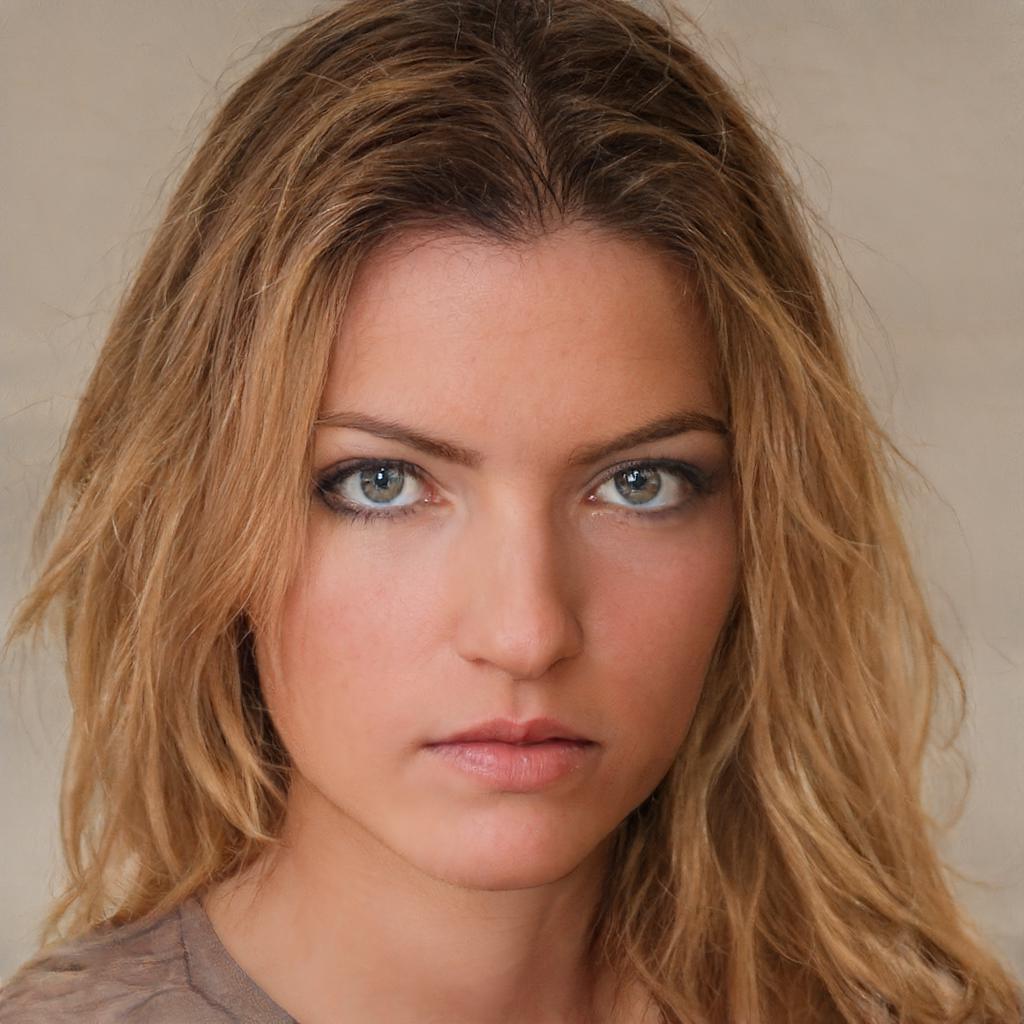}
        \includegraphics[width=1\linewidth]{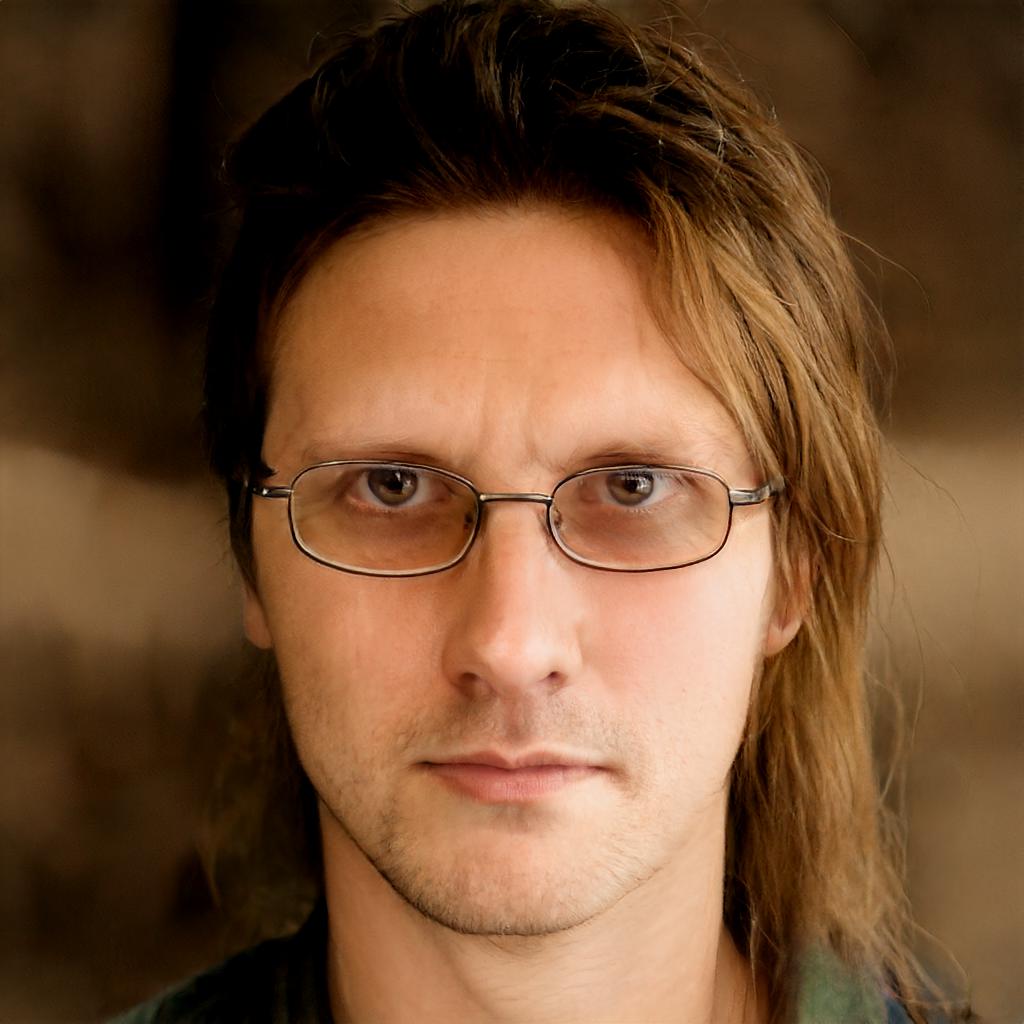}
        \caption{Inversion}
        \end{minipage}
    \label{fig:comp_input00}
  \end{subfigure}
  \hspace{-0.021\linewidth}
  \centering
    \begin{subfigure}[t]{0.187\linewidth}
        \begin{minipage}{1\linewidth}
        \includegraphics[width=1\linewidth]{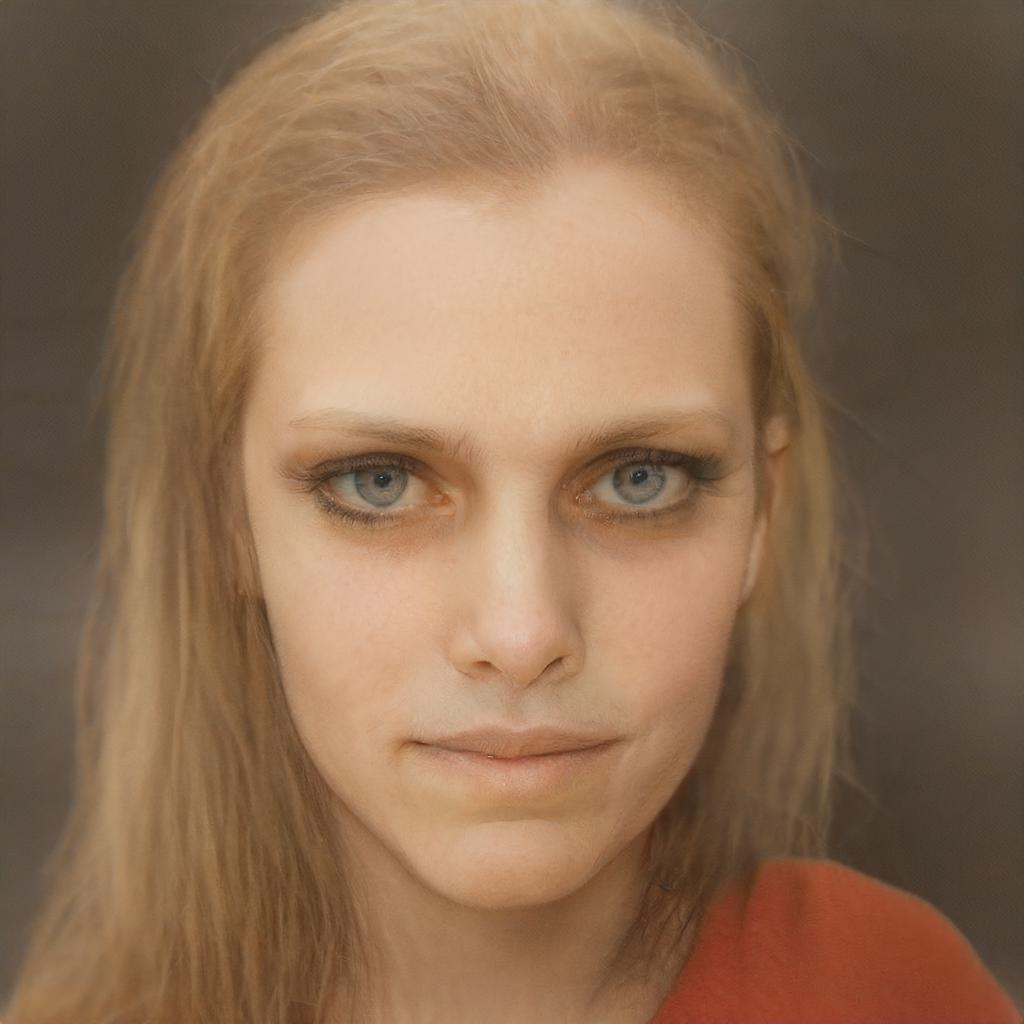}
        \includegraphics[width=1\linewidth]{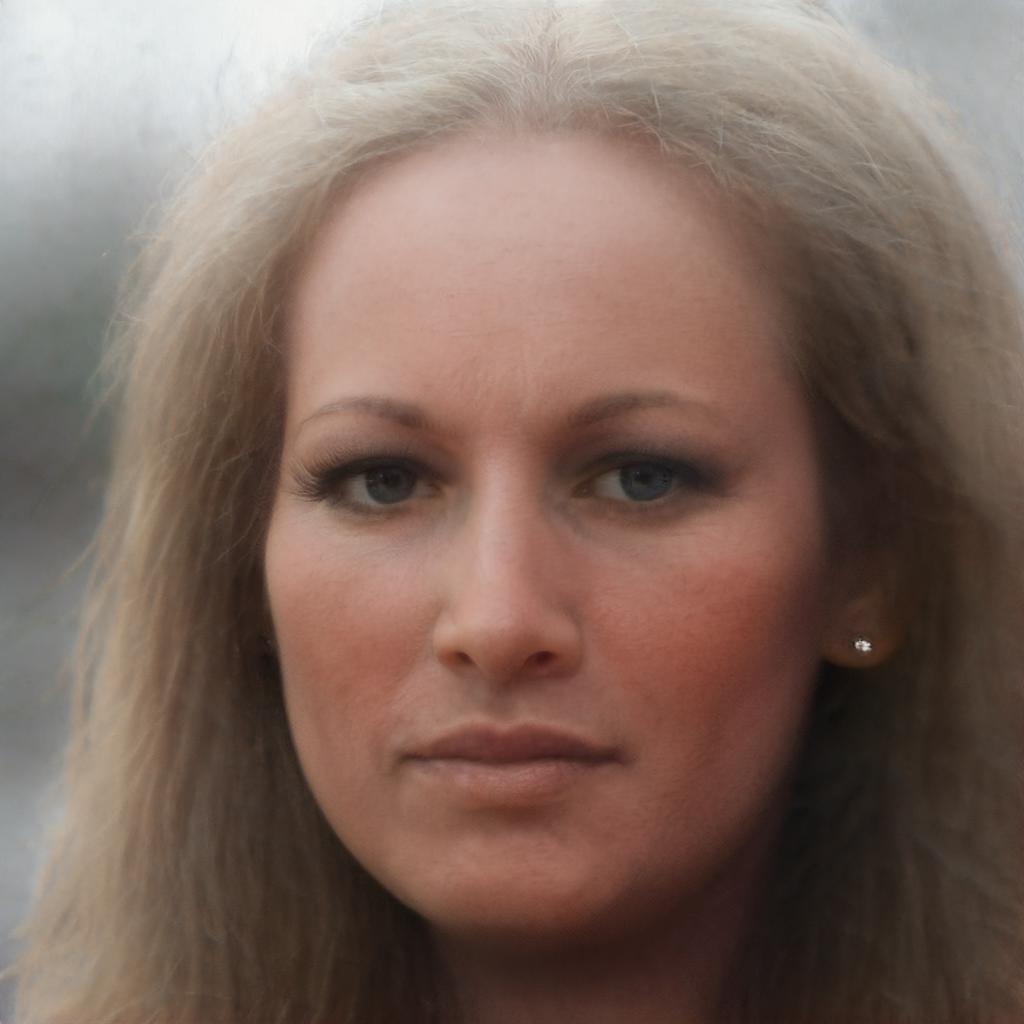}
        \includegraphics[width=1\linewidth]{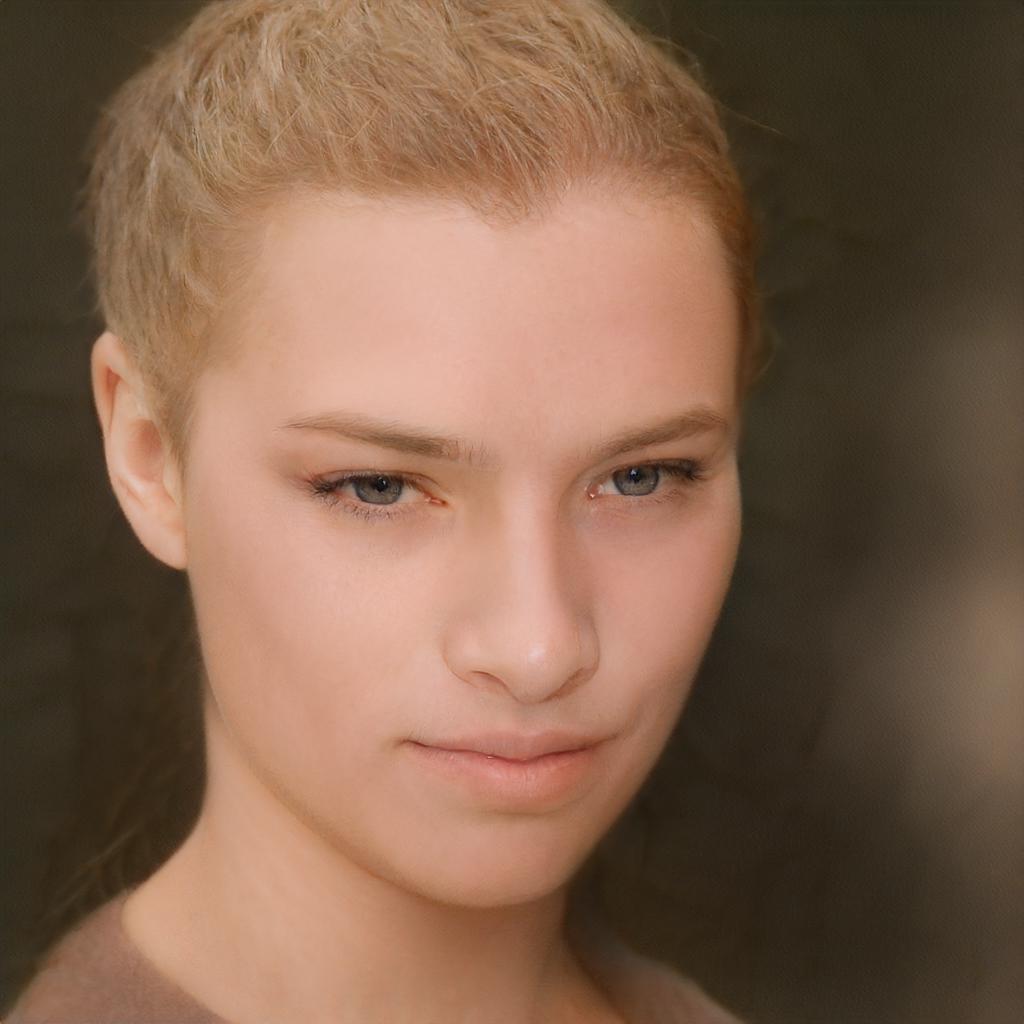}
        \includegraphics[width=1\linewidth]{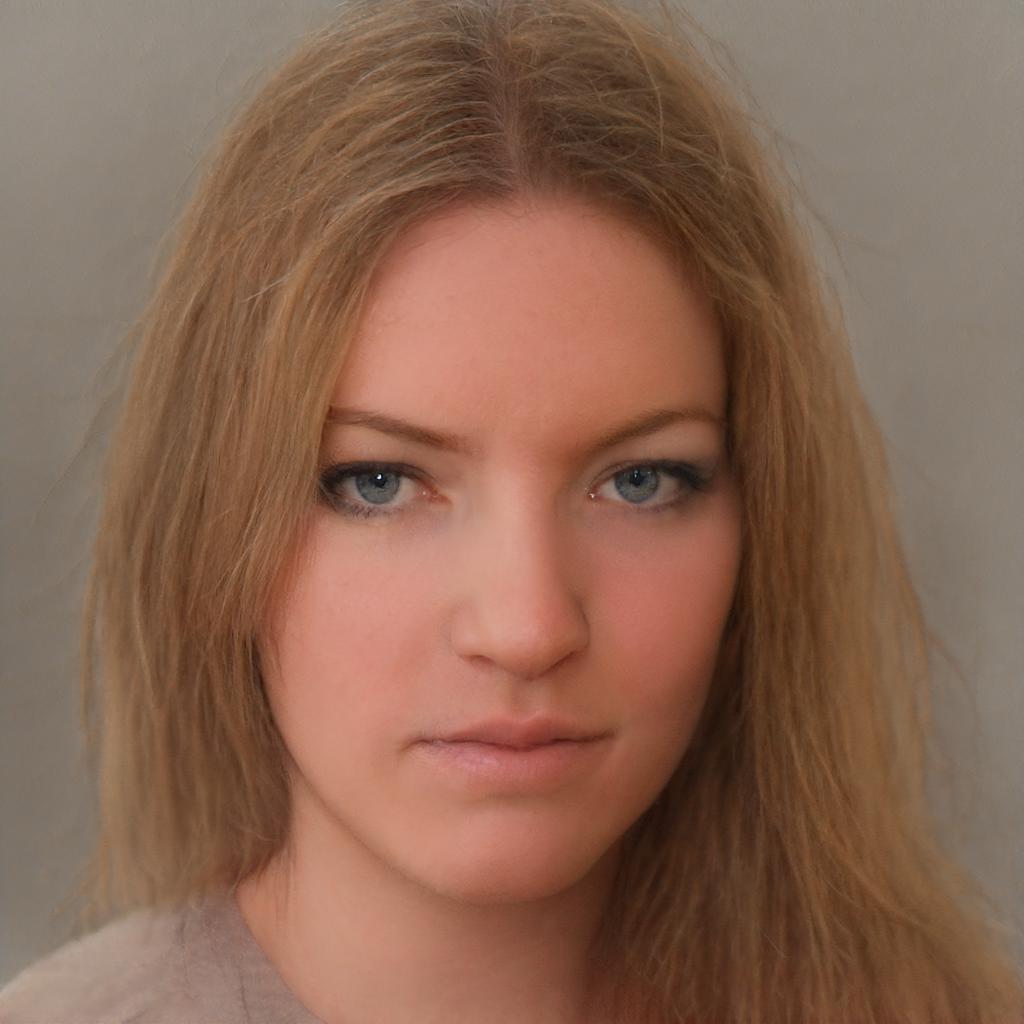}
        \includegraphics[width=1\linewidth]{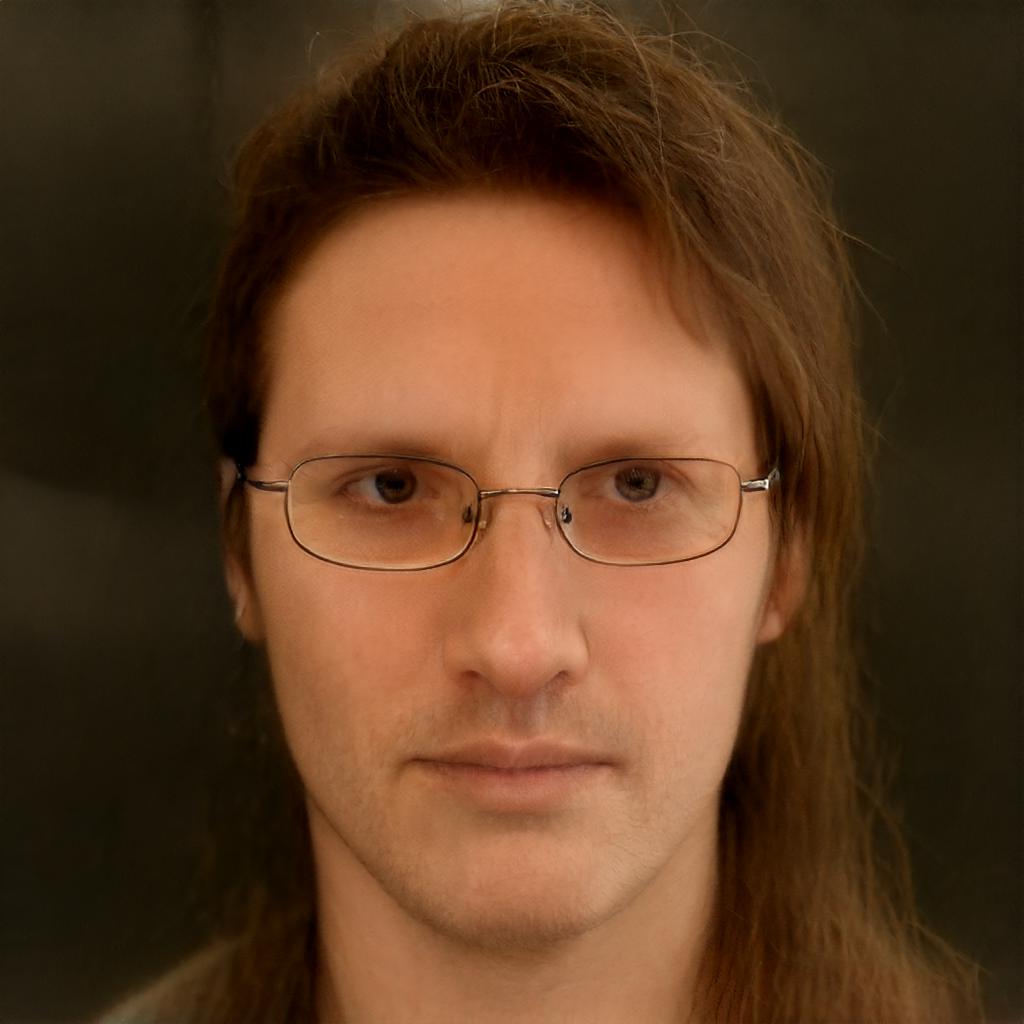}
        \caption{$a_{tar} = 0$}
        \end{minipage}
    \label{fig:comp_inver00}
  \end{subfigure}
  \hspace{-0.021\linewidth}
  \centering
    \begin{subfigure}[t]{0.187\linewidth}
        \begin{minipage}{1\linewidth}
        \includegraphics[width=1\linewidth]{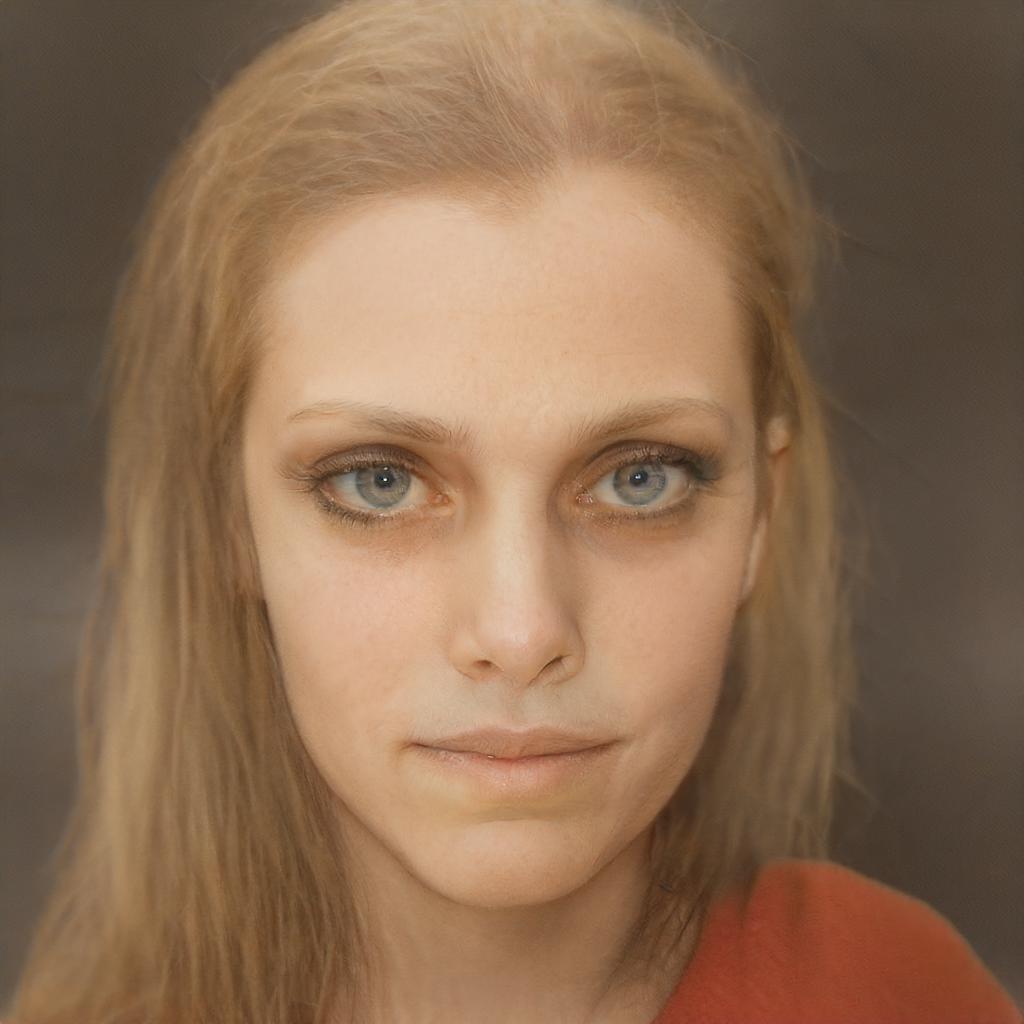}
        \includegraphics[width=1\linewidth]{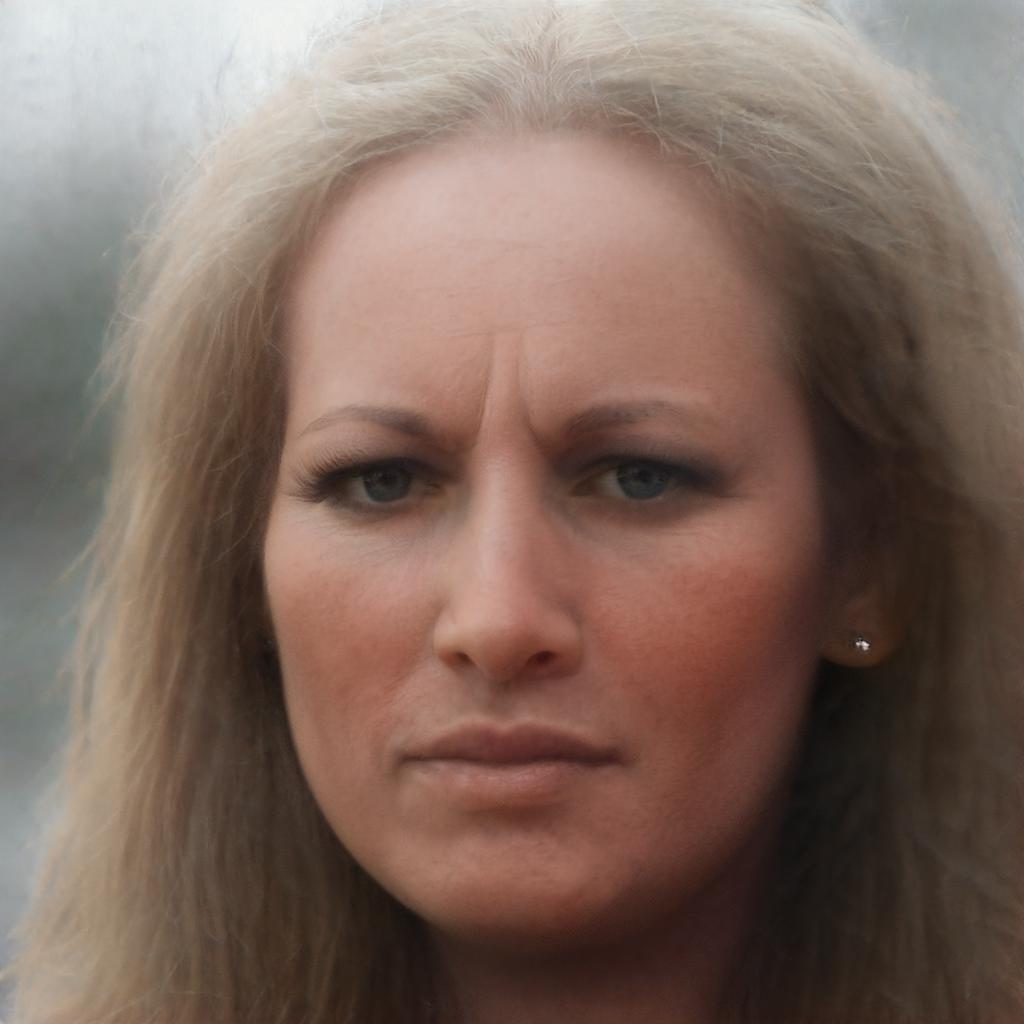}
        \includegraphics[width=1\linewidth]{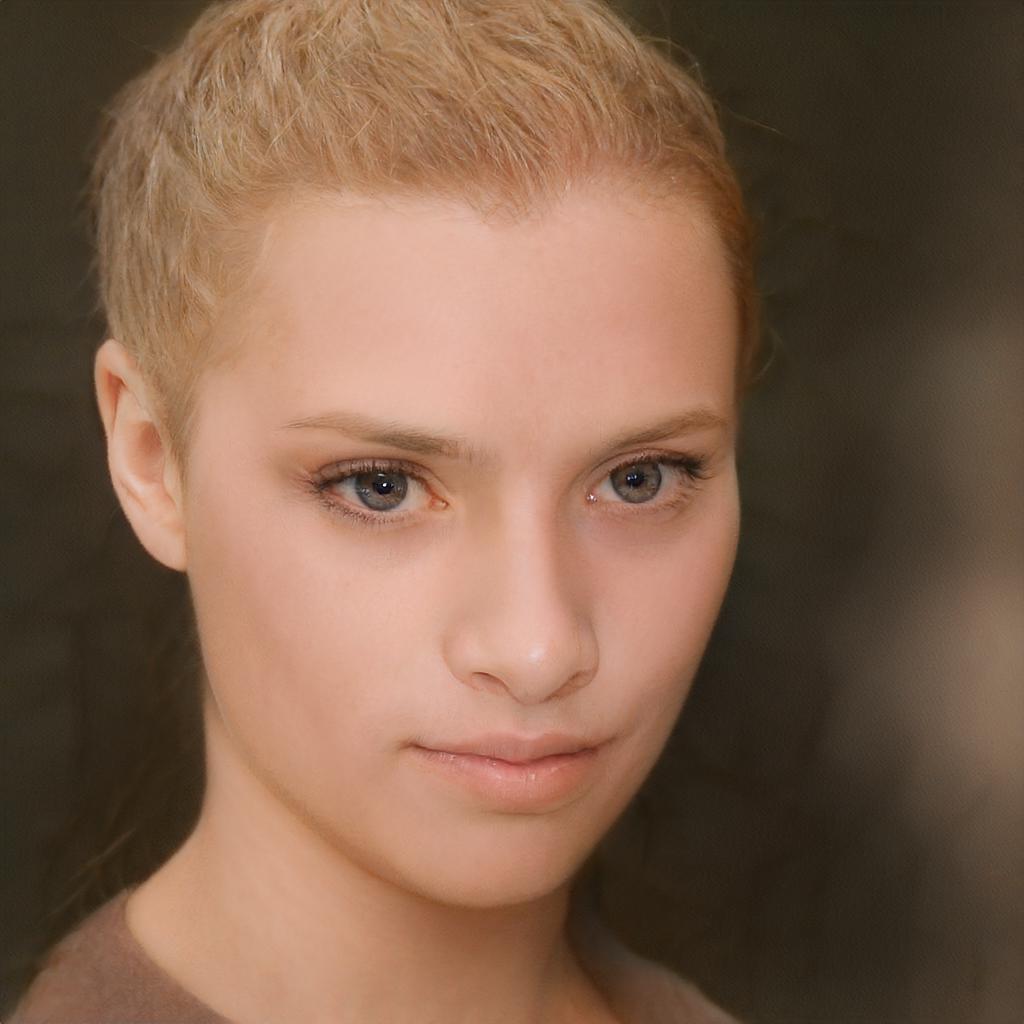}
        \includegraphics[width=1\linewidth]{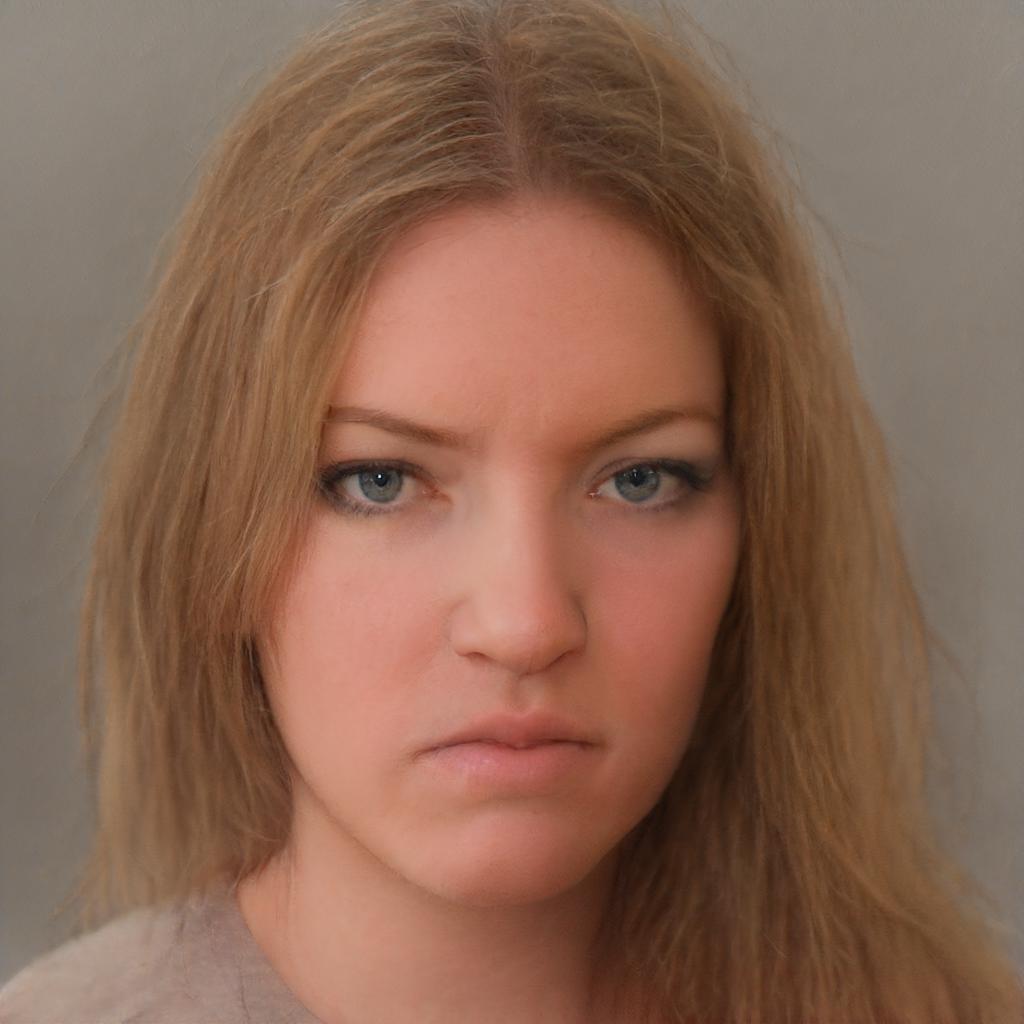}
        \includegraphics[width=1\linewidth]{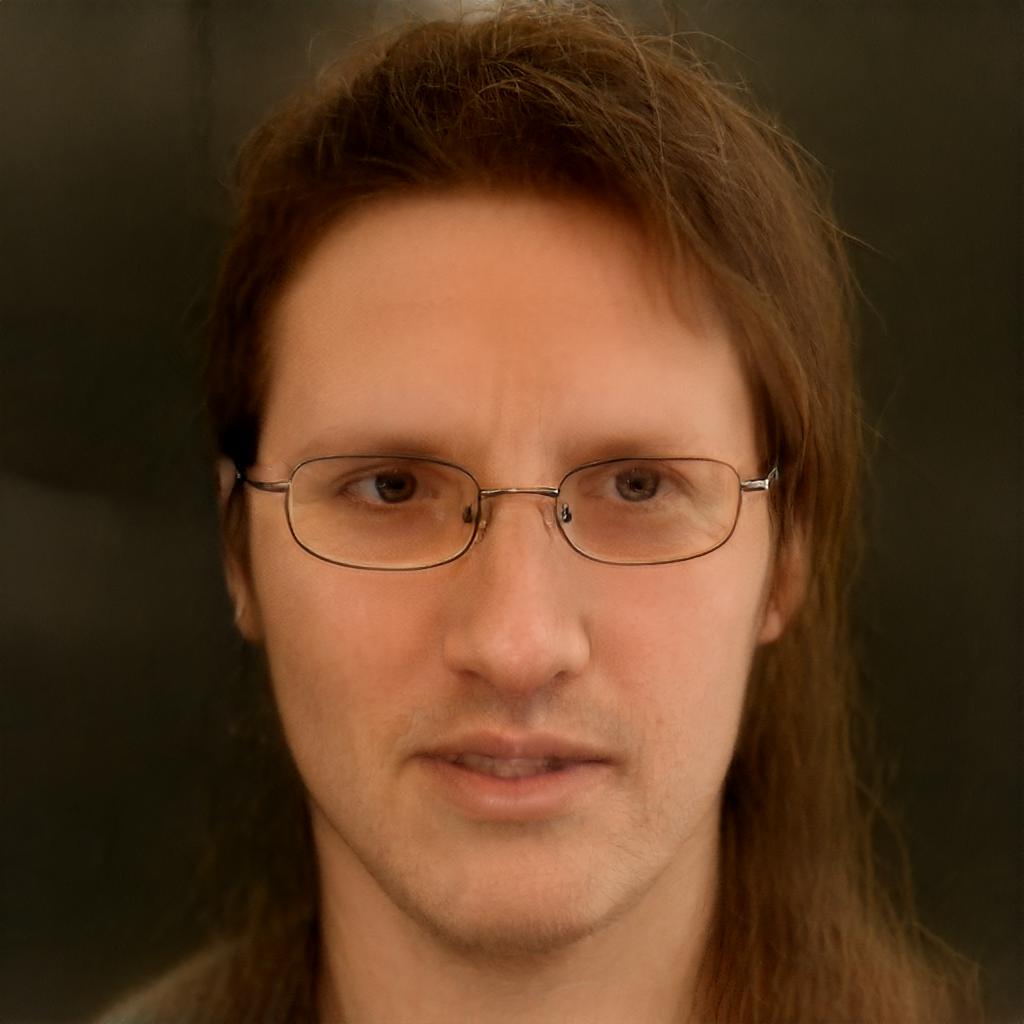}
        \caption{$a_{tar} = 2$}
        \end{minipage}
    \label{fig:comp_redir00}
  \end{subfigure}
  \hspace{-0.021\linewidth}
  \centering
    \begin{subfigure}[t]{0.187\linewidth}
        \begin{minipage}{1\linewidth}
        \includegraphics[width=1\linewidth]{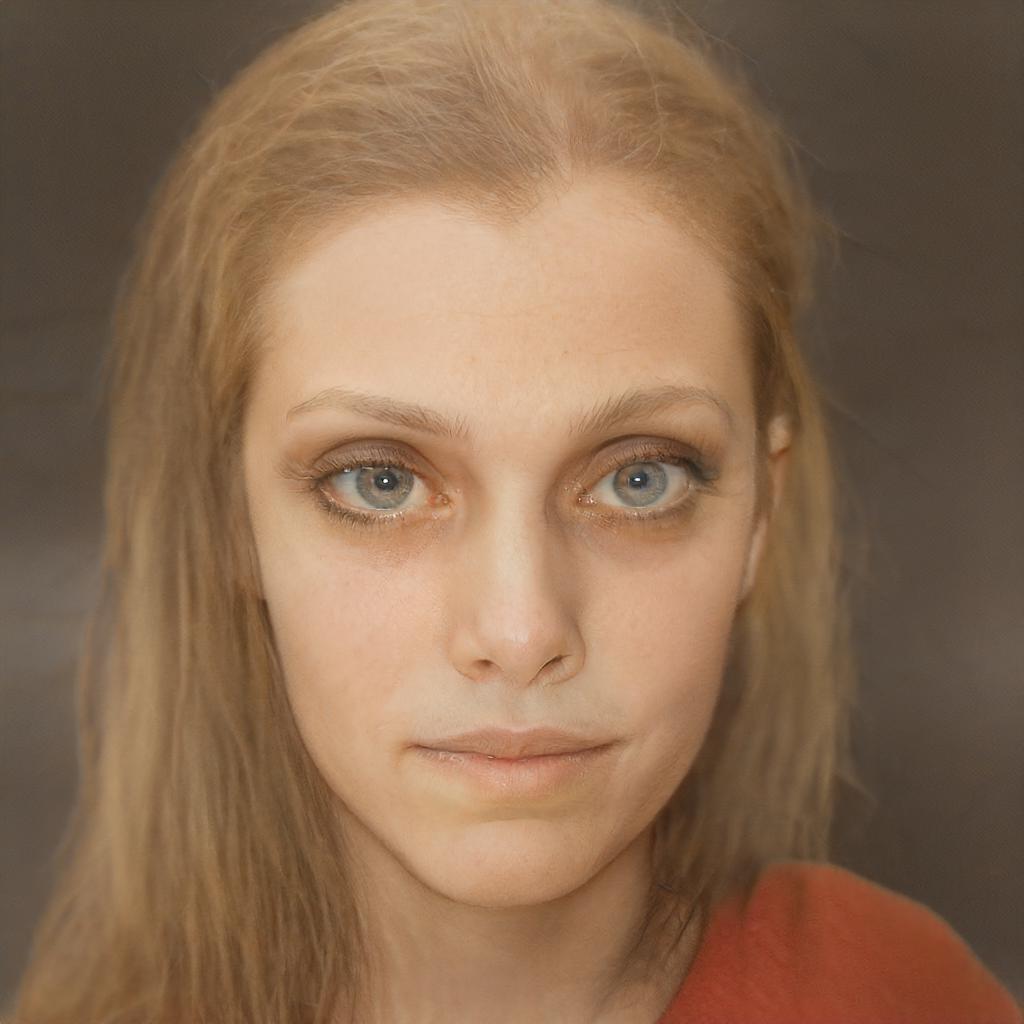}
        \includegraphics[width=1\linewidth]{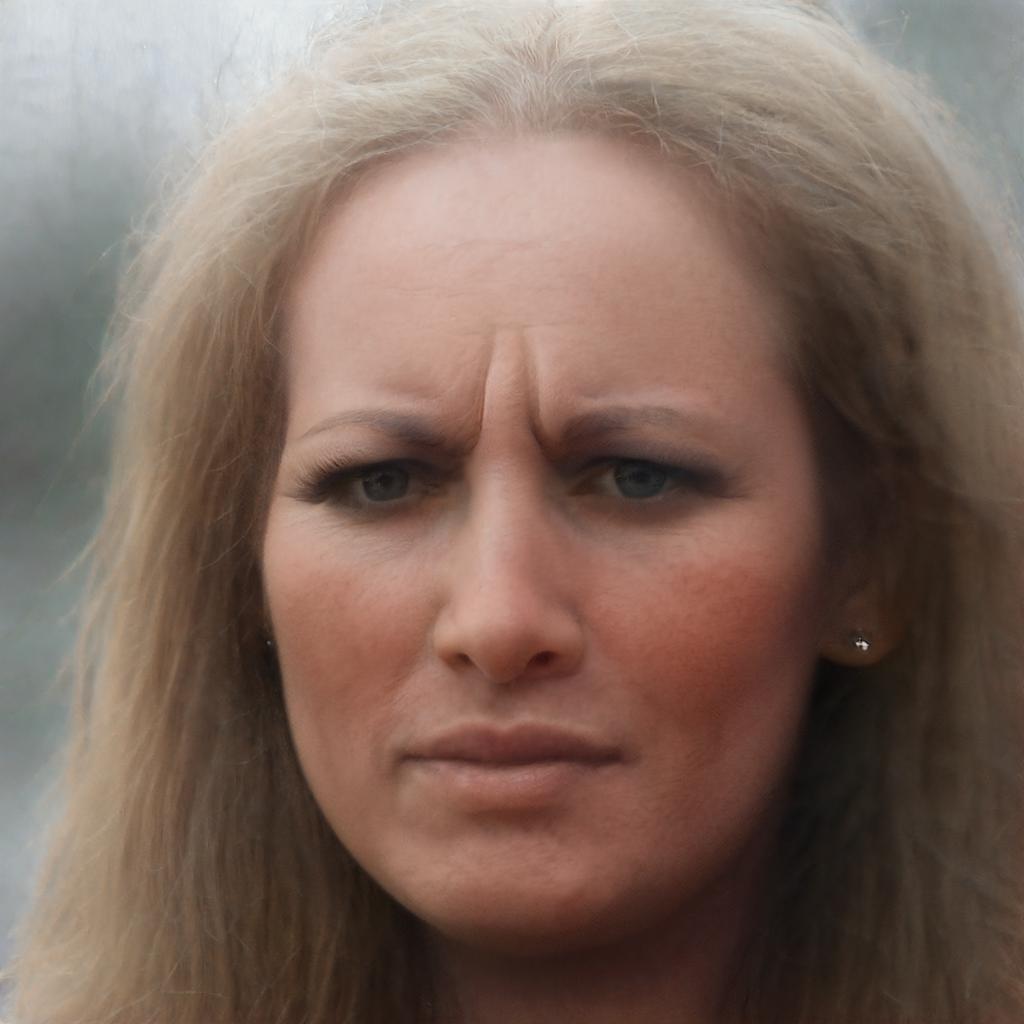}
        \includegraphics[width=1\linewidth]{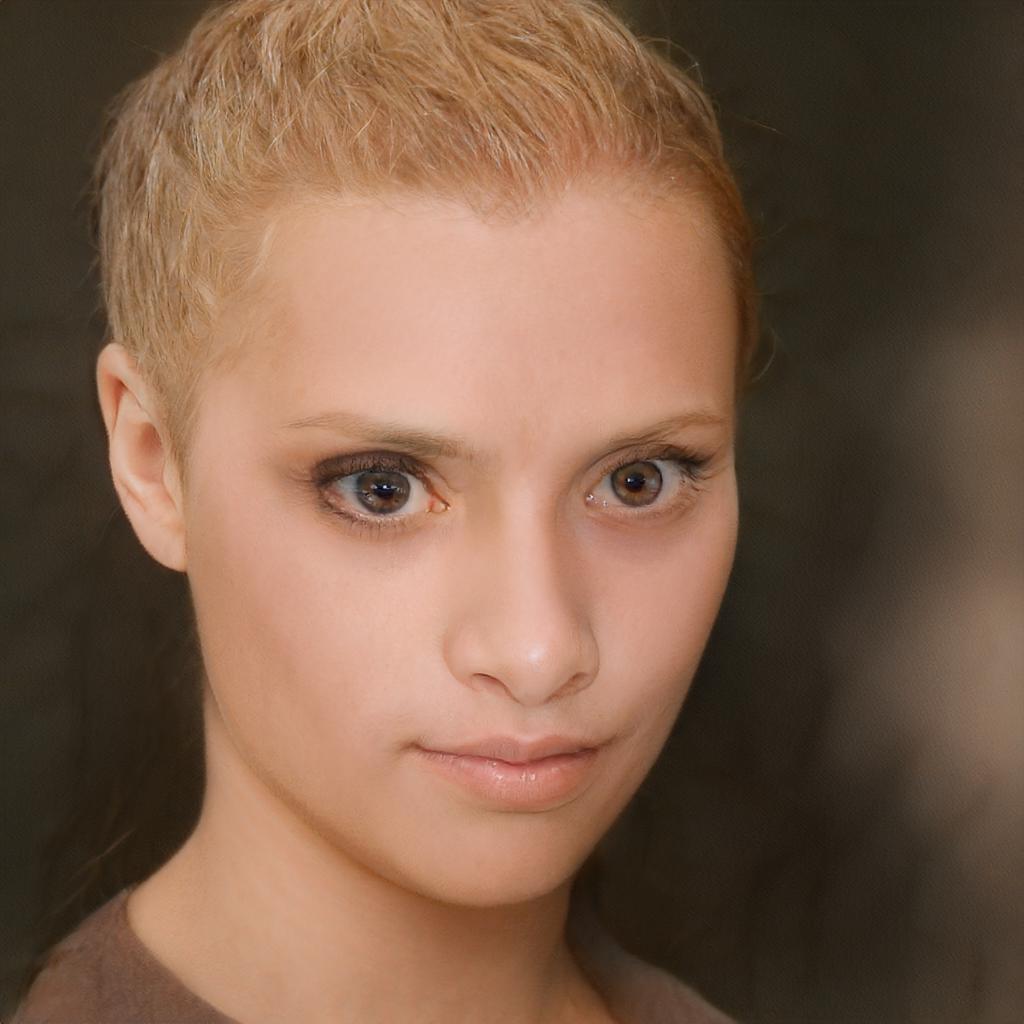}
        \includegraphics[width=1\linewidth]{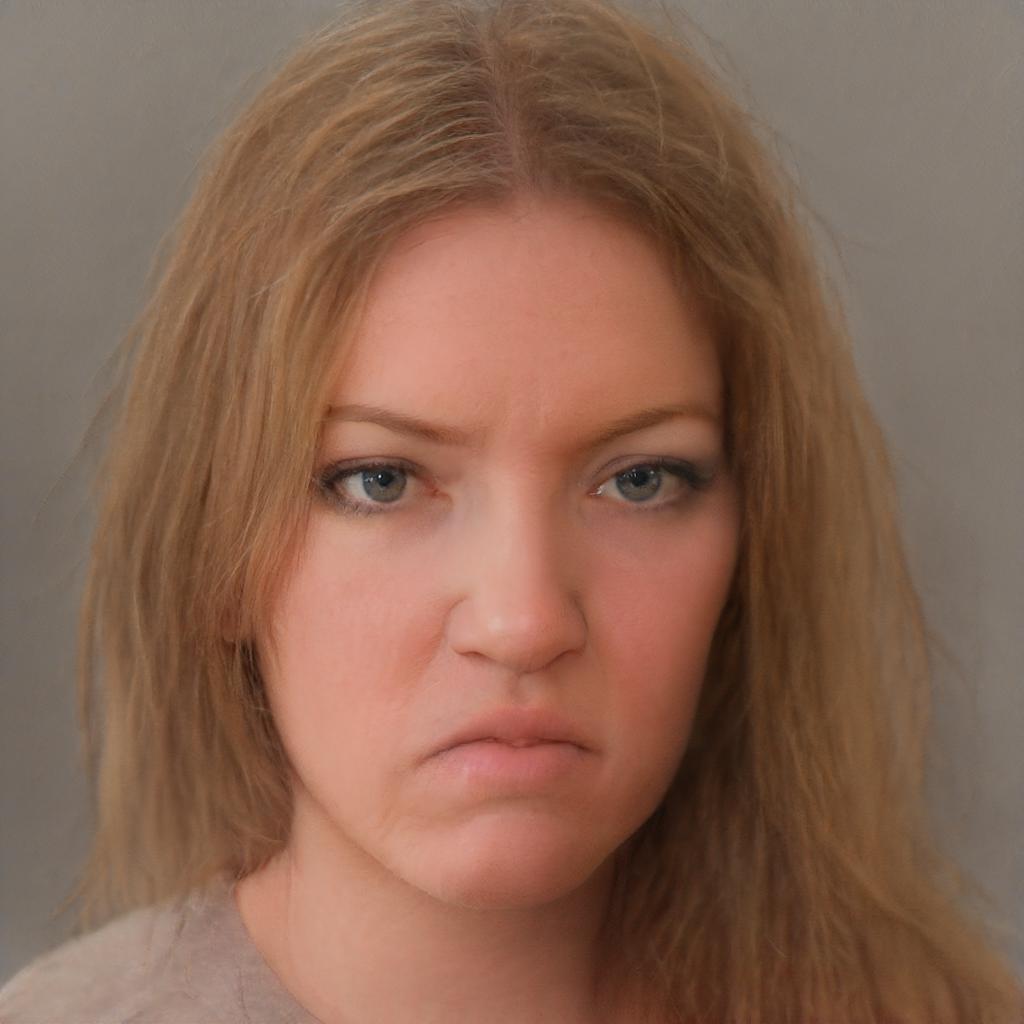}
        \includegraphics[width=1\linewidth]{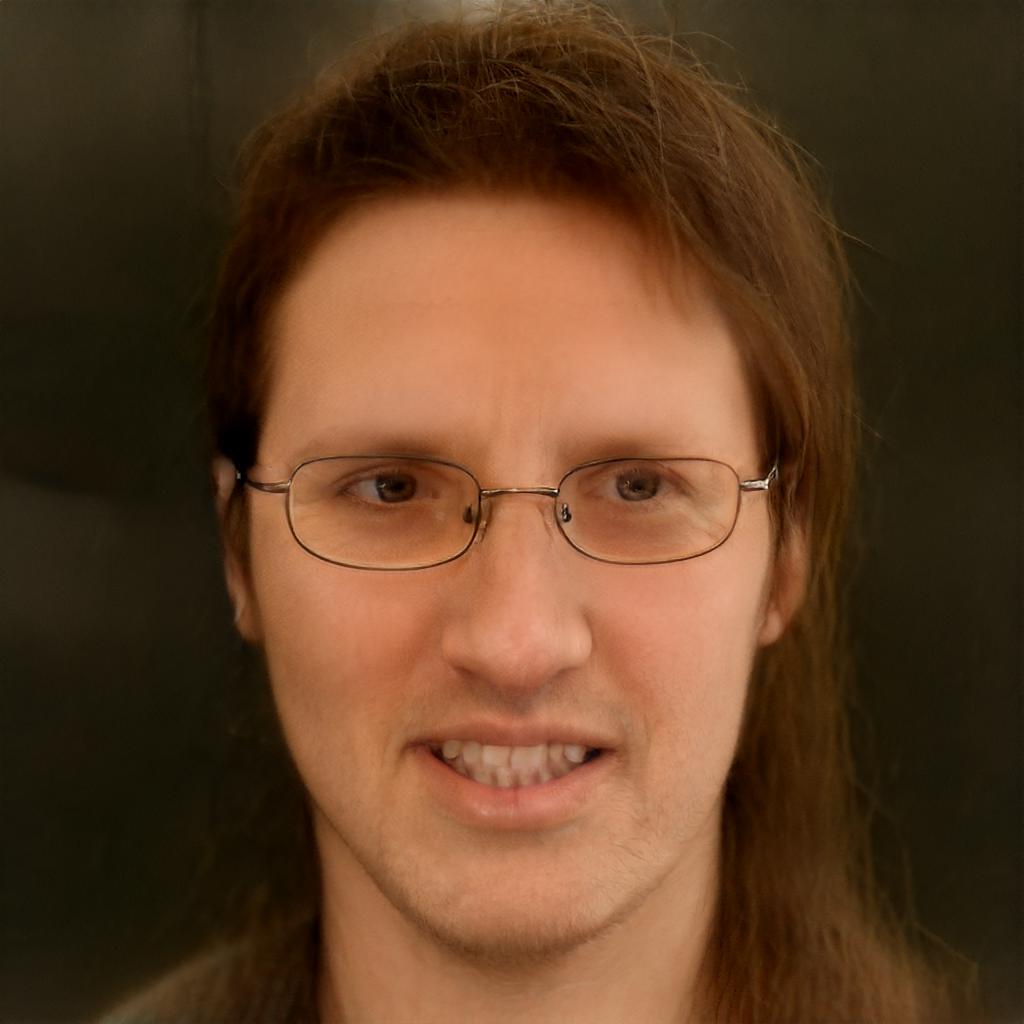}
        \caption{$a_{tar} = 4$}
        \end{minipage}
    \label{fig:comp_tar00}
  \end{subfigure}
  \hspace{-0.021\linewidth}
  \centering
    \begin{subfigure}[t]{0.187\linewidth}
        \begin{minipage}{1\linewidth}
        \includegraphics[width=1\linewidth]{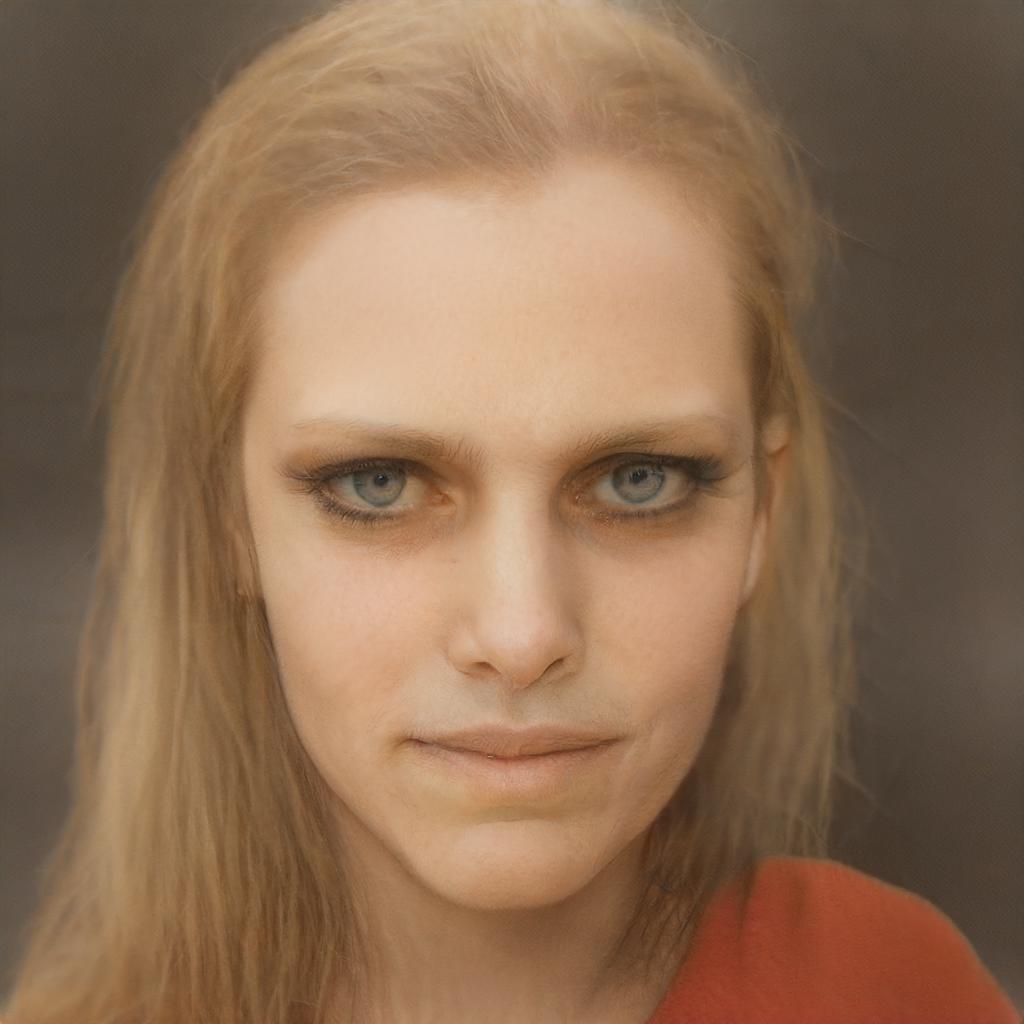}
        \includegraphics[width=1\linewidth]{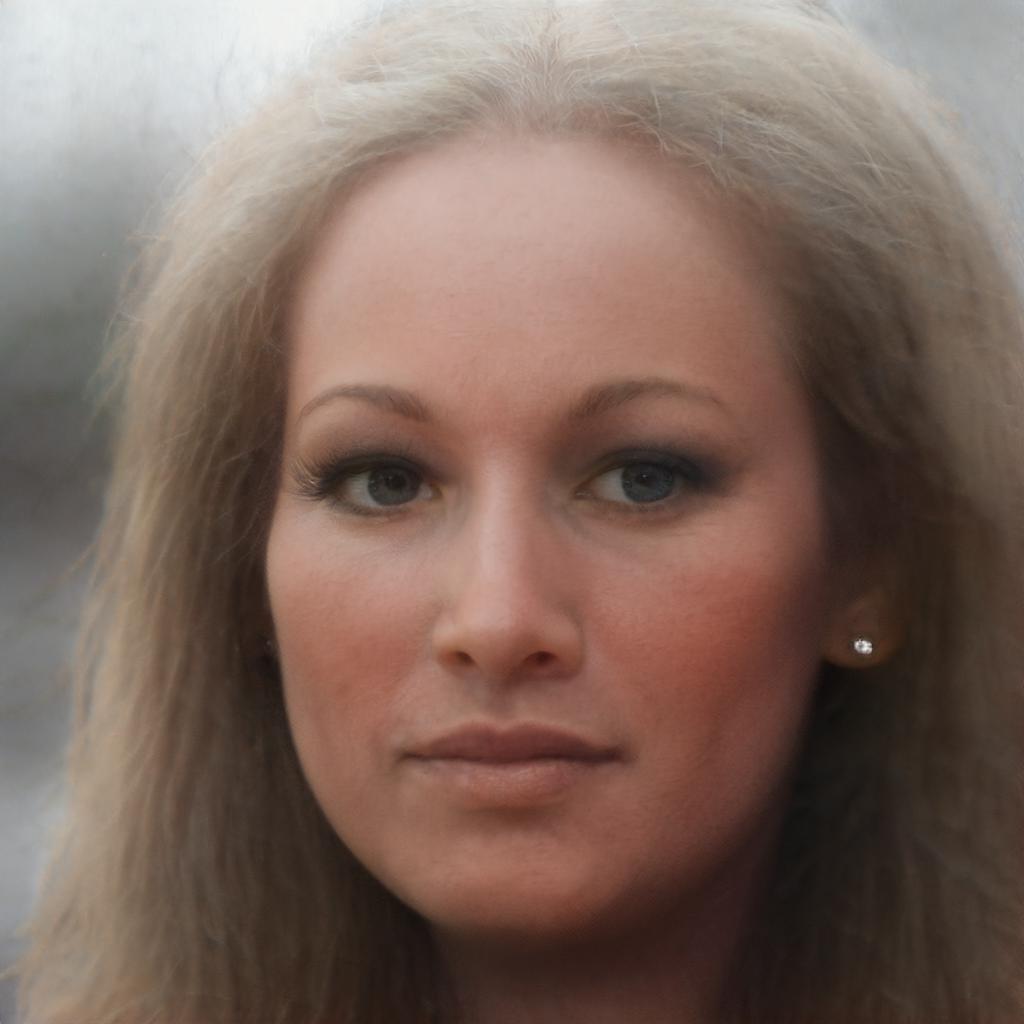}
        \includegraphics[width=1\linewidth]{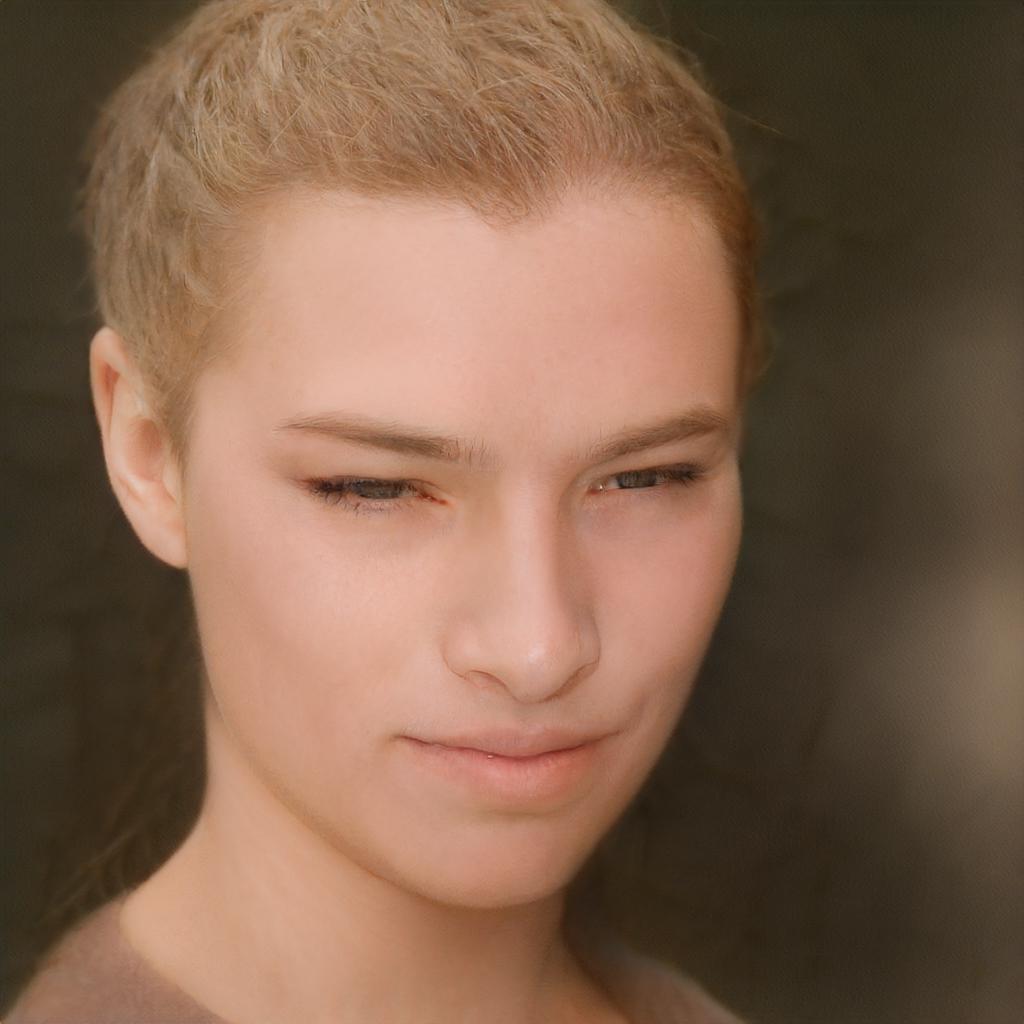}
        \includegraphics[width=1\linewidth]{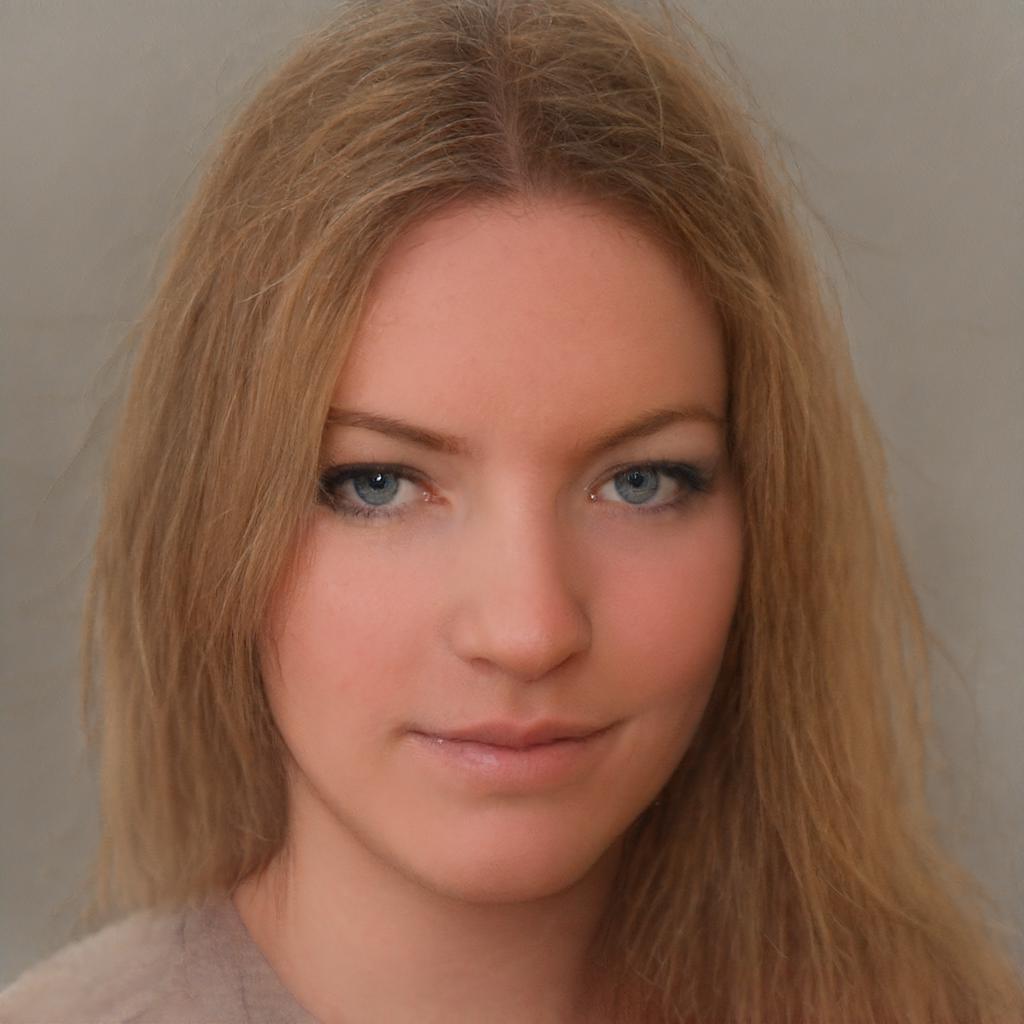}
        \includegraphics[width=1\linewidth]{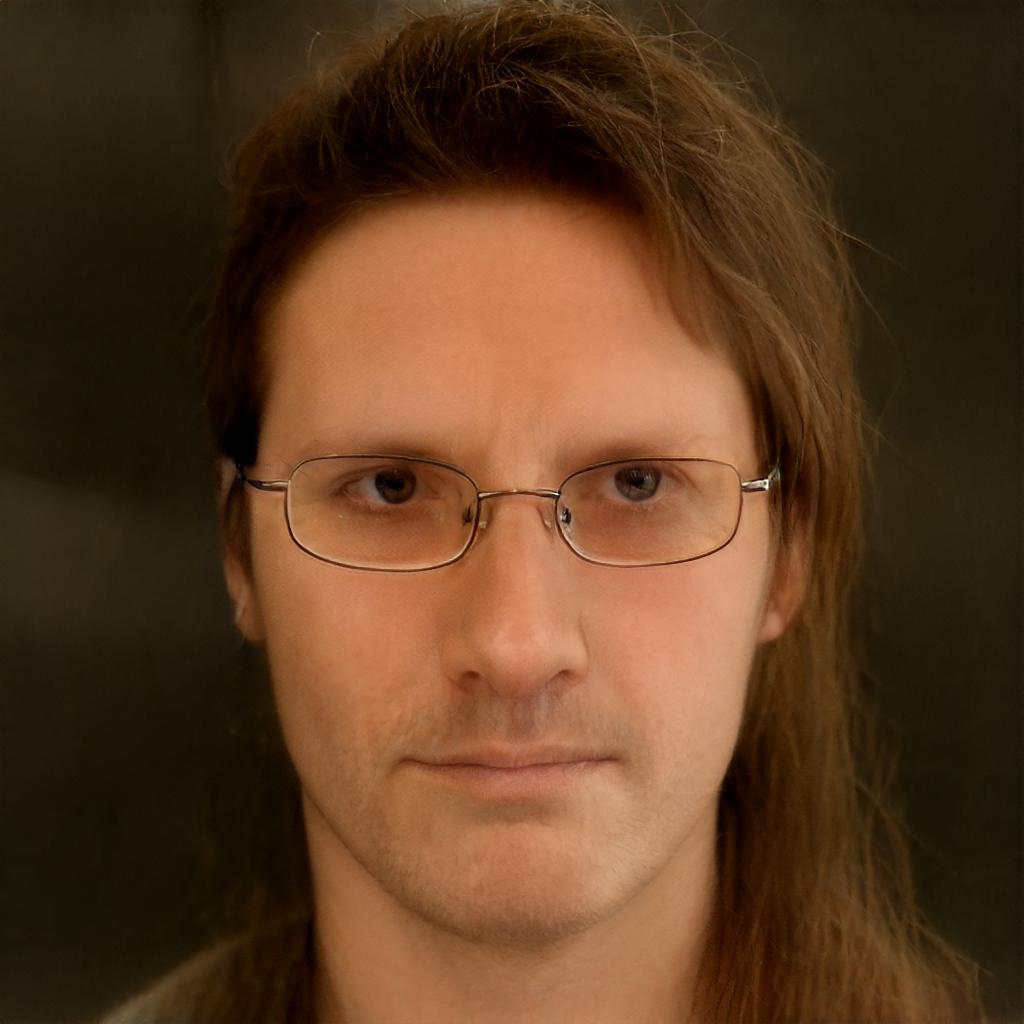}
        \caption{$a_{tar} = -2$}
        \end{minipage}
    \label{fig:comp_input01}
  \end{subfigure}
  \caption{Cross-dataset evaluation of single AU intensity manipulation in CelebA-HQ.
  The descriptions of AUs (from top to bottom) are Outer Brow Raiser, Brow Lowerer, Upper Lid Raiser, Lip Corner Depressor, and Lips Part.
  $a_{tar}$ represents the target intensity.} 
  \label{fig:celeba_single_au}
\end{figure}
\vspace{-1.5em}
\paragraph{Cross-Dataset Evaluation}
Fig. \ref{fig:celeba_single_au} presents the cross-dataset results involving single AU editing with multiple intensity levels on the CelebA-HQ dataset \cite{karras2017progressive}. 
AUEditNet exhibits the capability to achieve consecutive AU intensity manipulation. 
Notably, even in the absence of \uline{negative intensities} during training, AUEditNet produces reasonable editing outcomes. 
For instance, applying negative intensity to AU $5$ (Upper Lid Raiser) results in a generated image with partial eye closure.
Regarding AU $25$ (lips part), where intensity indicates mouth openness, providing negative intensity still maintains the closed configuration, aligning with the case of zero intensity instead of creating unrealistic results. 

\subsection{Quantitative Evaluation}
\label{subsec:quantitative}
\begin{table*}[t]
\centering
\resizebox{0.95\textwidth}{!}{
\begin{tabular}{@{}clccccccccccccc@{}}
\toprule
         &
  Method &
  AU1 &
  AU2 &
  AU4 &
  AU5 &
  AU6 &
  AU9 &
  AU12 &
  AU15 &
  AU17 &
  AU20 &
  AU25 &
  AU26 &
  Avg \\ \midrule \midrule
\multirow{6}{*}{\rotatebox{90}{ICC(3, 1) $(\uparrow)$}} &
  HR \cite{ntinou2021transfer} &
  .56 &
  .52 &
  .75 &
  .42 &
  .51 &
  .55 &
  .82 &
  \textbf{.55} &
  .37 &
  .21 &
  .93 &
  .62 &
  .57 \\
 &
  Aps \cite{sanchez2021affective} &
  .35 &
  .19 &
  \textbf{.78} &
  \textbf{.73} &
  .52 &
  \textbf{.65} &
  .81 &
  .49 &
  \textbf{.61} &
  .28 &
  .92 &
  .67
  &
  .58 \\
 &
  MAE-Face \cite{10439628} &
  \textbf{.740} &
  \textbf{.688} &
  .754 &
  .666 &
  \textbf{.653} &
  .584 &
  \textbf{.877} &
  .527 &
  .589 &
  \textbf{.331} &
  \textbf{.952} &
  \textbf{.721} &
  \textbf{.674} \\\cmidrule(l){2-2}
 & 
  DeltaEdit \cite{lyu2023deltaedit} &
  .091 &
  .058 &
  .114 &
  .034 &
  .383 &
  .065 &
  .694 &
  .008 &
  .004 &
  .041 &
  .581 &
  .166 &
  .179 \\
 &
  ReDirTrans \cite{jin2023redirtrans} &
  \textbf{.856} &
  \textbf{.631} &
  .851 &
  .436 &
  \textbf{.634} &
  \textbf{.278} &
  .862 &
  \textbf{.364} &
  .602 &
  .481 &
  .927 &
  .480 &
  .617 \\
 &
  AUEditNet &
  .848 &
  .559 &
  \textbf{.874} &
  \textbf{.600} &
  .577 &
  .230 &
  \textbf{.890} &
  .276 &
  \textbf{.669} &
  \textbf{.511} &
  \textbf{.950} &
  \textbf{.548} &
  \textbf{.628} \\ \midrule
\multirow{9}{*}{\rotatebox{90}{MSE $(\downarrow)$}} &
  HR \cite{ntinou2021transfer} &
  .41 &
  .37 &
  .70 &
  .08 &
  .44 &
  .30 &
  .29 &
  .14 &
  .26 &
  .16 &
  .24 &
  .39 &
  .32 \\
 &
  Aps \cite{sanchez2021affective} &
  .68 &
  .59 &
  \textbf{.40} &
  \textbf{.03} &
  .49 &
  \textbf{.15} &
  .26 &
  .13 &
  .22 &
  .20 &
  .35 &
  \textbf{.17} &
  .30 \\
 &
  MAE-Face \cite{10439628} &
  \textbf{.200} &
  \textbf{.186} &
  .514 &
  .032 &
  \textbf{.320} &
  .222 &
  \textbf{.221} &
  \textbf{.093} &
  \textbf{.204} &
  \textbf{.146} &
  \textbf{.164} &
  .260 &
  \textbf{.213} \\\cmidrule(l){2-2}
 &
  DeltaEdit \cite{lyu2023deltaedit} &
  .605 &
  .686 &
  1.311 &
  .031 &
  .513 &
  .485 &
  .570 &
  .080 &
  .424 &
  .454 &
  1.157 &
  .420 &
  .561 \\
 &
  ReDirTrans \cite{jin2023redirtrans} &
  \textbf{.181} &
  \textbf{.397} &
  .341 &
  .034 &
  \textbf{.453} &
  \textbf{.552} &
  .286 &
  \textbf{.070} &
  .225 &
  .333 &
  .247 &
  \textbf{.367} &
  .290 \\
 &
  AUEditNet &
  .191 &
  .445 &
  \textbf{.309} &
  \textbf{.029} &
  .492 &
  .579 &
  \textbf{.228} &
  .080 &
  \textbf{.188} &
  \textbf{.322} &
  \textbf{.169} &
  \textbf{.367} &
  \textbf{.283} \\ 
  \cmidrule(l){3-15}
  &
  ReDirTrans (\textit{N}) &
  \textbf{.045} &
  .117 &
  \textbf{.025} &
  \textbf{.019} &
  \textbf{.024} &
  .009 &
  .300 &
  .032 &
  .177 &
  \textbf{.032} &
  .803 &
  .427 &
  .167 \\
 &
  AUEditNet (\textit{N}) &
  .069 &
  \textbf{.101} &
  .098 &
  .024 &
  .036 &
  \textbf{.006} &
  \textbf{.227} &
  \textbf{.004} &
  \textbf{.014} &
  .063 &
  \textbf{.351} &
  \textbf{.228} &
  \textbf{.102} \\ \bottomrule
\end{tabular}}
\caption{Comparison to the state-of-the-art action unit (AU) intensity estimation and editing methods on DISFA \cite{mavadati2013disfa}. 
The `Method' column under each metric is categorized into two parts: 1. Upper part: AU intensity estimation methods; 2. Lower part: AU intensity editing methods. 
In the estimation task, we evaluate the performance by comparing the estimated intensities of the input image to the ground truth. 
For the editing task, the procedure begins with the editing of the input image based on the target conditions.
Then we acquire the estimated AU intensities from the edited image via the external pretrained estimator $H_{est}$. 
Finally, we compare these estimated intensities with the provided target conditions. 
`(\textit{N})' denotes the results obtained after the source attribute removal, where all AU intensities are set to zero. 
Each group's best result is highlighted in bold. 
Without extra facial data, MAE-Face becomes MAE-IN1k \cite{10439628}, leading to ICC dropping to .599.
}
\label{tab:icc_comparison}
\end{table*}
\paragraph{Accuracy of AU Intensity Editing.}
Table \ref{tab:icc_comparison} presents measurements of ICC and MSE for comparing the estimated AU intensities against ground truth. 
We categorize the methods under each evaluation metric based on their research directions, whether they focus on the estimation or editing of AU intensities in images. 
Among the editing methods, our proposed AUEditNet surpasses state-of-the-art facial attribute editing methods, especially in terms of the average performance across all $12$ AUs. 
When it comes to the performance of deactivating all AUs, AUEditNet achieves a substantial $38.92\%$ improvement in MSE compared to ReDirTrans \cite{jin2023redirtrans}.  
This illustrates the complete and accurate attribute removal process, which, in turn, contributes to enhanced final performance since attribute removal and addition are entirely reversible processes with shared trainable parameters. 

Furthermore, we expand our comparison to include both editing and estimation methods because the editing performance is also assessed using the same AU intensity estimation process. 
Moreover, the external estimator $H_{est}$ is trained on the same data as AU intensity estimation methods.
We still observe that the estimation performance, when evaluated with our edited face images, surpasses that of state-of-the-art AU intensity estimation methods on the DISFA test subset \cite{mavadati2013disfa, mavadati2012automatic}. 
This finding further solidifies the high level of consistency between the provided target intensities and the edited images generated by AUEditNet. 

\vspace{-1.5em}
\paragraph{Identity Preservation and Image Similarity.}
\begin{table}[t]
\resizebox{\linewidth}{!}{
\begin{tabular}{@{}ccccccc@{}}
\toprule
\multirow{2}{*}{Method} &
  \multirow{2}{*}{\begin{tabular}[c]{@{}c@{}}Target\\ Image\end{tabular}} &
  \multirow{2}{*}{} &
  \begin{tabular}[c]{@{}c@{}}Identity\\ Preservation\end{tabular} &
  \multirow{2}{*}{} &
  \multicolumn{2}{c}{\begin{tabular}[c]{@{}c@{}}Image\\ Similarity\end{tabular}} \\ \cmidrule(lr){4-4} \cmidrule(l){6-7} 
                            &          &  & Distance $(\downarrow)$  &  & L2 $(\downarrow)$  & LPIPS $(\downarrow)$  \\ \midrule \midrule
\multirow{2}{*}{\begin{tabular}[c]{@{}c@{}}GAN\\ Inversion \cite{tov2021designing}\end{tabular}}  & Real     &  & .368 &  & .025 & .173 \\
                            & Inverted &  & .278 &  & .011 & .065 \\ 
                            \midrule
\multirow{2}{*}{DeltaEdit \cite{lyu2023deltaedit}}  & Real     &  & \textbf{.396} &  & \textbf{.022} & \textbf{.165} \\
                            & Inverted &  & \textbf{.309} &  & \textbf{.011} & \textbf{.074} \\ 
                            \cmidrule(l){4-7}
\multirow{2}{*}{ReDirTrans \cite{jin2023redirtrans}} & Real     &  & .505          &  & [.024] & .175 \\
                            & Inverted &  & .479          &  & .018          & .153          \\ 
                            \cmidrule(l){4-7}
\multirow{2}{*}{AUEditNet}  & Real     &  & [.468]          &  & .026          & [.174]          \\
                            & Inverted &  & [.435]          &  & [.016]        & [.126] \\ \bottomrule
\end{tabular}}
\caption{Comparison of identity preservation and image similarity in facial attribute editing methods. 
`GAN Inversion' as a baseline illustrates that the accuracy of action unit intensity editing cannot be reflected in the performance of the Image Similarity criteria.
The best performance is indicated in bold, while the second best is highlighted within brackets.}
\label{tab:id_comparison}
\end{table}
Table \ref{tab:id_comparison} summarizes the performance of identity preservation and image similarity given image editing results. 
In addition to comparing the edited images with the real target images, we also conduct a comprehensive comparison using GAN-inverted images as the target. 
All three editing methods focus on the latent code editing, without adjusting the image encoder and generator. 
From the identity perspective, DeltaEdit \cite{lyu2023deltaedit} achieves the best performance, nearly matching the GAN inversion performance.  
However, this is at the cost of AU intensity manipulation accuracy, resulting in \uline{a decline of $71.50\%$ in ICC and $98.23\%$ in MSE} compared to AUEditNet. 
Comparing our AUEditNet with ReDirTrans \cite{jin2023redirtrans}, we observe the identity preservation improvements of $7.33\%$ and $9.19\%$ considering real and inverted images, respectively. 
These results further validate the effectiveness of our method's ability to achieve disentanglement and preserve identity during intensity manipulation. 

Regarding image similarity, DeltaEdit \cite{lyu2023deltaedit} continues to outperform the other two editing methods. 
However, when using the GAN inversion as the baseline to compare the inverted source image with the real or inverted target images separately, we find that the image similarity criteria still maintain good performance, even when dealing with different AU intensities between source and target images. 
In other words, the difference in AU intensities is not reflected over the image similarity. 
When compared to ReDirTrans \cite{jin2023redirtrans}, AUEditNet achieves comparable performance with the real target image and achieves better performance with the inverted one. 
These results further demonstrate AUEditNet's disentanglement ability when achieving AU intensity editing. 

\vspace{-1.5em}
\paragraph{Smile Manipulation.}
\begin{table}[t]
\centering
\resizebox{0.6\linewidth}{!}{
\begin{tabular}{@{}ccc@{}}
\toprule
\multirow{2}{*}{Method}                                        & \multicolumn{2}{c}{Smile Attribute} \\ \cmidrule(l){2-3} 
                                                               & $E_d$ $(\downarrow)$        & $\rho$ $(\uparrow)$  \\ \midrule \midrule
Talk-to-Edit \cite{jiang2021talk}                              & 0.212       & 40.9        \\
StyleFlow \cite{abdal2021styleflow}                            & \textbf{0.099}       & 88.9        \\
Do \textit{et al.} \cite{do2023quantitative} \footnotesize(W/ StyleGAN2)              & 0.103       & 96.9        \\
AUEditNet                                                      & \textbf{0.099}       & \textbf{121.3}      \\ \bottomrule
\end{tabular}}
\caption{Comparison of smile intensity manipulation performance on the FFHQ test dataset. AUEditNet achieves the best performance given identity preservation ($E_d$) and manipulation efficiency ($\rho$).}
\label{tab:quant_comp}
\end{table}

We evaluate smile attribute manipulation using metrics proposed in \cite{do2023quantitative} on the FFHQ dataset \cite{karras2019style}. 
Specifically, we modify the intensities of AU $6$ (Cheek Raiser) and AU $12$ (Lip Corner Puller) across eight levels simultaneously to enable  smile intensity editing \cite{girard2019reconsidering}. 
Table \ref{tab:quant_comp} provides comparisons based on identity preservation ($E_d$) and manipulation efficiency ($\rho$). 
The result indicates that AUEditNet better preserves identity when an attribute undergoes the same quantity of change than others. 

\subsection{Expression Transfer}
\label{subsec:exp_transfer}
\begin{figure}
    \captionsetup[subfigure]{labelformat=empty}
    \captionsetup[subfigure]{justification=centering}
    \centering
        \begin{minipage}{0.3cm}
        \rotatebox{90}{\scriptsize{~~~~~~~~ Anger ~~~~~~~~~~~ Disgust ~~~~~~~~~~ Happiness ~~~~~~~~ Sadness ~~~~~~~~~~~~ Fear ~~~}}
        \end{minipage}%
    \begin{subfigure}[t]{0.187\linewidth}
        \begin{minipage}{1\linewidth}
        \includegraphics[width=1\linewidth]{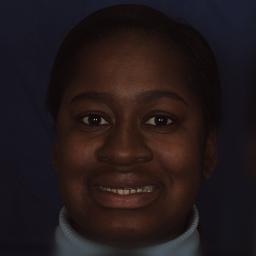}
        \includegraphics[width=1\linewidth]{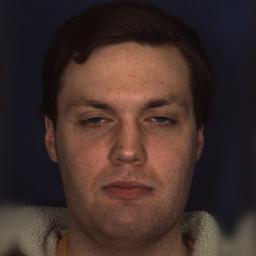}
        \includegraphics[width=1\linewidth]{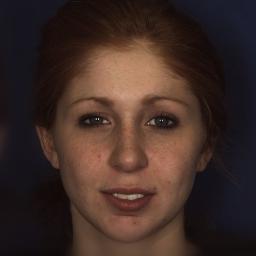}
        \includegraphics[width=1\linewidth]{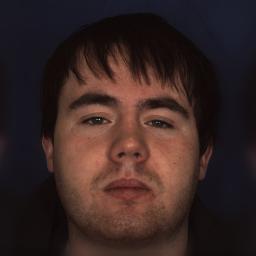}
        \includegraphics[width=1\linewidth]{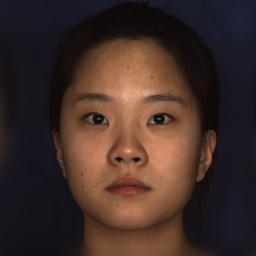}
        \caption{Source}
        \end{minipage}
    \label{fig:comp_input}
  \end{subfigure}
  \hspace{-0.021\linewidth}
  \centering
    \begin{subfigure}[t]{0.187\linewidth}
        \begin{minipage}{1\linewidth}
        \includegraphics[width=1\linewidth]{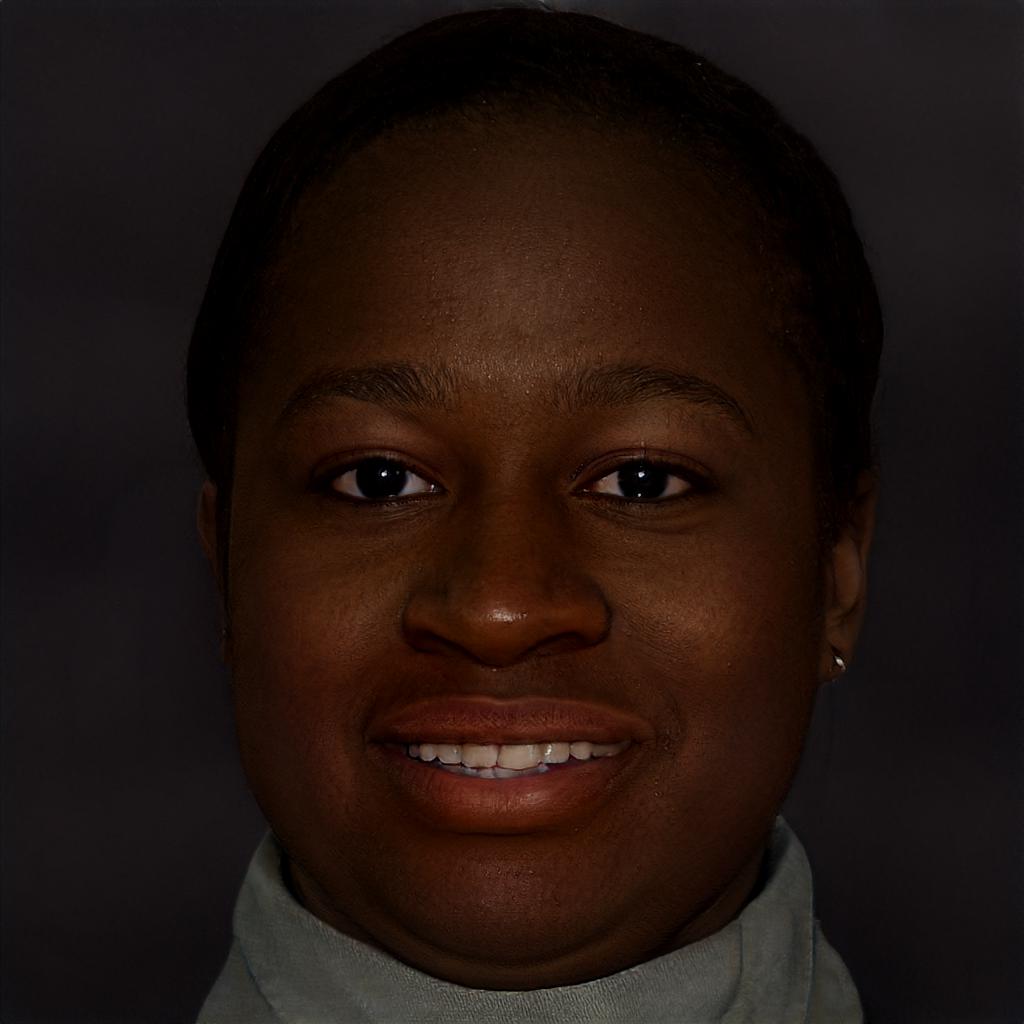}
        \includegraphics[width=1\linewidth]{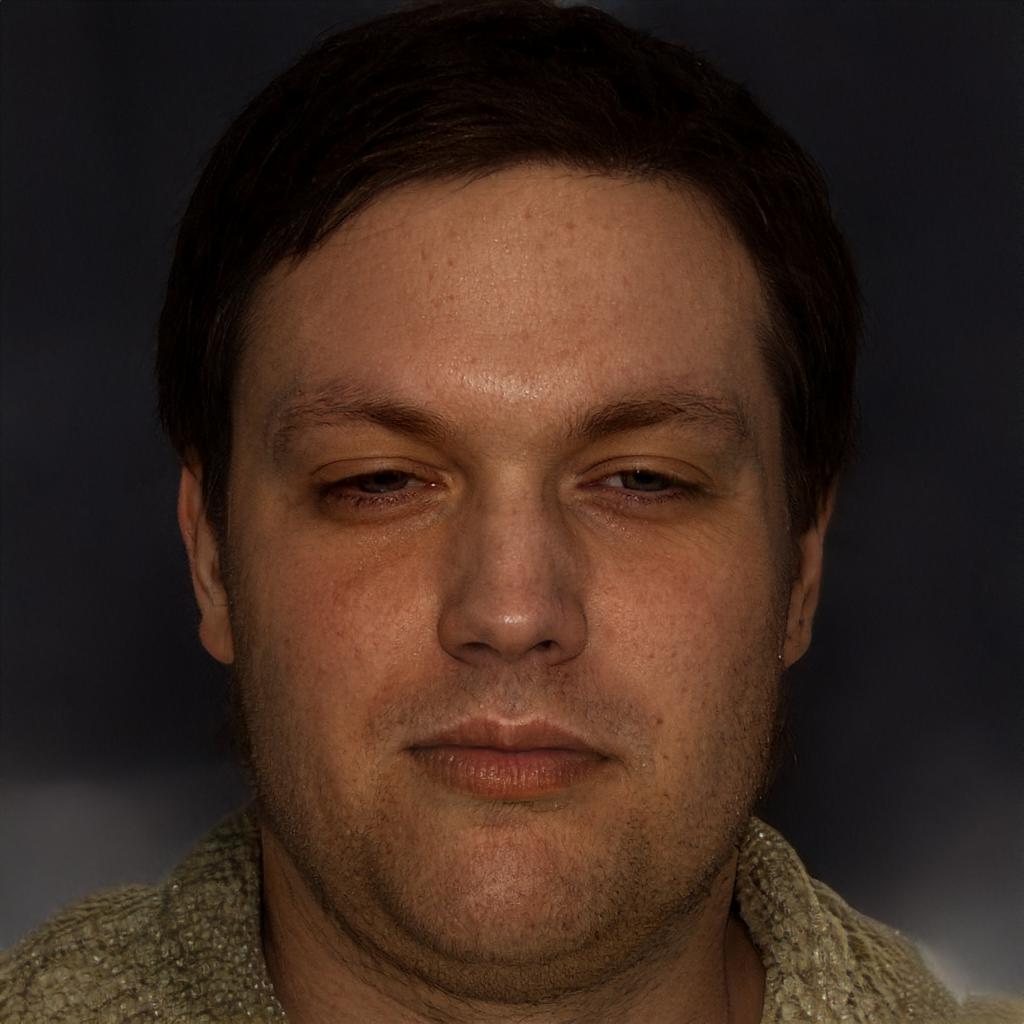}
        \includegraphics[width=1\linewidth]{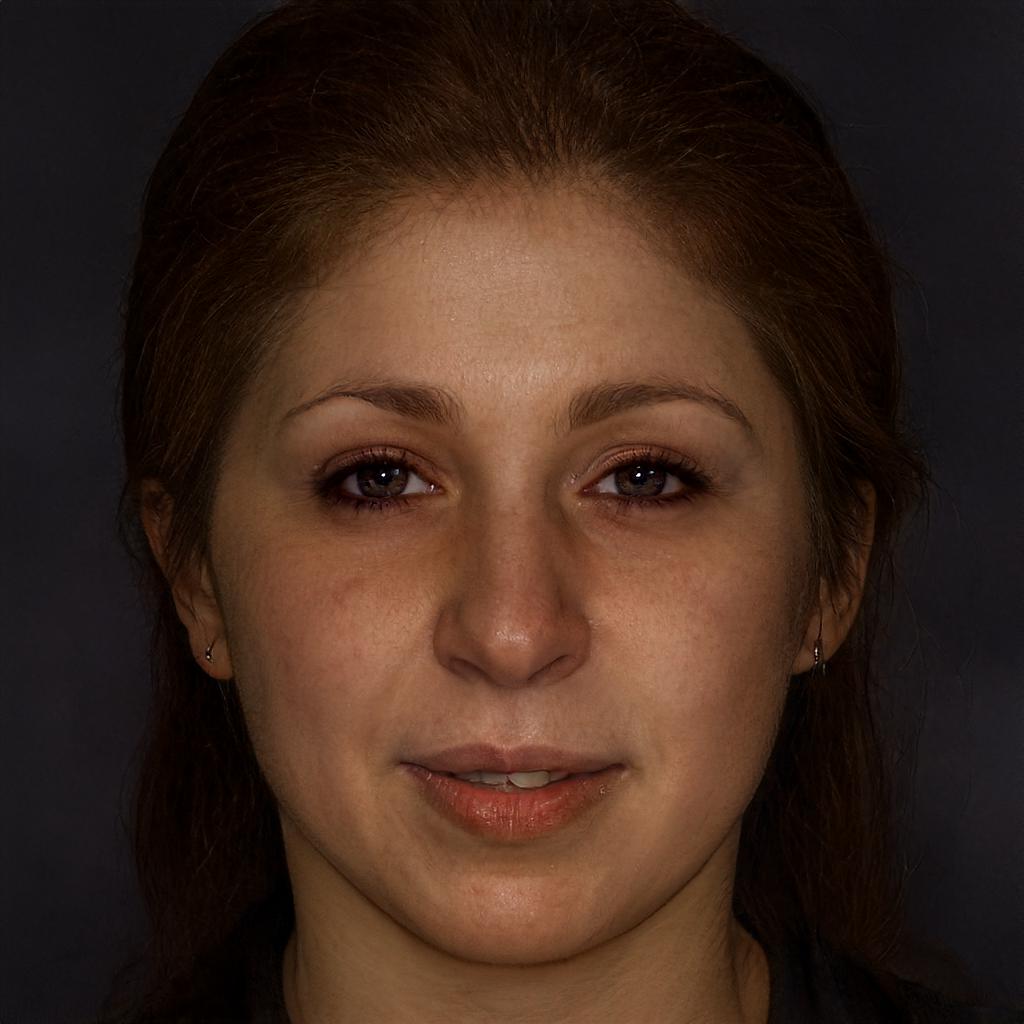}
        \includegraphics[width=1\linewidth]{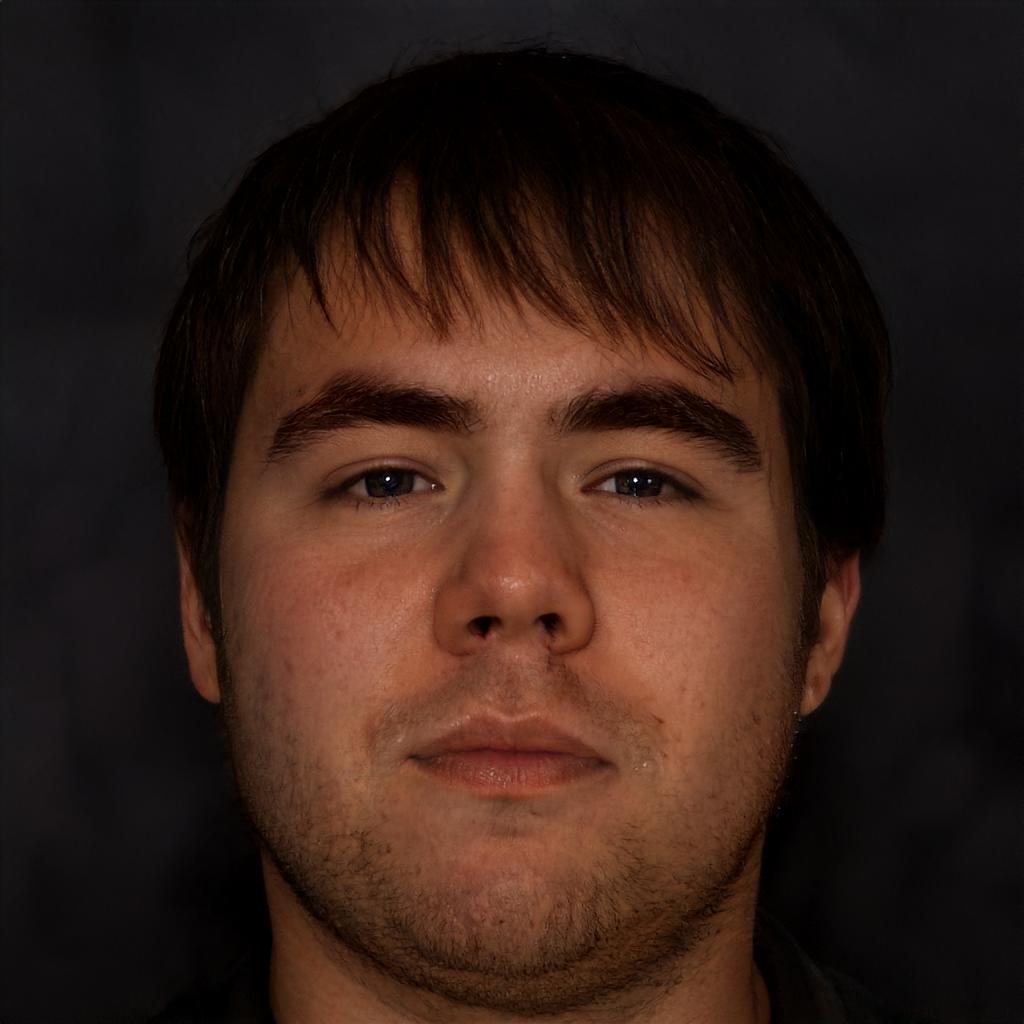}
        \includegraphics[width=1\linewidth]{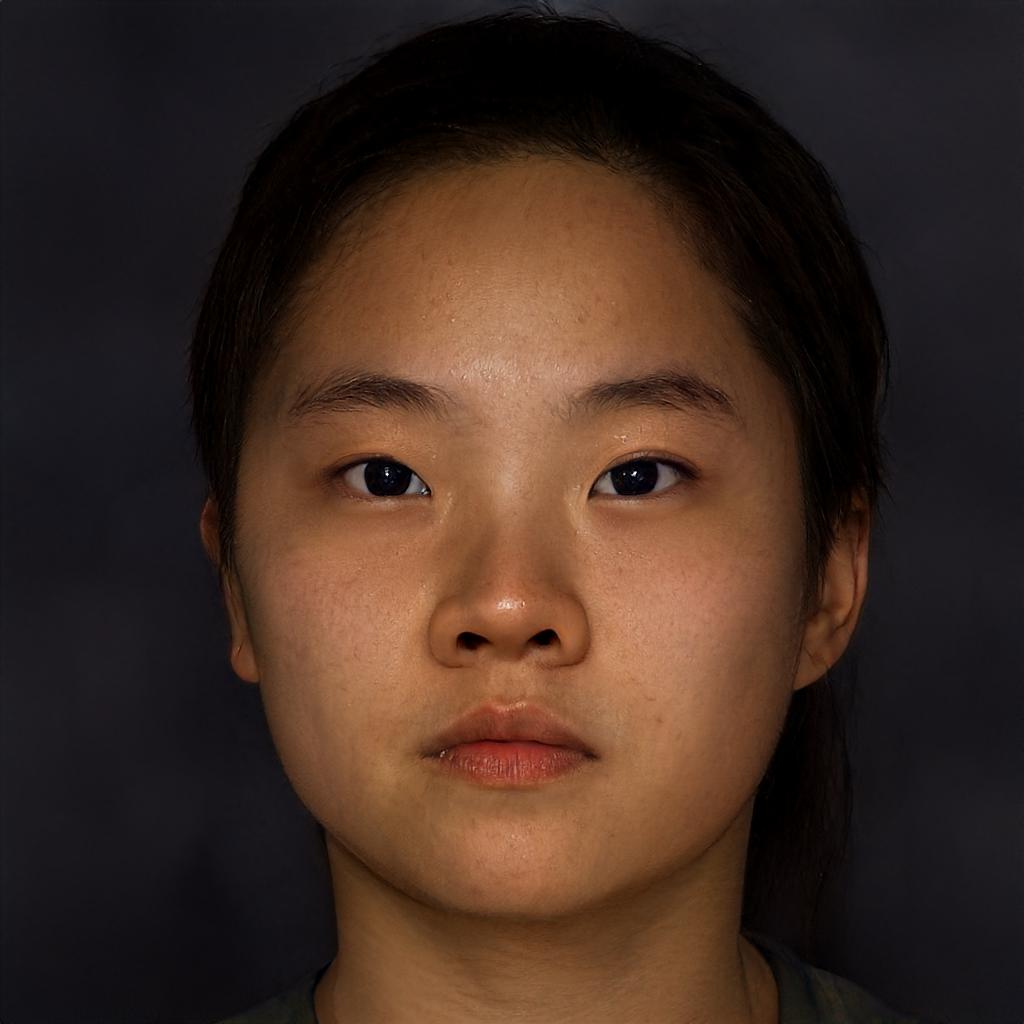}
        \caption{Inversion}
        \end{minipage}
    \label{fig:comp_inver}
  \end{subfigure}
  \hspace{-0.021\linewidth}
  \centering
    \begin{subfigure}[t]{0.187\linewidth}
        \begin{minipage}{1\linewidth}
        \includegraphics[width=1\linewidth]{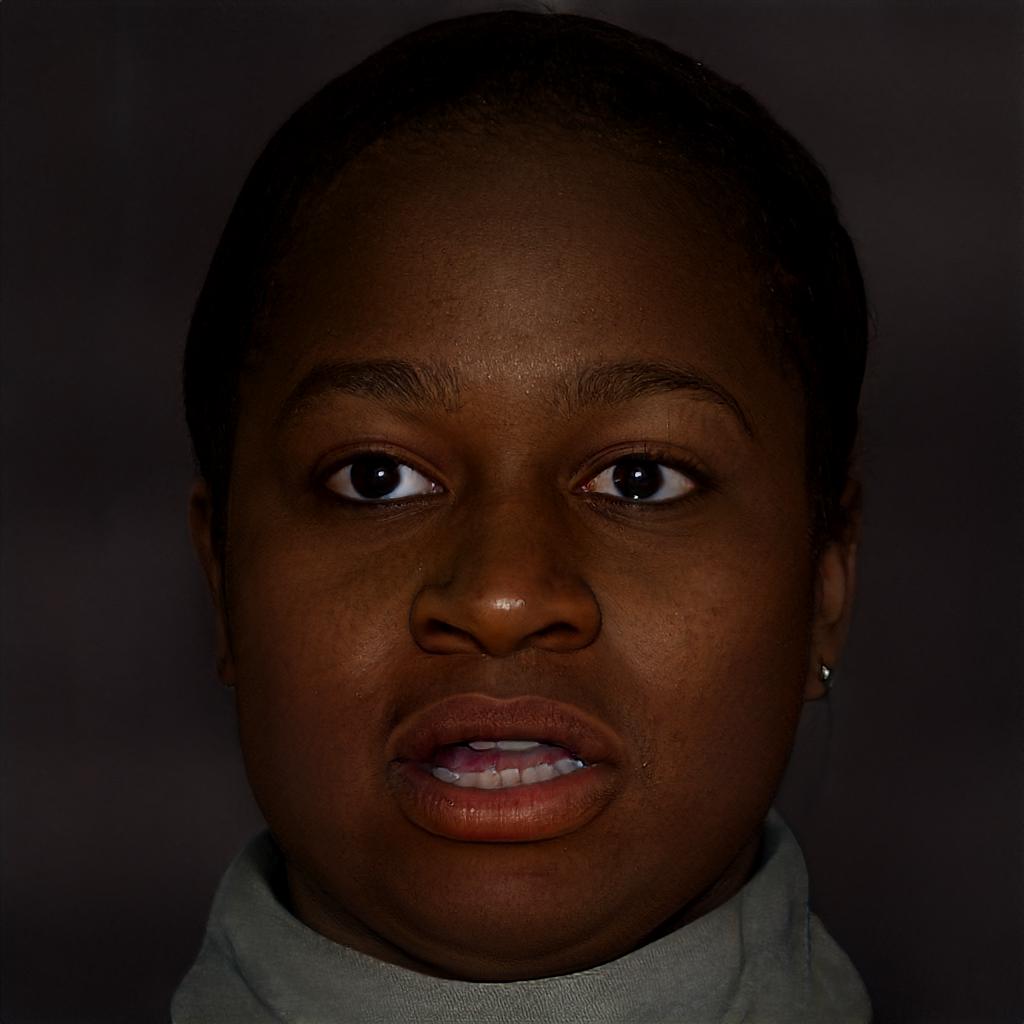}
        \includegraphics[width=1\linewidth]{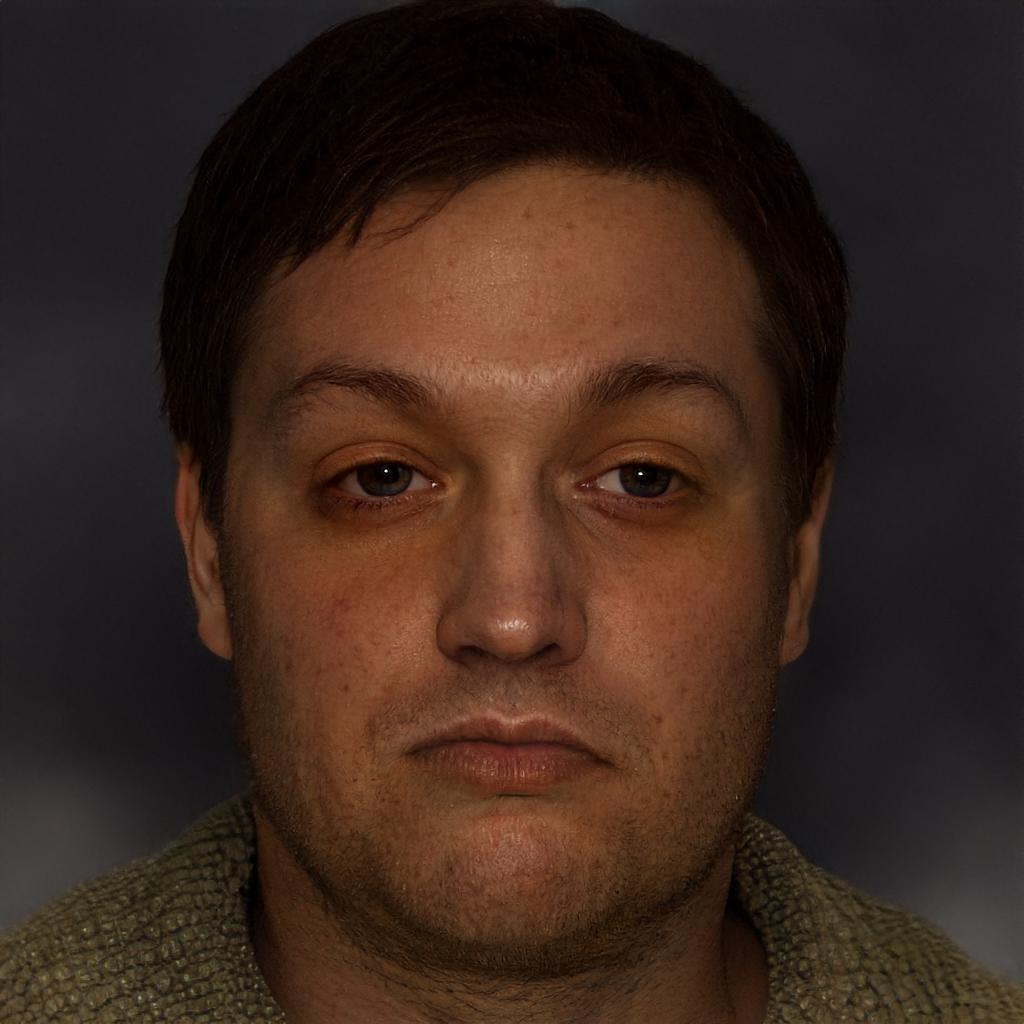}
        \includegraphics[width=1\linewidth]{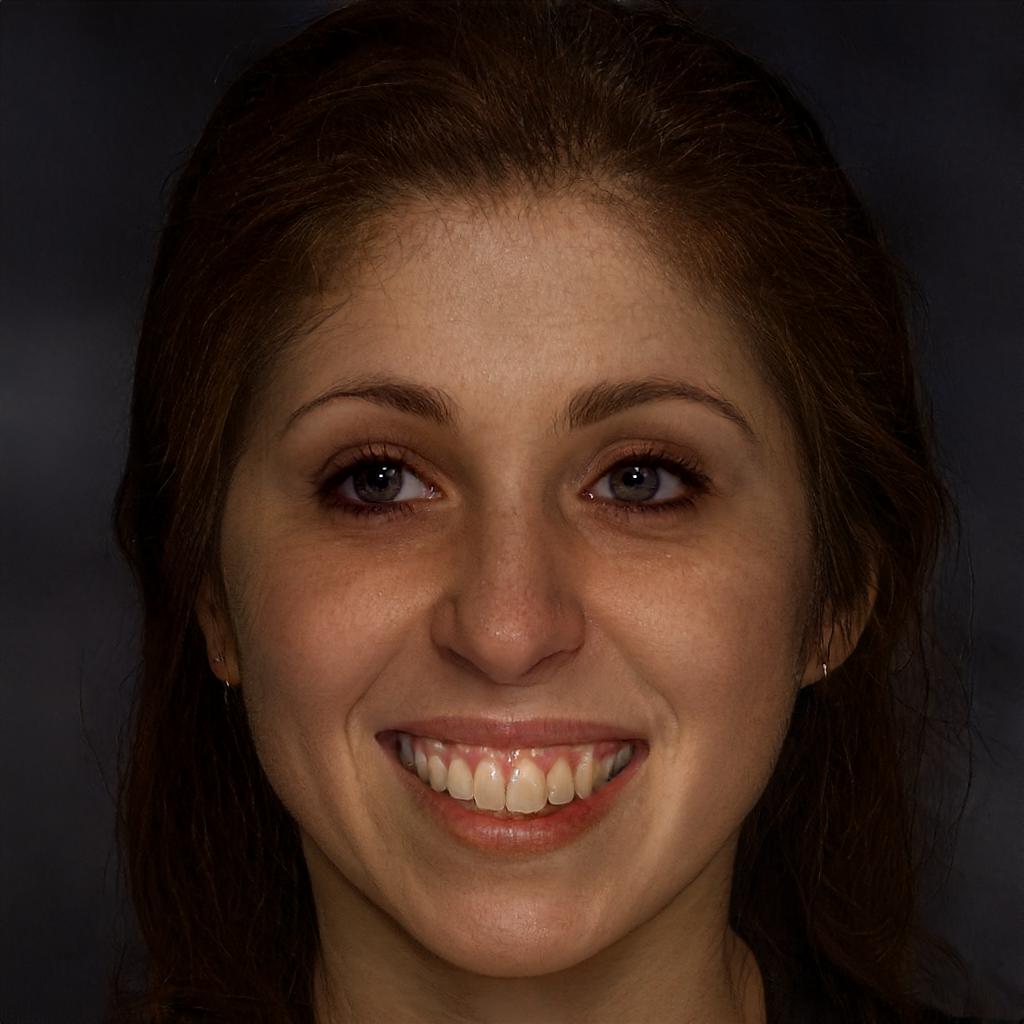}
        \includegraphics[width=1\linewidth]{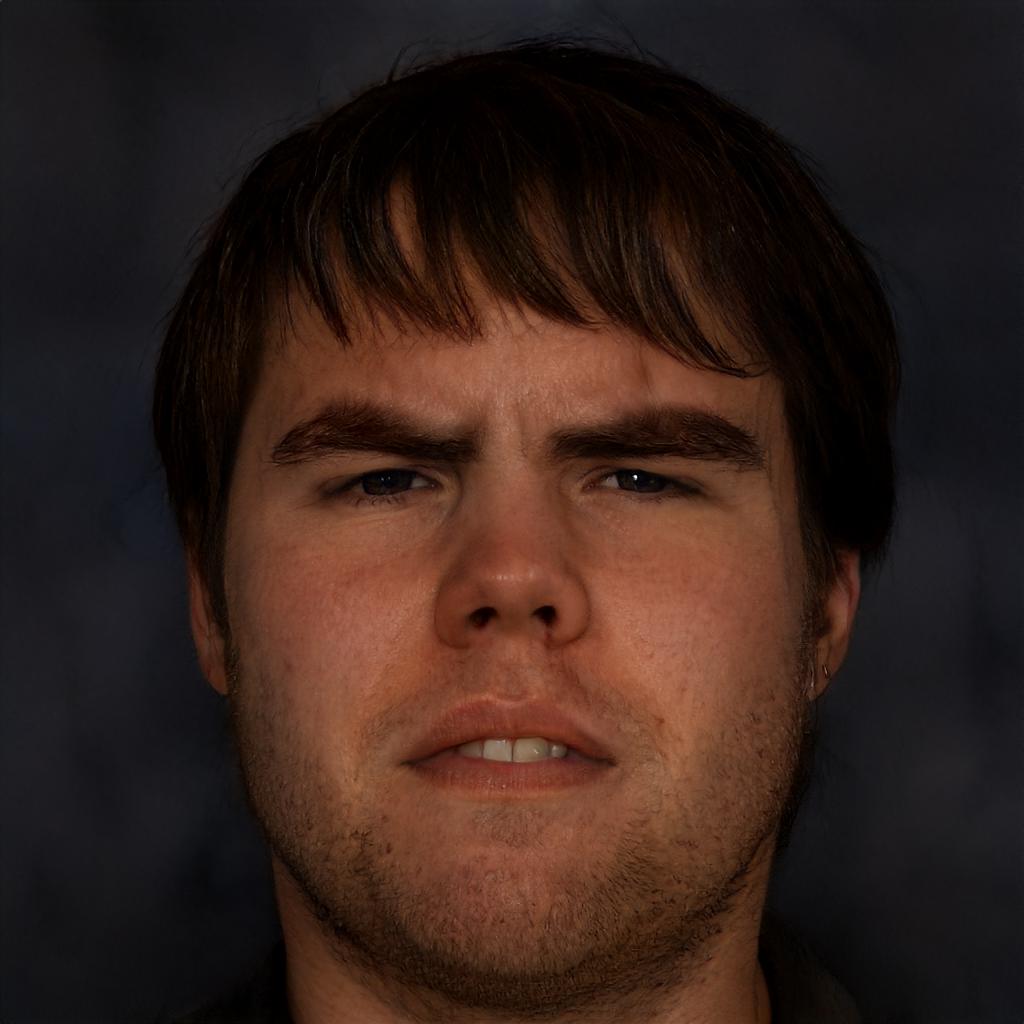}
        \includegraphics[width=1\linewidth]{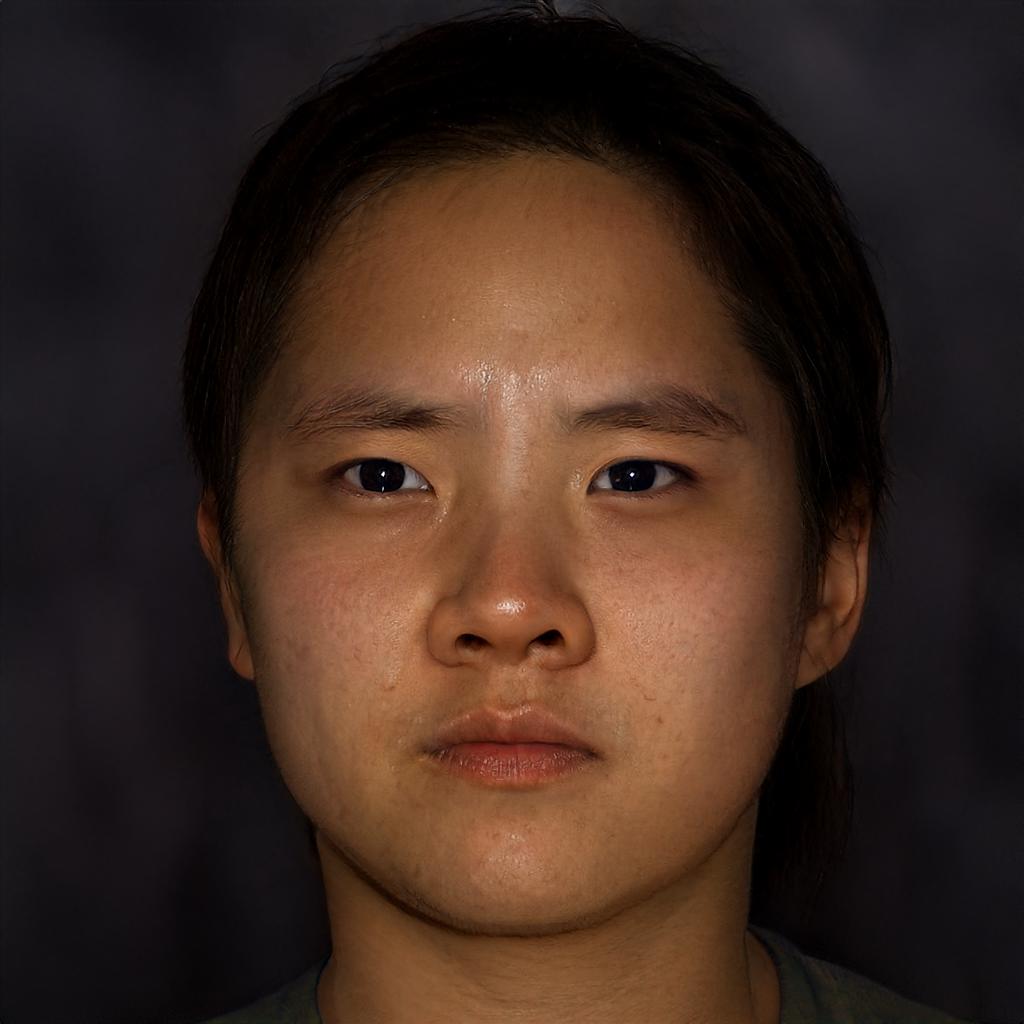}
        \caption{Edited}
        \end{minipage}
    \label{fig:comp_redir}
  \end{subfigure}
  \hspace{-0.021\linewidth}
  \centering
    \begin{subfigure}[t]{0.187\linewidth}
        \begin{minipage}{1\linewidth}
        \includegraphics[width=1\linewidth]{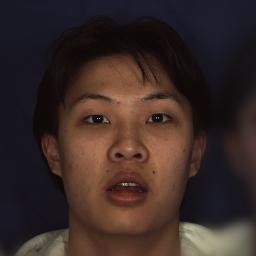}
        \includegraphics[width=1\linewidth]{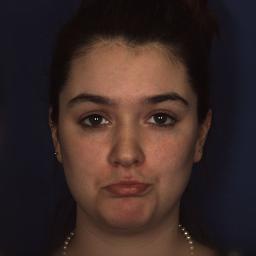}
        \includegraphics[width=1\linewidth]{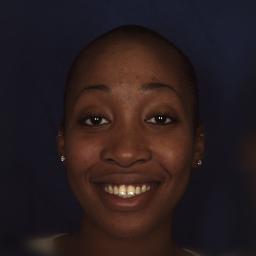}
        \includegraphics[width=1\linewidth]{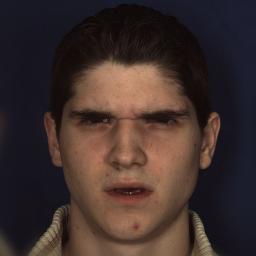}
        \includegraphics[width=1\linewidth]{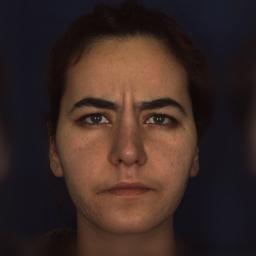}
        \caption{Target}
        \end{minipage}
    \label{fig:comp_tar}
  \end{subfigure}
  \caption{AU intensity manipulation conditioned on target images to achieve facial expression transfer on the BU-4DFE dataset. The fine-grained facial expressions, such as AU $17$ (Chin Raiser) in `Sadness' and AU $25$ (Lips Part) in `Disgust', are transferred accurately.} 
  \label{fig:bu4dfe_exp_transfer}
\end{figure}
Setting individual AU values is a cumbersome process and requires expertise for achieving desired expression synthesis \cite{papantoniou2022neural}. 
In contrast, our proposed AUEditNet demonstrates the capability to directly transfer facial expressions from target images without the need for retraining the network. 
The process involves inputting the target image with the desired expression into the target branch in Fig. \ref{fig:network}. 
Instead of employing removal and addition processes, we directly feed the estimated embeddings of the target image into the decoder $\Phi_{dec}^j$, similar to the procedure in the source branch, to acquire editing residuals with target facial expressions.
Fig. \ref{fig:bu4dfe_exp_transfer} shows expression transfer results on the BU-4DFE dataset \cite{zhang2013high}. 
The edited images demonstrate the contributions of AU intensity manipulation to the facial expression reenactment. 

\subsection{Ablation Study}
\begin{table}[t]
\centering
\resizebox{\linewidth}{!}{
\begin{tabular}{@{}crcccccc@{}}
\toprule
\multicolumn{2}{c}{\multirow{2}{*}{Model}} &
  \multicolumn{3}{c}{\begin{tabular}[c]{@{}c@{}}Target\\ (Removal \& Addition)\end{tabular}} &
   &
  \multicolumn{2}{c}{\begin{tabular}[c]{@{}c@{}}Neutral\\ (Removal)\end{tabular}} \\ \cmidrule(lr){3-5} \cmidrule(lr){7-8} 
\multicolumn{2}{c}{}                                     & MSE $\downarrow$   & ICC $\uparrow$  & ID $\downarrow$   &  & MSE $\downarrow$  & ID $\downarrow$   \\ \midrule \midrule
\multirow{3}{*}{\rotatebox{90}{Training}} & w/o $\mathcal{L}_R$ \& $\mathcal{L}_P$           
                                                         & 0.388 & 0.584 & 0.502 &  & 0.253 & 0.440 \\
                          & w/o $\mathcal{L}_F$          & 0.507 & 0.356 & 0.480 &  & 0.467 & 0.454 \\
                          & w/o $\mathcal{L}_{ID}$       & 0.288 & 0.619 & 0.533 &  & 0.115 & 0.515 \\ 
\midrule 
\multirow{4}{*}{\rotatebox{90}{Design}}   & Sngl.        & 0.317 & 0.598 & 0.545 &  & 0.621 & 0.724 \\
                          & + Encoder, Decoder           & 0.290 & 0.617 & 0.505 &  & 0.167 & 0.600 \\
                          & + Dual.                      & 0.288 & 0.617 & 0.471 &  & 0.111 & 0.439 \\ 
                          & + Label Mapping              & \textbf{0.283} & \textbf{0.628} & \textbf{0.468} &  & \textbf{0.102} & \textbf{0.426} \\
\bottomrule
\end{tabular}}
\caption{Ablation Study for AUEditNet.}
\label{rebtab:ablation}
\end{table}
Table \ref{rebtab:ablation} shows the results of ablation studies for AUEditNet.  
In \textbf{module design}, the integration of dual branch (+ Dual.) leads to improvements in both AU manipulation accuracy (MSE, ICC) and ID preservation (ID). 
Notably, in the `Removal' case for neutral face generation, MSE and ID get $33.5\%$ and $26.8\%$ improvements, respectively. 
The evaluation in this removal-only process is valuable for assessing whether unrelated information is introduced into the target AU space during editing, which is often invisible when `Removal and Addition' processes are implemented. 
Level-wise label mapping can further improve manipulation accuracy. 
Regarding \textbf{training loss}, a well-trained AU intensity estimator ($\mathcal{L}_F$) plays \underline{a more crucial role} than a paired target image ($\mathcal{L}_R$ \& $\mathcal{L}_P$). 
This observation aligns with the fact that pixel-wise MSE and perceptual loss may not effectively capture AU motions.
The absence of ID loss leads to a performance drop in ID. 
However, it also loosens constraints on latent code editing, resulting in more accurate AU manipulation.

\section{Conclusion}
\label{sec:conclusion}
In this work, we achieved accurate AU intensity manipulation in high-resolution synthetic face images. 
Our method allows conditioning manipulation on intensity values or target images without retraining the network or requiring extra estimators. 
This pipeline presents a promising solution for editing facial attributes despite the dataset’s limited subject count. 
We validated our method both qualitatively and quantitatively through extensive experiments. 
The performance boost with synthetic augmented data confirms the quality of generated samples, mitigating the challenge of data scarcity.
In the future, we aim to explore weakly- (or self-) supervised methods to further advance AU intensity manipulation.

\clearpage

\normalem
{\small
\bibliographystyle{ieeenat_fullname}
\bibliography{references}
}

\clearpage

\end{document}


\title{\paperTitle}
\author{\authorBlock}
\maketitlesupplementary

\input{12_appendix}

\clearpage
\clearpage
\normalem
{\small
\bibliographystyle{ieeenat_fullname}
\bibliography{11_references}
}